\documentclass[11pt]{article}

\usepackage[final]{acl}

\usepackage{times}
\usepackage{latexsym}

\usepackage[T1]{fontenc}

\usepackage[utf8]{inputenc}

\usepackage{microtype}

\usepackage{inconsolata}

\usepackage{graphicx}

\usepackage{subcaption}
\usepackage{booktabs}
\usepackage{hyperref}
\usepackage{algorithm}
\usepackage{algpseudocode}

\usepackage{amsmath}
\usepackage{amssymb}
\usepackage{mathtools}
\usepackage{amsthm}

\usepackage[utf8]{inputenc} 
\usepackage[T1]{fontenc}    
\usepackage{url}            
\usepackage{amsfonts}       
\usepackage{nicefrac}       
\usepackage[table]{xcolor}
\usepackage{colortbl}
\usepackage{times}
\usepackage{latexsym}
\usepackage{multirow}
\usepackage{xspace}
\usepackage{enumitem}
\usepackage{pifont}
\newcommand{\cmark}{\ding{51}}%
\newcommand{\xmark}{\ding{55}}%
\usepackage{textcomp}
\usepackage{multicol}
\usepackage{wrapfig}
\usepackage{longtable}
\usepackage[frozencache]{minted}
\usepackage{adjustbox}

\definecolor{citecolor}{HTML}{0071bc}
\definecolor{darkblue}{rgb}{0,0.08,0.5}
\definecolor{Gray}{gray}{0.9}
\definecolor{revision}{rgb}{0, 0.0, 0.0}

\newcommand{\cc}{\cellcolor{gray!20}}
\newcommand{\ccrow}{\rowcolor{gray!20}}  
\newcommand{\ms}[1]{{\tiny$\pm$#1}}
\def \name{ReFeri}

\title{Training-free LLM Verification via Recycling Few-shot Examples}

\author{First Author \\
  Affiliation / Address line 1 \\
  Affiliation / Address line 2 \\
  Affiliation / Address line 3 \\
  \texttt{email@domain} \\\And
  Second Author \\
  Affiliation / Address line 1 \\
  Affiliation / Address line 2 \\
  Affiliation / Address line 3 \\
  \texttt{email@domain} \\}

\author{
  Dongseok Lee$^{1}$ \quad Jimyung Hong$^{1}$ \quad
  Dongyoung Kim$^{2}$ \quad Jaehyung Kim$^{1}$ \\  
  $^{1}$Yonsei University \qquad $^{2}$KAIST \\
  \texttt{\{ehdtjr1220, jaehyungk\}@yonsei.ac.kr} \\
}

\begin{document}
\maketitle
\begin{abstract}
Although large language models (LLMs) have achieved remarkable performance, the inherent stochasticity of their reasoning processes and varying conclusions present significant challenges. 
Majority voting or Best-of-N with external verifiers has been explored to mitigate this, but these approaches are limited in applicability or require additional training. 
To address this, we propose a novel framework that \textbf{Re}cycles \textbf{Fe}w-shot examples to ve\textbf{ri}fy LLM outputs (\name{}). 
Our key idea is to utilize the given few-shot examples not only to generate outputs, but also to evaluate the candidate outputs. 
Specifically, \name{} combines a \textit{forward confidence score} with a \textit{backward reconstruction penalty} to select candidates that follow few-shot guidance while avoiding demonstration-specific overfitting.
Experiments with three different LLMs across seven diverse tasks demonstrate that our framework significantly improves average accuracy of LLMs---achieving an average relative gain of 8.2\%---through effective response selection. \footnote{{Code: \href{https://github.com/dongseok1220/Referi}{github.com/dongseok1220/Referi}.}}
\end{abstract}

\section{Introduction}\label{sec:1}

Recently, large language models (LLMs) have shown remarkable performance on complex reasoning tasks, such as math, coding, and robotics \citep{claude3.5,dubey2024llama,openai2024gpto1,team2023gemini}. 
To further enhance the reasoning capacity of LLMs, various approaches have been proposed, ranging from in-context learning at test time \citep{wei2022chain,kojima2022large} to recent reinforcement learning methods \citep{qu2024recursive,guo2025deepseek}. 
However, the stochastic nature of LLMs still poses challenges, since different reasoning paths with varying conclusions can be generated for the same input, making it difficult to identify the optimal response. 
Majority voting approaches \citep{wang2023self, aggarwal2023let} have been widely adopted to mitigate such randomness by aggregating multiple LLM outputs.
However, these approaches are applicable only when the answer can be easily extracted and aggregated. 
Consequently, they are difficult to apply to open-ended text generation tasks, such as code generation and personalized dialogue generation \citep{jain2025livecodebench, salemi2024lamp}.

To address this challenge, selecting the most promising solution among multiple LLM outputs, often termed \textit{Best-of-N}, has gained attention \citep{snell2024scaling, gui2024bonbon}. 
Training external verifiers \citep{cobbe2021training, lightman2024let} is a representative approach, but it necessitates extensive task-specific labeled data and training. 
Consequently, \textit{training-free} alternatives, such as the \textit{LLM-as-Judge} framework \citep{zheng2023judging}, or recent methods leveraging the model's internal confidence \citep{kang2025scalable, wang2024chain}, have been widely explored. 
However, these methods rely on the model's \textit{intrinsic knowledge} encoded within the parameters, which can be miscalibrated or unreliable when the model encounters a new domain or complex reasoning \citep{yuan2024self, zhang2025generative}.
Essentially, asking an LLM to verify its own output in an unfamiliar domain without external guidance is prone to failure.

\begin{figure*}[t]
\centering
\includegraphics[width=1.0\textwidth]{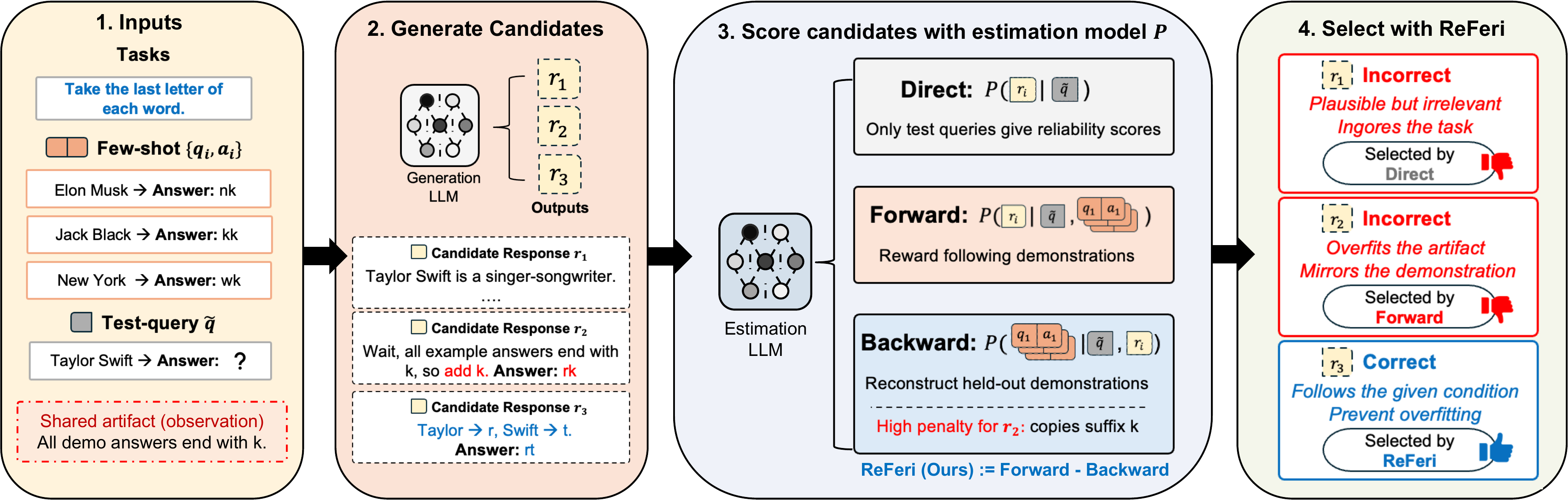}
\caption{
\textbf{An overview of \name{}.} For $N$ candidate responses from LLMs, \name{} evaluates each candidate by combining a forward confidence score (\textit{how well the candidate follows the few-shot examples}) and a backward reconstruction penalty (\textit{that penalizes candidates
that mimic the surface patterns or artifacts of few-shot examples}). 
This example illustrates how \name{} regularizes selection when forward/direct fail.}
\vspace{-0.1in}
\label{fig:method}
\end{figure*}

Motivated by this limitation, we suggest a new perspective: \textit{utilizing few-shot examples to verify and select among multiple LLM outputs}. 
Although few-shot examples are often omitted in recent instruction-tuned LLMs due to their zero-shot capabilities \citep{guo2025deepseek,sprague2025cot}, we argue that they remain the most direct and effective medium for conveying human priors. 
Even if an LLM has not encountered a specific task during training, few-shot examples serve as explicit grounding signals that guide the model on the desired reasoning logic and output format. 

In this work, we propose \textbf{\name{}}, a novel framework that \textbf{Re}cycles \textbf{Fe}w-shot examples to ve\textbf{ri}fy LLM outputs.
The core idea of \name{} is to utilize the few-shot examples not only to generate answers but also to rigorously evaluate them. 
To this end, \name{} exploits the bidirectional relationship between few-shot examples and the target query via a \textit{Bayes-inspired} decomposition.
Specifically, \name{} estimates the quality of each candidate using two complementary signals: (i) \textit{forward confidence score} that measures the likelihood of the candidate given the few-shot examples, ensuring the output follows the provided guidance; and (ii) \textit{backward reconstruction penalty} that suppresses candidates that are \textit{overly specific} to the few-shot demonstrations. 
Intuitively, a large backward penalty implies that the candidate mimics the surface patterns or artifacts (\textit{i.e.}, overfitting) of the few-shot examples, rather than reflecting the underlying reasoning logic. 
Consequently, \name{} enables robust selection without additional model training, effectively balancing the guidance of few-shot examples with the model's generalization capability.

We validate the effectiveness of \name{} across three different LLMs (GPT‑4o, GPT‑4o‑mini, and LLaMA‑3.1‑8B) and seven different benchmarks. 
When selecting one response among five candidates generated by few-shot chain-of-thought prompting, \name{} generally outperforms other training-free methods across all tasks, with an average relative gain of 8.2\% over random selection and 2.6\% over the strongest previous baseline.
Importantly, these gains persist even with a lightweight likelihood estimator, yielding a favorable cost--accuracy trade-off and indicating that \name{}'s effectiveness stems from its Bayes-inspired scoring design rather than estimator capacity alone. 
We further show that \name{} scales reliably with the number of candidate responses, remains robust across estimation models, and generalizes well to open-ended generation tasks. 
We also empirically validate the role of the backward reconstruction penalty in capturing demonstration-specific dependence.
\section{Method}

\subsection{Problem formulation}

Let us denote a given LLM as $\mathcal{M}$ and a target test query as $\widetilde{q}$. 
We assume access to a set of $K$ few-shot examples $\mathbf{X}=\{\mathbf{x}_i\}_{i=1}^{K}, \mathbf{x}_i=({q}_i,{a}_i)$, where each pair consists of a query $q_i$ and its corresponding ground-truth answer $a_i$.
Then, \textit{few-shot prompting} incorporates the few-shot examples $\mathbf{X}$ as the context to guide $\mathcal{M}$ in generating a response:
\begin{equation}
    r_n \sim \mathcal{M}(\widetilde{q},\mathbf{X}),
    \label{eq.generate}
\end{equation}
where multiple non-identical predictions $r_n, n=1,\dots,N$ can be sampled. 
Then, our goal is to identify the optimal response $r_{n^*}$ among $N$ candidates. 
Existing approaches typically employ a scoring function or decision rule for this selection. 
For example, Self-Consistency \citep{wang2023self} aggregates outputs via majority voting, assuming the answer is unique and strictly extractable. 
Alternatively, Best-of-N uses an external verifier $R_\phi$ such as reward models \citep{cobbe2021training,lightman2024let} to score the predictions and select the highest-scoring candidate:
\begin{equation}
    r_{n^*} = \arg\max_{n=1,\dots,N} R_\phi (\mathbf{y}_n),
\end{equation}
where $\mathbf{y}_n=(\widetilde{q}, r_n)$. 
However, these approaches face significant challenges: voting is ill-suited for open-ended generation, while training a task-specific verifier requires expensive labeled data and often fails to generalize to unseen domains.

\subsection{\name{}: Recycling few-shot examples for training-free LLM verification}\label{sec:3.2}

Let $P$ be an estimation model (\textit{e.g.}, a pre-trained LLM) capable of computing the likelihood $P(\mathbf{y}_n) = P(r_n \mid \widetilde{q})$ of a response $r_n$ given a query $\widetilde{q}$.
\footnote{$P$ can be different from the generation model $\mathcal{M}$; if $\mathcal{M}$ is a black-box LLM, using white-box $P$ is necessary.} 
Then, our goal is to identify the best response $r_{n^*}$ that maximizes this likelihood:
\begin{equation}
    {n^*} = \arg\max_{n=1, \dots, N} P(\mathbf{y}_n).\label{eq.argmax}
\end{equation}
We note that likelihood-based selection has shown effectiveness in finding a high-quality reasoning path \citep{wang2024chain}. 
However, directly estimating $P(\mathbf{y}_n)$ is not always effective, as it relies entirely on the model's intrinsic knowledge, which may be uncalibrated in unfamiliar domains (see our empirical results in Appendix \ref{app:add_direct}). 
Furthermore, when there is a mismatch between $\mathcal{M}$ and $P$, the estimated likelihoods can be unreliable.
To address this, we reinterpret $P(\mathbf{y}_n)$ with few-shot examples $\mathbf{X}$, through Bayes’ rule: 
\begin{equation}
    \small
    P(\mathbf{y}_n) = \frac{P(\mathbf{y}_n \mid \mathbf{X}) \cdot P(\mathbf{X})}{P(\mathbf{X} \mid \mathbf{y}_n)}.
\end{equation}
Then, in log-space, this yields a decomposition into two intuitive forward and backward terms:
\begin{equation}
    \small
    \begin{aligned}
         \underbrace{\log P(\mathbf{y}_n)}_{\text{direct}}
        &= \underbrace{\log P(\mathbf{y}_n \mid \mathbf{X})}_{\text{forward}} \\
        &\quad - \underbrace{\big( \log P(\mathbf{X} \mid \mathbf{y}_n) - \log P(\mathbf{X})\big)}_{\text{backward}}.
    \end{aligned}
\label{eq:log-bayes}
\end{equation}
While Eq.~\ref{eq:log-bayes} represents a mathematical identity, we leverage this decomposition as a strategic design lens to overcome the aforementioned reliability issues. 
Instead of depending on the model's potentially uncalibrated confidence, this approach anchors the verification process in the explicit few-shot examples. 
Specifically, \name{} substitutes the uncertain direct likelihood with two signals: a \textit{forward confidence score} and a \textit{backward reconstruction penalty}, which we define below.

\paragraph{Forward confidence score.} 
Intuitively, the confidence of candidate response $r_n$ to test query $\widetilde{q}$ is captured by $\log P(\mathbf{y}_n|\mathbf{X})$; 
this score is high when $r_n$ aligns well with the reasoning patterns in the few-shot examples $\mathbf{X}$. 
This forward score has certain advantages over direct estimation of $P(\mathbf{y}_n)$, as it allows the estimation to be grounded in the few-shot examples and hence reduces the reliance on its prior knowledge alone. 
As a result, the forward score provides a more context-aware and robust estimation, especially important in unfamiliar or domain-shifted scenarios. 
In addition, when the estimation model $P$ is equal to generation LLM $\mathcal{M}$, the forward score can be freely obtained during generation of $r_n$. 
Formally, under the autoregressive LM $P$, the forward score is derived as:
\begin{equation}
    \small
    \begin{aligned}
        S_{\tt Forw}(\mathbf{y}_n)
        &:=\log P(\mathbf{y}_n \mid \mathbf{X}) \\
        &= {\color{revision} \frac{1}{L}}\sum_{t=1}^{L} \log P(r_{n,t} \mid \widetilde{q}, \mathbf X, r_{n,<t}),\label{eq.forward}
    \end{aligned}
\end{equation}
where each candidate response is a sequence of $L$ tokens $r_n = (r_{n,1}, \dots, r_{n,L})$. 

\paragraph{Backward reconstruction penalty.} 

The backward term $\log P(\mathbf{X}|\mathbf{y}_n) - \log P(\mathbf{X})$ measures how much conditioning on the candidate pair $\mathbf{y}_n=(\widetilde{q}, r_n)$ increases the likelihood of reconstructing the few-shot examples. 
A large backward term indicates that the candidate is overly tailored to the few-shot examples, capturing demonstration-specific patterns rather than reasoning specific to the target query; hence, the backward term acts as a penalty.
We note that a large backward term may also reflect reasoning patterns shared with the demonstrations (see Section~\ref{sec:4.3} for further analysis).
Under the assumption of mutual independence between few-shot examples, the backward term can also be derived similarly to Eq.~\ref{eq.forward}. 
To better utilize given few-shot examples, we refine the backward term using a leave-one-out strategy through prompt replacement; we construct new demonstration $\widetilde{\mathbf{X}}_{i}$ by replacing $i$-th example $\mathbf{x}_i=(q_i,a_i)$ with $\mathbf{y}_n$:
\begin{equation}
    \widetilde{\mathbf{X}}_{i} := \mathbf{X}_{-i} \cup \{\mathbf{y}_n\},
    \label{eq.replace}
\end{equation}
where $\mathbf{X}_{-i}$ denotes the few-shot examples excluding $\mathbf{x}_i$. 
Then, by including $\widetilde{\mathbf{X}}_{i}$ during estimation for $\mathbf{x}_i$ as additional input context similar to the forward term, we define the modified backward term:
\begin{equation}
    \small
    \begin{aligned}
        S_{\tt Back}(\mathbf{y}_n)
        &:=\log P(\mathbf{X} \mid \mathbf{y}_n) - \log P(\mathbf{X}) \\
        = &\frac{1}{K}\sum_{i=1}^{K} 
        \big(\log P(a_i \mid q_i, \widetilde{\mathbf{X}}_{i}) - \log P(a_i \mid q_i)\big).\label{eq.backward}
    \end{aligned}
\end{equation}
Including the remaining examples $\mathbf{X}_{-i}$ enables more accurate likelihood estimation for target example $\mathbf{x}_i$ by leveraging the in-context learning capability of $P$ (see Appendix \ref{app:add_ablation}). 
In addition, similar to Eq.~\ref{eq.forward}, $\log P(a_i|q_i)$ and $\log P(a_i|q_i, \widetilde{\mathbf{X}}_{i})$ can be computed via token-level decomposition.

While Eq.~\ref{eq.backward} is theoretically rigorous, computing it for all $K$ examples is expensive.
More importantly, we argue that the backward signal is dominated by the examples most semantically related to the query, while distant examples contribute negligible or noisy signals.
Therefore, we propose a \textit{nearest-neighbor approximation} that focuses verification on the local semantic neighborhood.
Specifically, we identify the single most relevant example $\mathbf{x}_{i^\dagger}$ using a pre-trained embedding model $E$:
\begin{equation}
    \small
    i^\dagger = \arg\max_{i=1,\dots,K}
    \cos\big(E(\widetilde{q}), E(q_i)\big).
    \label{eq:selection}
\end{equation}
Then, we define the approximated backward term:
\begin{equation}
    \small
    \widetilde{S}_{\tt Back}(\mathbf{y}_n) := \log P(a_{i^\dagger} \mid q_{i^\dagger}, \widetilde{\mathbf{X}}_{i^\dagger}) - \log P(a_{i^\dagger} \mid q_{i^\dagger}) 
    \label{eq.new_backward}
\end{equation}
We emphasize that the backward term is not intended as a standalone measure of response quality; rather, it is interpreted jointly with the forward score through their difference (see Section~\ref{sec:4.3}).

\paragraph{Final score.} 
Finally, motivated by Eq.~\ref{eq:log-bayes}, we design our main selection score $S_{\tt Fin}$ by subtracting the backward penalty from the forward confidence score to find the most promising output $r_{n^\star}$:
\begin{equation}
\small
    \begin{aligned}
      &{n^{\star}} = \arg \max_{n=1, \dots, N} S_{\tt Fin}(\mathbf{y}_n), \\
        &S_{\tt Fin}(\mathbf{y}_n):=
         S_{\tt Forw}(\mathbf{y}_n) - \widetilde{S}_{\tt Back}(\mathbf{y}_n).
    \end{aligned}
  \label{eq.combined}
\end{equation}

\noindent The overall algorithm is presented in Algorithm \ref{alg:main}. 
\section{Experiments}\label{sec:4}

\subsection{Setups}\label{sec:4.1}

\paragraph{Datasets.} 
We evaluate our method on seven benchmarks encompassing diverse reasoning paradigms.
(1) \textit{MATH500} \citep{lightman2024let}; a 500-problem subset of MATH \citep{hendrycks2021measuring}.
(2) \textit{MMLU-Pro} \citep{wang2024mmlu}; 4,200 examples, including 300 randomly sampled questions per domain.
(3) \textit{HotpotQA} \citep{yang2018hotpotqa}; 500 samples from \citep{kim2024sure}, a multi‑hop question‑answering benchmark.
(4) \textit{DROP} \citep{dua2019drop}; 500 randomly sampled questions from this reading comprehension benchmark.
(5) GPQA-diamond \citep{rein2024gpqa} (\textit{GPQA}); 198 graduate-level questions for complex reasoning.
(6,7) MuSR \citep{sprague2310musr}; 256 examples in Object Placement (\textit{MuSR-op}) and 250 in Team Allocation (\textit{MuSR-ta}) assessing spatial and relational reasoning. 
Further details are provided in Appendix \ref{app:dataset}.

\begin{table*}[t]
\caption{\textbf{Main results.} Overall performance on seven reasoning benchmarks comparing the proposed \textbf{\name{}} with training-free baselines under three generation LLMs. 
Unless otherwise specified, \name{} uses LLaMA-3.1-8B as the estimation model; 
\name{}$_{\text{1B}}$ denotes the same method with a LLaMA-3.2-1B estimator.
The best and second-best scores within each generation model are highlighted in \textbf{bold} and \underline{underlined}, respectively.}
\centering
\resizebox{\textwidth}{!}{
\begin{tabular}{c|c|ccccccc|c}
\toprule
\multirow{2}{*}{Models} & \multirow{2}{*}{Methods} & MuSR-ta & MuSR-op & GPQA & MATH500 & DROP & HotpotQA & MMLU-PRO & \multirow{2}{*}{Avg.} \\
& & (Acc.) & (Acc.) & (Acc.) & (Acc.) & (EM / F1) & (EM / F1) &  (Acc.) & \\
\midrule
\multirow{8}{*}{\rotatebox{90}{GPT-4o}} 
&Zero-shot CoT   & 67.5 \ms{0.8} & 62.1 \ms{0.4} & 49.5 \ms{0.8} & 77.1 \ms{0.6} & 74.2 \ms{0.8} / 84.9 \ms{0.4} & 37.8 \ms{0.3} / 50.3 \ms{0.4} & 74.0 \ms{0.2} & 63.2 \ms{0.2} \\
&Few-shot CoT    & 87.2 \ms{0.6} & 69.6 \ms{0.6} & 47.3 \ms{1.7} & 75.5 \ms{0.1} & 80.4 \ms{0.2} / 89.0 \ms{0.2} & 44.9 \ms{0.4} / 58.6 \ms{0.3} & 73.7 \ms{0.2} & 68.4 \ms{0.3} \\
&LEAP            & 88.1 \ms{1.6} & 68.0 \ms{1.2} & 47.8 \ms{2.8} & 75.2 \ms{0.4} & 81.0 \ms{0.5} / 89.4 \ms{0.4} & 44.5 \ms{0.6} / 57.8 \ms{0.5} & 73.9 \ms{0.2} & 68.4 \ms{0.4} \\
&USC             & 85.9 \ms{1.9} & 71.5 \ms{0.7} & 48.2 \ms{1.6} & 77.3 \ms{0.3} & 81.8 \ms{0.3} / 90.2 \ms{0.1} & 45.9 \ms{0.3} / 60.2 \ms{0.5} & 74.9 \ms{0.5} & 69.4 \ms{0.5} \\
&CoT-WP          & 88.3 \ms{0.2} & 67.5 \ms{1.4} & 49.5 \ms{1.8} & 78.1 \ms{0.6} & \underline{83.3 \ms{0.1}} / 90.5 \ms{0.8} & 46.3 \ms{0.9} / 59.6 \ms{0.9} & 74.4 \ms{0.3} & 69.6 \ms{0.0} \\
&Self-Certainty  & 88.9 \ms{0.5} & 71.1 \ms{2.1} & 50.2 \ms{0.8} & 77.7 \ms{0.4} & 81.1 \ms{0.8} / 89.4 \ms{0.3} & 43.9 \ms{0.4} / 57.7 \ms{0.1} & 74.6 \ms{0.4} & 69.6 \ms{0.4} \\
&\cc\name{}      & \cc\textbf{90.8 \ms{0.4}} & \cc\underline{73.7 \ms{2.2}} & \cc\textbf{51.8 \ms{0.3}} & \cc\textbf{78.5 \ms{0.6}} & \cc\textbf{83.6 \ms{0.2} / 90.9 \ms{0.6}} & \cc\textbf{46.9 \ms{0.4} / 61.0 \ms{0.7}} & \cc\textbf{75.4 \ms{0.3}} & \cc\textbf{71.5 \ms{0.4}} \\
&\cc\name{}$_{\text{1B}}$ & \cc\underline{90.7 \ms{0.2}} & \cc\textbf{74.1 \ms{3.8}} & \cc\underline{51.0 \ms{1.3}} & \cc\underline{78.3 \ms{0.6}} & \cc\underline{83.3 \ms{0.5}} / \underline{90.7 \ms{0.1}} & \cc\underline{46.5 \ms{0.8} / 60.5 \ms{0.2}} & \cc\underline{75.1 \ms{0.1}} & \cc\underline{71.3 \ms{0.4}} \\
\midrule
\multirow{8}{*}{\rotatebox{90}{GPT-4o-mini}} 
&Zero-shot CoT   & 57.6 \ms{1.3} & 58.9 \ms{2.7} & 41.8 \ms{1.0} & 75.8 \ms{0.6} & 77.2 \ms{0.7} / \textbf{85.1 \ms{0.8}} & 31.5 \ms{0.4} / 41.6 \ms{0.4} & 63.0 \ms{0.1} & 58.0 \ms{0.1} \\
&Few-shot CoT    & 77.5 \ms{0.5} & 60.3 \ms{0.8} & 42.4 \ms{1.1} & 74.7 \ms{0.7} & 76.5 \ms{0.2} / 82.9 \ms{0.2} & 33.6 \ms{0.3} / 44.9 \ms{0.2} & 63.0 \ms{0.2} & 61.1 \ms{0.2} \\
&LEAP            & 74.9 \ms{3.2} & 60.3 \ms{3.2} & 43.6 \ms{0.3} & 74.5 \ms{0.1} & 75.4 \ms{0.4} / 82.5 \ms{0.5} & 33.3 \ms{0.6} / 44.5 \ms{0.5} & 63.1 \ms{0.2} & 60.7 \ms{0.1} \\
&USC             & 76.5 \ms{2.5} & 60.5 \ms{1.0} & \underline{44.6 \ms{1.2}} & 76.4 \ms{1.2} & \underline{78.8 \ms{1.7}} / \underline{85.0 \ms{1.0}} & \underline{35.1 \ms{0.2}} / \underline{47.0 \ms{0.3}} & 64.2 \ms{0.5} & 62.3 \ms{0.5} \\
&CoT-WP          & 79.3 \ms{1.3} & 58.5 \ms{2.6} & 41.9 \ms{0.9} & 77.0 \ms{0.7} & 76.9 \ms{0.6} / 82.7 \ms{0.6} & 34.7 \ms{0.9} / 45.9 \ms{0.9} & \underline{64.6 \ms{0.4}} & 61.8 \ms{0.5} \\
&Self-Certainty  & 81.7 \ms{1.6} & 60.1 \ms{0.6} & 41.6 \ms{2.8} & 76.8 \ms{1.0} & 76.7 \ms{0.4} / 83.1 \ms{0.6} & 34.7 \ms{0.1} / 45.9 \ms{0.4} & 63.9 \ms{0.8} & 62.2 \ms{0.1} \\
&\cc\name{}      & \cc\textbf{83.1 \ms{0.2}} & \cc\underline{62.0 \ms{0.6}} & \cc\underline{44.6 \ms{2.6}} & \cc\textbf{78.2 \ms{0.4}} & \cc\textbf{79.1 \ms{0.5}} / 84.7 \ms{0.8} & \cc\textbf{35.7 \ms{0.4} / 47.4 \ms{0.6}} & \cc\textbf{64.9 \ms{0.4}} & \cc\textbf{63.9 \ms{0.5}} \\
&\cc\name{}$_{\text{1B}}$ & \cc\underline{82.8 \ms{0.4}} & \cc\textbf{62.9 \ms{1.4}} & \cc\textbf{45.3 \ms{1.2}} & \cc\underline{77.7 \ms{0.2}} & \cc78.3 \ms{0.8} / 83.9 \ms{1.0} & \cc\underline{35.1 \ms{0.5}} / 46.6 \ms{0.5} & \cc\underline{64.6 \ms{0.5}} & \cc\underline{63.8 \ms{0.2}} \\
\midrule
\multirow{8}{*}{\rotatebox{90}{LLaMA-3.1-8B}} 
&Zero-shot CoT   & 43.3 \ms{1.5} & 51.9 \ms{1.2} & 20.9 \ms{0.7} & 42.9 \ms{1.3} & 59.7 \ms{0.7} / 65.9 \ms{0.5} & 15.7 \ms{0.6} / 21.7 \ms{0.5} & 40.1 \ms{0.4} & 39.2 \ms{0.2} \\
&Few-shot CoT    & 64.5 \ms{2.2} & 53.4 \ms{0.8} & 23.7 \ms{0.3} & 42.4 \ms{0.5} & 61.1 \ms{0.7} / 66.5 \ms{0.9} & 19.3 \ms{0.3} / 25.4 \ms{0.4} & 39.2 \ms{0.4} & 43.4 \ms{0.5} \\
&LEAP            & 64.7 \ms{3.9} & 53.3 \ms{5.1} & 26.8 \ms{1.8} & 41.9 \ms{0.5} & 57.0 \ms{1.1} / 62.6 \ms{1.3} & 19.9 \ms{0.2} / 26.7 \ms{0.1} & 36.5 \ms{0.7} & 42.9 \ms{1.2} \\
&USC             & 67.7 \ms{4.5} & 54.8 \ms{2.2} & 28.1 \ms{3.1} & 48.5 \ms{1.2} & 68.6 \ms{0.9} / 74.3 \ms{1.3} & \underline{25.5 \ms{1.1}} / \underline{33.4 \ms{1.0}} & \textbf{45.8 \ms{0.4}} & 48.4 \ms{1.1} \\
&CoT-WP          & 71.9 \ms{0.6} & 53.1 \ms{1.4} & 30.5 \ms{2.5} & 48.1 \ms{1.7} & \textbf{71.0 \ms{0.5}} / 75.1 \ms{0.6} & \textbf{25.7 \ms{0.1}} / 33.2 \ms{0.4} & \underline{45.4 \ms{0.7}} & 49.4 \ms{0.4} \\
&Self-Certainty  & 72.4 \ms{3.6} & \underline{55.9 \ms{0.6}} & 30.8 \ms{2.6} & \textbf{50.7 \ms{2.2}} & 70.8 \ms{1.1} / \underline{76.2 \ms{1.0}} & 25.2 \ms{0.5} / 32.9 \ms{1.0} & 44.5 \ms{0.8} & 50.0 \ms{1.1} \\
&\cc\name{}      & \cc\underline{76.5 \ms{3.4}} & \cc\textbf{57.5 \ms{0.5}} & \cc\textbf{33.5 \ms{1.7}} & \cc\underline{50.2 \ms{1.6}} & \cc\underline{70.9 \ms{1.6}} / \textbf{76.3 \ms{0.5}} & \cc\textbf{25.7 \ms{0.7} / 33.6 \ms{0.6}} & \cc44.7 \ms{0.5} & \cc\textbf{51.3 \ms{0.6}} \\
&\cc\name{}$_{\text{1B}}$ & \cc\textbf{76.9 \ms{3.2}} & \cc55.5 \ms{0.8} & \cc\underline{31.8 \ms{2.0}} & \cc50.1 \ms{1.7} & \cc\underline{70.9 \ms{1.4}} / 75.8 \ms{1.0} & \cc25.3 \ms{0.5} / 32.9 \ms{0.6} & \cc44.2 \ms{0.5} & \cc\underline{50.7 \ms{0.5}} \\
\bottomrule 
\end{tabular}
}
\label{tab:main}
\vspace{-0.1in}
\end{table*}

\paragraph{Baselines.}
We compare \name{} against six widely-used methods that require no additional training, with some reflecting different uses of few-shot examples:
(1) \textit{Zero-shot CoT} \citep{kojima2022large} appends a trigger phrase ("Let's think step by step.") to each query without exemplars.
(2) \textit{Few-shot CoT} \citep{brown2020language} prepends a fixed set of few examples.
(3) \textit{LEAP} \citep{zhang2024context} improves few-shot prompting by inducing mistakes on few examples, extracts task-specific principles through self-reflection, and applies these principles to unseen questions. 
(4) \textit{USC} \citep{chen2023universal} asks LLM to select the consistent answer from responses. 
(5) \textit{CoT-WP} \citep{wang2024chain} scores each candidate by the top-1/top-2 token-probability gap at answer positions.
(6) \textit{Self-Certainty} \citep{kang2025scalable} measures the KL-divergence between the model's next token distribution and the uniform distribution as a selection metric. See Appendix \ref{app:baselines}.

\paragraph{Implementation details.}
For the experiments, we use three target LLMs for response generation: \texttt{gpt-4o-2024-08-06} (\textit{GPT-4o}) \citep{openai2024gpt4o}, \texttt{gpt-4o-mini-2024-07-18} (\textit{GPT-4o-mini}) \citep{openai2024gpt4omini}, and \texttt{LLaMA-3.1-8B-Instruct} (\textit{LLaMA-3.1-8B}) \citep{dubey2024llama}.
We generate $N{=}5$ responses per query using a temperature of $1.0$ to encourage diverse candidates.
For Zero-shot CoT, Few-shot CoT and LEAP, we report the average accuracy across five, which can be viewed as randomly selecting the response. 
For USC, CoT-WP, Self-Certainty, and \name{}, we use the same candidates generated from Few-shot CoT and employ LLaMA-3.1-8B as the estimation (or LLM-judge) model.
For computing similarity in the backward term (Eq. \ref{eq:selection}), we employ \texttt{all-mpnet-base-v2}, a lightweight embedding model with 110M parameters.
See Appendix \ref{app:implementation}.

\subsection{Main results}\label{sec:4.2}

Table~\ref{tab:main} summarizes the experimental results on seven reasoning benchmarks across three LLMs. 
We report results averaged across three independent runs for all methods to mitigate the probabilistic nature of LLM sampling. 
Across all LLMs and benchmarks, \name{} improves the average accuracy by {8.2\%} relative to Few-shot CoT, which corresponds to random selection.
Compared to the strongest baseline, Self-Certainty, \name{} achieves an average relative improvement of {2.6\%}. 
To verify that these gains are not an artifact of sampling variation, we conduct two-sided paired $t$-tests on the seven-task average over the three matched runs.
All comparisons against CoT-WP and Self-Certainty favor \name{}, with five reaching $p<.05$ and the remaining one at $p{=}.055$. 
See Appendix ~\ref{sec:significance} for all 95\% confidence intervals and unadjusted $p$-values.

Interestingly, CoT-WP and Self-Certainty provide notable benefits for smaller open-source models (\textit{e.g.}, LLaMA), whose samples are often verbose or stylistically variable; however, their effectiveness diminishes for stronger models such as GPT-4o-mini and GPT-4o.
This is likely because candidates from these models are uniformly high-quality and induce very similar predictive distributions, making entropy-based uncertainty insufficient to separate subtle differences. 
This also suggests confidence-based selection is vulnerable to miscalibration under estimator--generator mismatch.
In addition, we observe that the effectiveness of USC varies substantially across tasks and is sensitive to response ordering (see Appendix \ref{app:prompt_based_result}).

In contrast, \name{} explicitly incorporates few-shot demonstrations for validation via two complementary terms.
Consequently, even where model-internal confidence becomes unreliable due to uniformly confident outputs, \name{} maintains robustness by grounding verification in the provided demonstrations.
This also explains the \name{}$_{\text{1B}}$ result in Table~\ref{tab:main}: the gain primarily reflects \name{}'s Bayes-inspired use of few-shot examples rather than estimator capacity, which helps a 1B estimator preserve most of the improvement.
It also contributes to the cost efficiency of \name{} as shown in the cost analysis in Section \ref{sec:4.3}.

\begin{figure*}[t]
\centering
\begin{subfigure}[t]{0.31\textwidth}
  \centering
  \includegraphics[width=\linewidth]{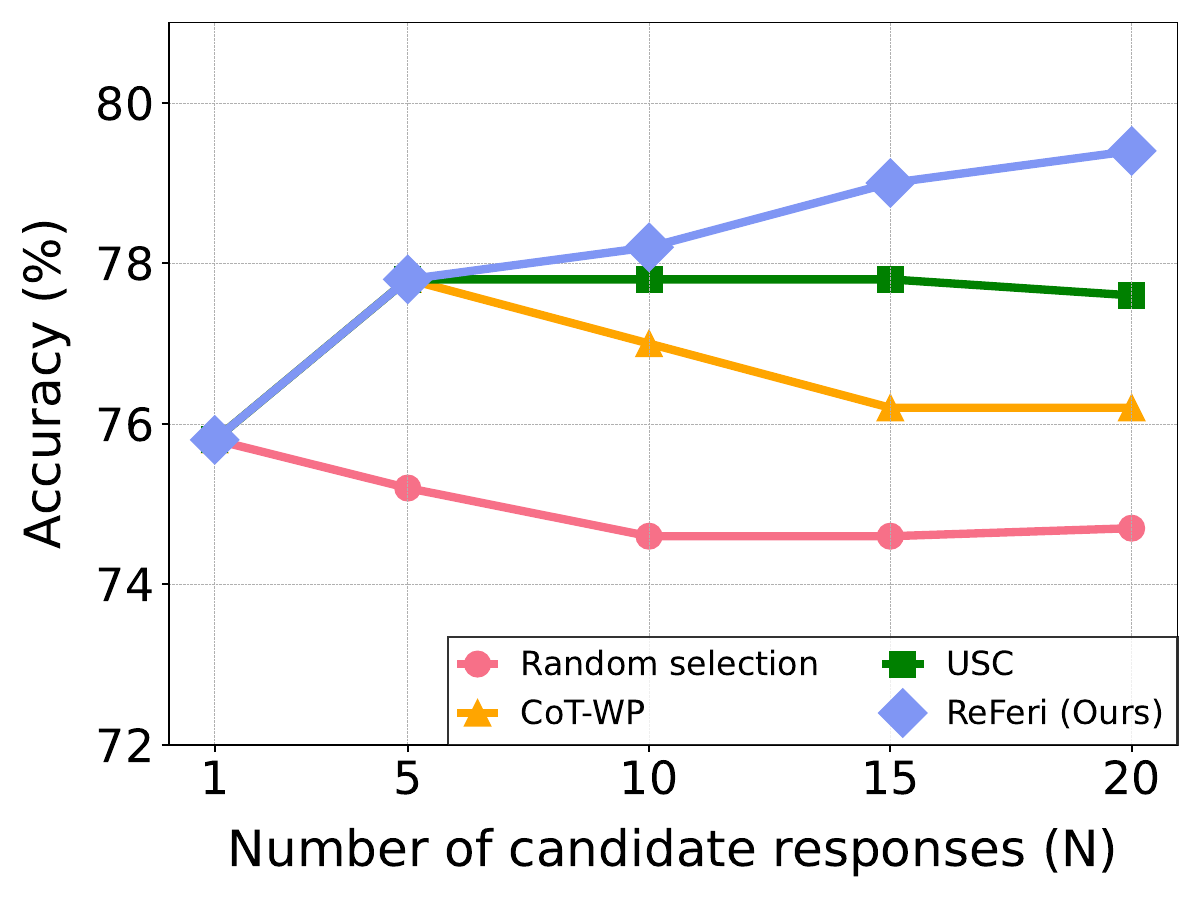}
  \caption{MATH500}
  \label{fig:task_a}
\end{subfigure}\hfill
\begin{subfigure}[t]{0.31\textwidth}
  \centering
  \includegraphics[width=\linewidth]{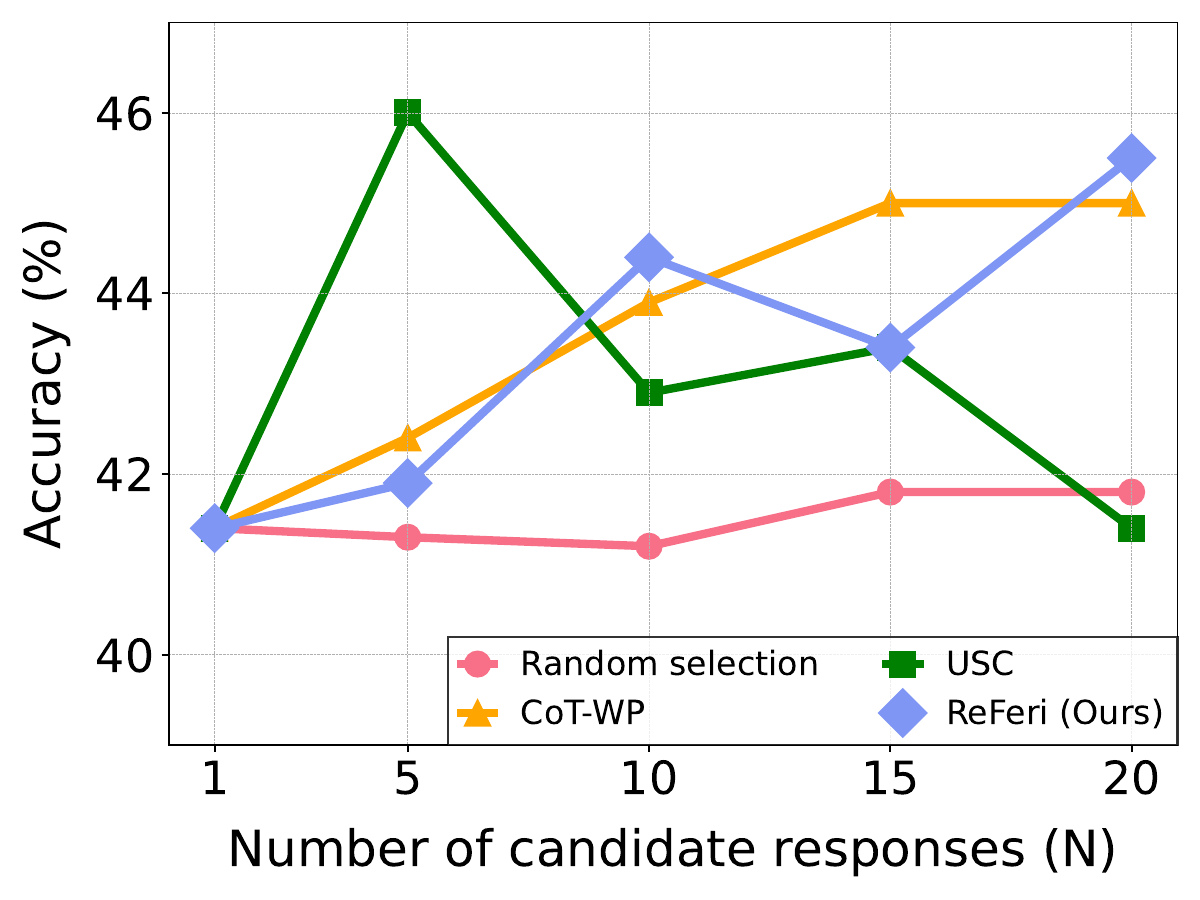}
  \caption{GPQA}
  \label{fig:task_b}
\end{subfigure}\hfill
\begin{subfigure}[t]{0.31\textwidth}
  \centering
  \includegraphics[width=\linewidth]{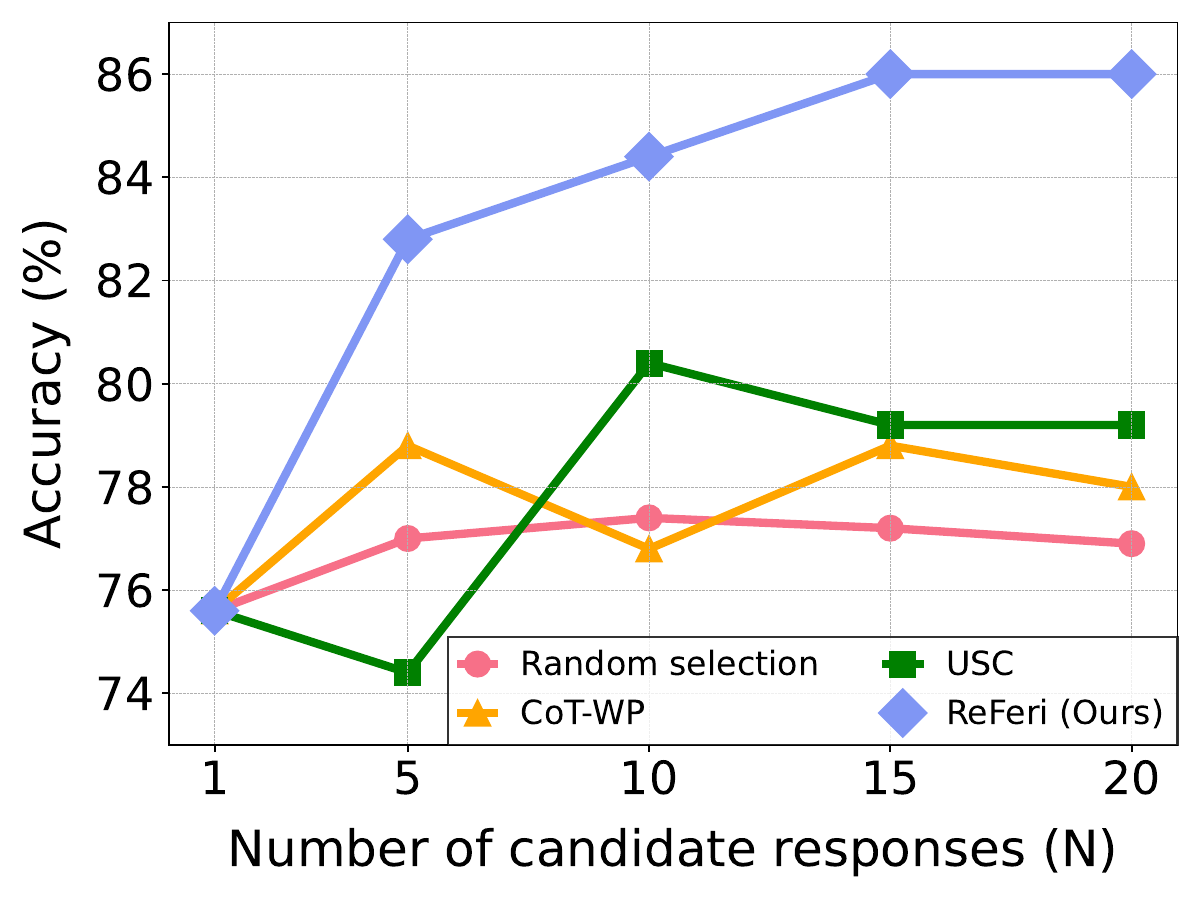}
  \caption{MuSR-ta}
  \label{fig:task_c}
\end{subfigure}
\caption{{\textbf{Test-time scaling.} Accuracy of \name{} versus other training-free selection methods (Random selection, CoT-WP and USC) on {MATH500}, {GPQA}, and {MuSR-ta}. GPT-4o-mini generates different numbers of candidate responses ($N=1,5,10,15, 20$) using Few-shot CoT.}}
\vspace{-0.1in}
\label{fig:best_of_n}
\end{figure*}

\paragraph{Test-time scaling.} 
To assess whether \name{} scales effectively with the number of candidate outputs similar to the conventional Best-of-$N$ method, we evaluate performance as the candidate pool grows.
Specifically, we test $N{=}\{1, 5, 10, 15, 20\}$ candidates on three representative tasks--\textit{MATH500}, \textit{GPQA}, and \textit{MuSR-ta}--using GPT-4o-mini as the generation model under Few-shot CoT. 
As $N$ increases, \name{} yields consistent improvements. 
While random selection accuracy decreases on MATH500, \name{} consistently selects higher-quality responses, improving from 75.8\% at $N{=}1$ to 79.4\% at $N{=}20$.
On GPQA, \name{} raises performance from 41.4\% at $N{=}1$ to 45.5\% at $N{=}20$.
The largest gain is observed on MuSR-ta, where accuracy increases from 75.6\% to 86.0\%.

In contrast, the performance of CoT-WP and USC even degrades as the number of candidates increases, suggesting that these methods do not capture what is truly plausible among the candidates. 
Notably, USC demonstrates strong performance on GPQA when $N{=}5$, but its accuracy declines as $N$ increases, highlighting sensitivity to the candidate set size.
Overall, these results confirm that \name{} scales well with more candidates, demonstrating effectiveness and reliability in test-time scaling.

\begin{table}[t]
\centering
\caption{\textbf{Comparison with Self-Consistency.} Seven-task average accuracy under three generation LLMs. Borda weights each vote by the rank induced by the \name{} score ($p{=}1$).}
\label{tab:self_consistency_main}
\begingroup
\small
\setlength{\tabcolsep}{4.5pt}
\begin{tabular}{@{}lccc@{}}
\toprule
Methods & LLaMA-3.1-8B & GPT-4o-mini & GPT-4o \\
\midrule
Self-Consistency        & 50.1 & \textbf{64.2} & 70.7 \\
\name{}                 & \underline{51.9} & 63.4 & \underline{71.1} \\
\name{} + Borda         & \textbf{52.4} & \underline{64.9} & \textbf{71.3} \\
\bottomrule
\end{tabular}
\endgroup
\vspace{-0.08in}
\end{table}

\subsection{Additional analyses}\label{sec:4.3}
In this section, we present additional analyses of \name{}. 
For experiments, we use candidate responses from the first run in Table \ref{tab:main} except for Table \ref{tab:ablation}.
More results are in Appendix \ref{app:more_exp}.

\paragraph{Comparison with Self-Consistency.}
As noted in Section~\ref{sec:1}, standard \textit{Self-Consistency} \citep{wang2023self} requires answers to be normalized into a discrete format for voting, and is therefore inapplicable to open-ended tasks. 
We therefore use USC as the corresponding baseline in Table~\ref{tab:main}. 
Nevertheless, when answer-level voting is feasible, \name{} can complement Self-Consistency by providing a weight for each candidate's vote. 
As shown in Table~\ref{tab:self_consistency_main}, this weighted voting consistently improves over conventional majority voting across all three generators. 
These results suggest that \name{} can serve not only as a standalone response selector, but also as an auxiliary signal for existing voting-based aggregation methods. 
See Appendix~\ref{app:self_cons} for further details, including full task-wise accuracy.

\begin{table}[t]
\small
\centering
\caption{\textbf{Ablation study.} We report a scoring variant averaged over three runs sampled from LLaMA-3.1-8B responses, corresponding to the setup in Table \ref{tab:main}. The notations F, B, and D stand for the forward, backward, and direct term, respectively.}
\label{tab:ablation}
\begin{adjustbox}{width=\columnwidth}
\begin{tabular}{c|ccc|ccc}
  \toprule
   & F & B & D & MATH500 & GPQA & MuSR-ta \\ \midrule
   & \textcolor{red}\xmark & \textcolor{red}\xmark & \textcolor{red}\xmark & 42.4 & 23.7 & 64.5  \\
   & \textcolor{red}\xmark & \textcolor{red}\xmark & \textcolor{green}\cmark & 50.3 & 32.6 & 68.4  \\
   & \textcolor{green}\cmark & \textcolor{red}\xmark & \textcolor{red}\xmark & 50.1 & 32.3 & 76.9 \\ 
\name{} (Ours)        & \textcolor{green}\cmark & \textcolor{green}\cmark & \textcolor{red}\xmark & 50.2 & 33.5 & 76.5 \\ 
\bottomrule
\end{tabular}
\end{adjustbox}
\vspace{-0.1in}
\end{table}

\paragraph{Ablation study.} 
To understand how each component contributes to \name{}, we conduct an ablation study on the proposed scoring function (Eq. \ref{eq.combined}).
Table \ref{tab:ablation} shows that the direct score can be effective on relatively familiar tasks like MATH500, where the model's inherent prior knowledge is often sufficient.
However, its performance degrades on complex or less familiar benchmarks (\textit{e.g.}, GPQA and MuSR-ta), demonstrating that directly estimating likelihood loses reliability when the model's confidence is miscalibrated.
In contrast, the forward score exhibits superior performance on MuSR-ta, where few-shot guidance is crucial. 
Simultaneously, forward score underperforms the direct score on MATH500 and GPQA, suggesting that conditioning on demonstrations is not always beneficial.
This is because accidental patterns in few-shot examples are not aligned with valid solutions for a given query, leading to an excessive preference for candidates that closely mimic the demonstrations. 
However, incorporating the backward reconstruction penalty addresses this problem, as shown in the last row of Table \ref{tab:ablation}, implying the complementary roles of the forward and backward terms.
Extending this ablation to all three generators and seven tasks, \name{} achieves the best task average for every generator, never below both Direct and Forward score, with consistent gains across the three matched runs ($p<.05$). 
See Appendix~\ref{app:add_ablation}.

\begin{figure}[t]
\centering
\begin{subfigure}[t]{0.49\columnwidth}
  \centering
  \includegraphics[width=\linewidth]{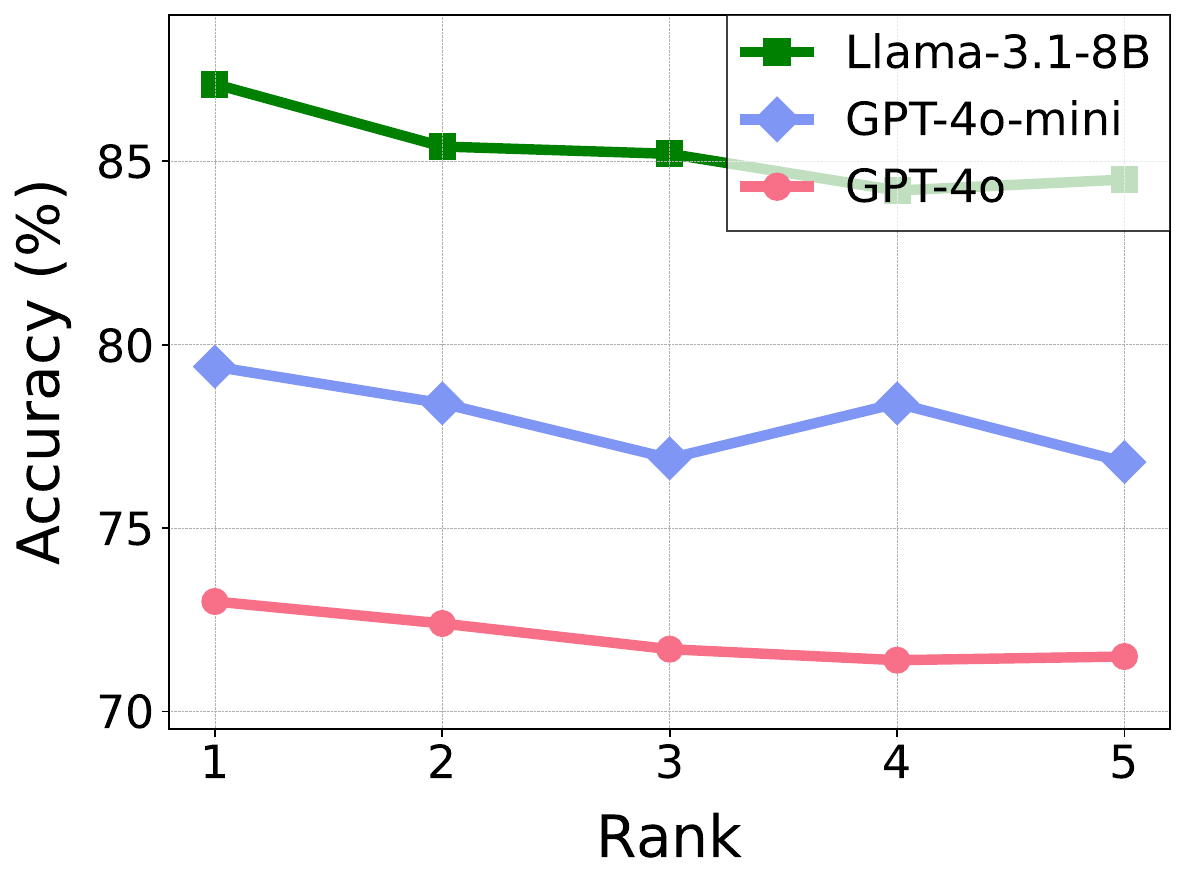}
  \caption{MATH500}
  \label{fig:rank_math500}
\end{subfigure}
\hfill
\begin{subfigure}[t]{0.49\columnwidth}
  \centering
  \includegraphics[width=\linewidth]{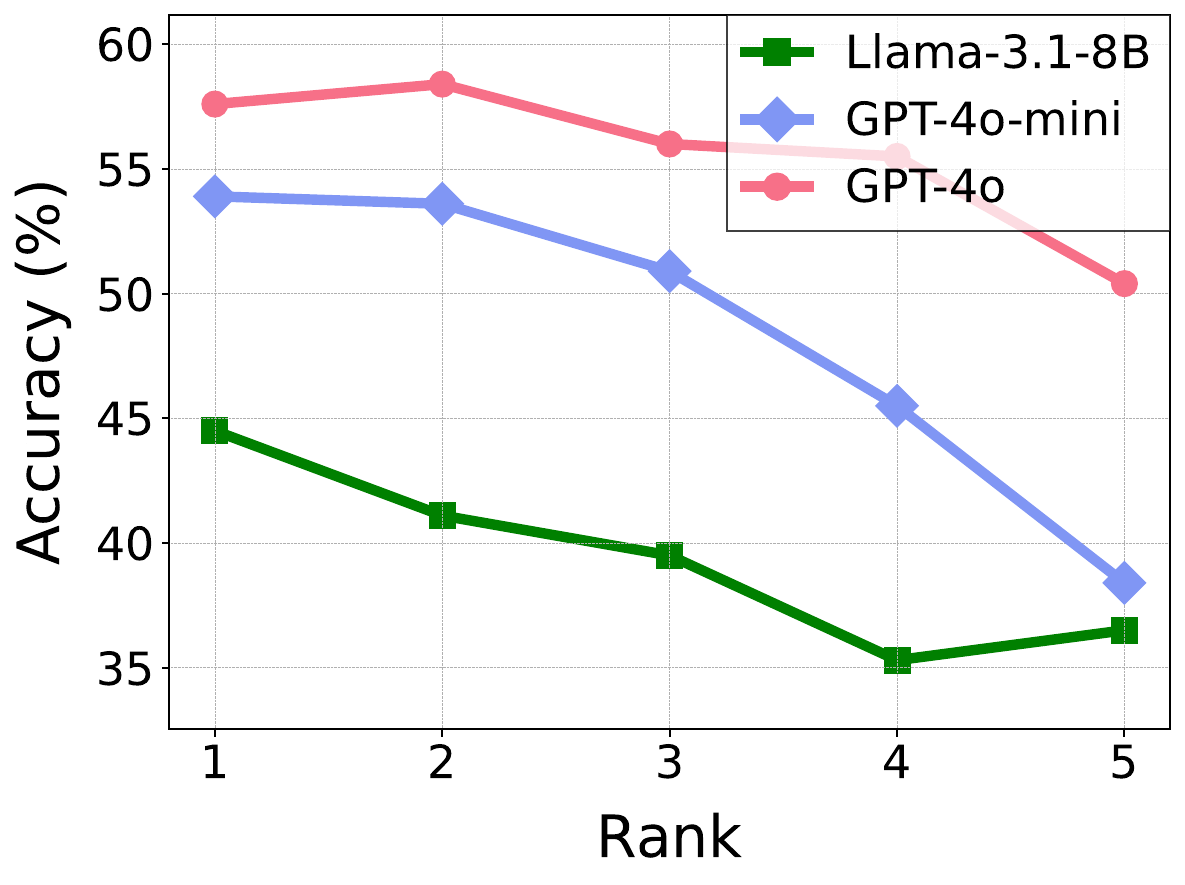}
  \caption{MuSR-ta}
  \label{fig:rank_musrta}
\end{subfigure}
\caption{\textbf{What the backward term captures.} 
Candidates are ranked by the raw backward term, where Rank 1 denotes the largest backward term. 
We measure how well it reconstructs the original few-shot answers.
}
\label{fig:rank_two_tasks}
\vspace{-0.1in}
\end{figure}

\paragraph{Deeper analysis of the backward term.} 
To verify that the backward term captures surface-level overfitting to the few-shot examples, we test the following intuition: candidates with larger backward terms should be more effective at recovering the original few-shot answers.
Concretely, we rank the generated outputs by the raw backward term and use each output as a one-shot demonstration to isolate the effect of candidate response under the same setup (see Appendix \ref{app:add_ablation} for the simplified variant).
For consistency with our pipeline, we employ LLaMA-3.1-8B as the generation model.
As shown in Figure \ref{fig:rank_two_tasks}, accuracy monotonically decreases with rank: candidates with larger backward terms better reconstruct the original few-shot answers, a behavior that is precisely the failure mode we aim to suppress. 
In other words, these are candidates that achieve high scores by reproducing demonstration-specific structures rather than by reasoning about the target query.
Since the forward score already rewards consistency with the few-shot guidance, subtracting the backward term prevents \name{} from selecting such mimicry.

\begin{figure}[t]
\centering
\begin{subfigure}[t]{0.49\columnwidth}
  \centering
  \includegraphics[width=\linewidth]{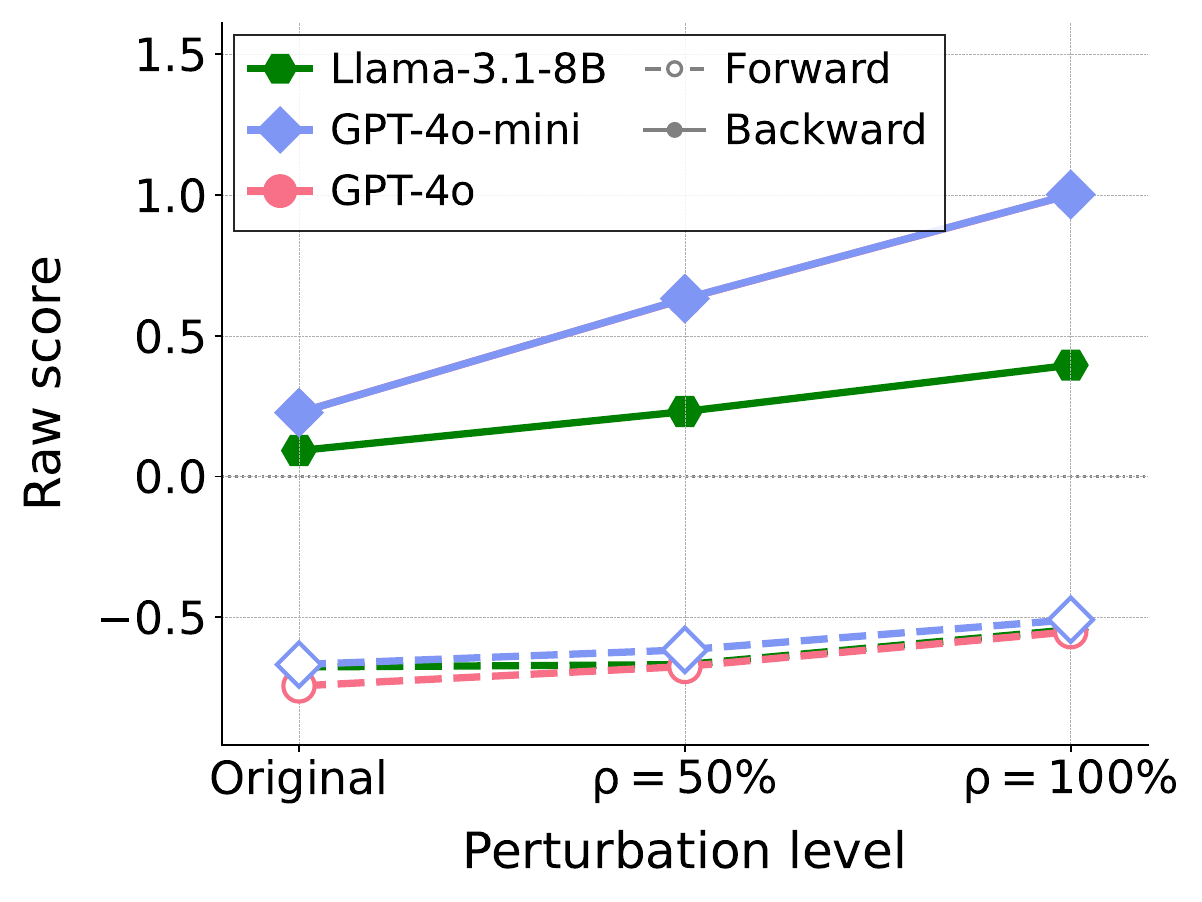}
  \caption{MATH500}
\end{subfigure}
\hfill
\begin{subfigure}[t]{0.49\columnwidth}
  \centering
  \includegraphics[width=\linewidth]{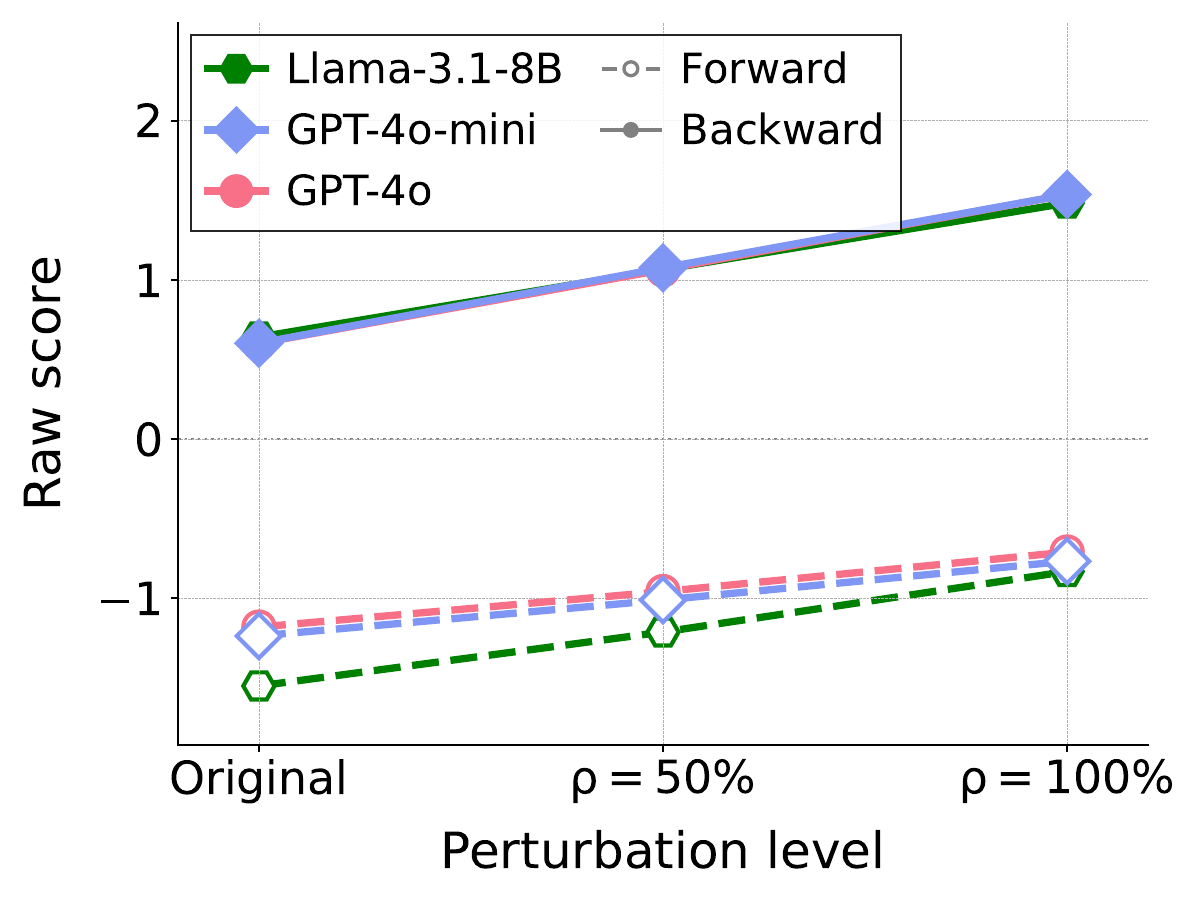}
  \caption{MuSR-ta}
\end{subfigure}
\caption{\textbf{Perturbation with demonstration fragments.} 
Few-shot fragments increase the raw backward term more sharply than the forward score.}
\label{fig:perturb}
\vspace{-0.1in}
\end{figure}
\begin{table}[t]
\centering
\caption{\textbf{Perturbation applied to response selection.} Selection accuracy when the first $\rho\%$ of the retrieved demonstration answer is prepended to a low-forward incorrect candidate.}
\label{tab:perturb_selection}
\begingroup
\begin{adjustbox}{width=\columnwidth}
\begin{tabular}{l l ccc}
\toprule
Estimator & Method & $\rho{=}0$ & $\rho{=}50$ & $\rho{=}100$ \\
\midrule
\multirow{2}{*}{LLaMA-3.1-8B} & Forward & 45.29 & 42.03 & 33.31 \\
& \name{} (Ours) & 45.29 & \textbf{47.48} & \textbf{47.48} \\
\midrule
\multirow{2}{*}{LLaMA-3.2-1B} & Forward & 45.41 & 41.38 & 30.71 \\
& \name{} (Ours) & 45.47 & \textbf{47.60} & \textbf{47.66} \\
\bottomrule
\end{tabular}
\end{adjustbox}
\endgroup
\vspace{-0.1in}
\end{table}
\begin{table}[t]
\centering
\caption{\textbf{Matched-pair analysis of the backward term.} Pair accuracy for the highest-forward correct and incorrect candidates in each query, grouped by which is more similar to the retrieved demonstration answer.}
\label{tab:matched_pair}
\begingroup
\begin{adjustbox}{width=\columnwidth}
\begin{tabular}{l cc}
\toprule
 Pair group & Forward & \name{} (Ours) \\
\midrule
Correct-more-similar ($n=316$) & \textbf{74.68} & 74.05 ($-0.63$) \\
Incorrect-more-similar ($n=314$) & 44.90 & \textbf{47.13} ($+2.23$) \\
\bottomrule
\end{tabular}
\end{adjustbox}
\endgroup
\vspace{-0.1in}
\end{table}

To further verify the backward term's sensitivity to demonstration artifacts, we conduct a perturbation experiment: we prepend the first $\rho\%$ of the few-shot examples to each candidate response and recalculate the forward and backward terms.
As shown in Figure \ref{fig:perturb}, inserting demonstration fragments causes a significant increase in the backward term, whereas the change in forward score is relatively small.
We next test how this sensitivity affects response selection by perturbing a \textit{low-forward incorrect} candidate. We compare the resulting selection accuracy of Forward and \name{} across three generators (GPT-4o, GPT-4o-mini, and LLaMA-3.1-8B) and three tasks (MATH500, GPQA, and MuSR-ta). In Table~\ref{tab:perturb_selection}, as $\rho$ increases, forward-only accuracy drops from 45.3\% to 33.3\%, whereas \name{} reaches 47.5\%, as the injected content incurs a larger backward penalty.

A high backward score can also reflect legitimate reuse of demonstration reasoning. However, this effect is asymmetric (Table~\ref{tab:matched_pair}): when the correct candidate is more demonstration-similar, the backward term reduces pairwise accuracy by only 0.6\%, as that candidate retains strong forward support, whereas it yields a 2.2\% gain when the incorrect candidate is more similar. Consistently, the raw backward score correlates with candidate--demonstration similarity, but not with correctness ($|r|\le0.09$) or response length ($|r|\le0.23$).
Together with the low agreement between forward and backward selections reported in Table~\ref{tab:agreement} (average agreement of 0.22), these results demonstrate that the backward term provides an orthogonal and mechanistically interpretable signal rather than redundant likelihood information, supporting its role as a backward reconstruction penalty.
Further details are provided in Appendix~\ref{app:backward_analysis}.

\begin{figure}[t]
    \centering
    \includegraphics[width=1.0\columnwidth]{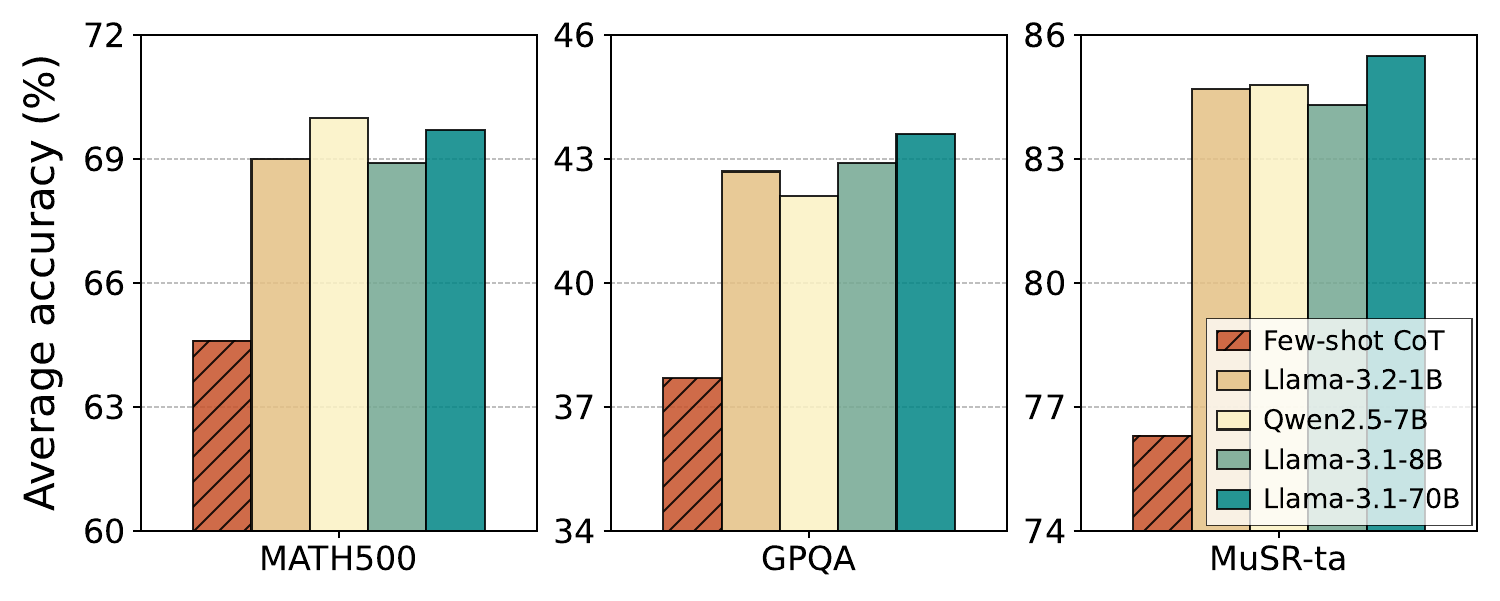}
    \vspace{-0.25in}
    \caption{\textbf{Estimation models.} Each bar shows the average accuracy of three generation LLMs.}
    \label{fig:estimation}
    \vspace{-0.1in}
\end{figure}

\begin{figure}[t]
    \centering
    \includegraphics[width=1.0\columnwidth]{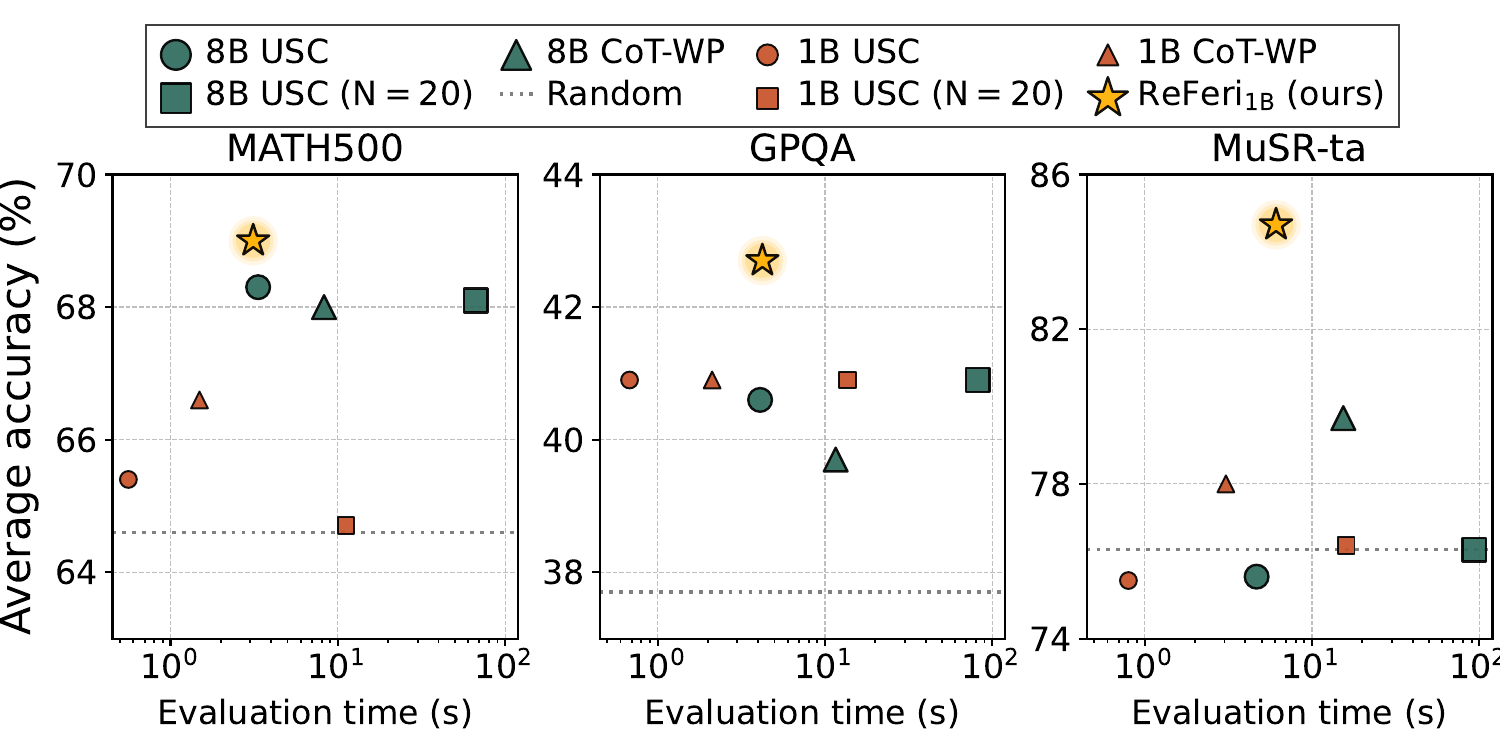}
    \vspace{-0.2in}
    \caption{\textbf{Cost--accuracy trade-off.} 
    Per-instance evaluation cost (seconds, single NVIDIA RTX A6000 GPU) against accuracy on selected benchmarks.
    }
    \label{fig:cost_main}
    \vspace{-0.1in}
\end{figure}

\paragraph{Estimation models.} 
To examine robustness to the choice of estimation model $p_{\theta}$, we evaluate \name{}'s performance using three LLMs of diverse scales and backbones: \texttt{LLaMA-3.2-1B-Inst}, \texttt{Qwen-2.5-7B-Inst}, and \texttt{LLaMA-3.1-70B-Inst}.  
The generation model is fixed (either GPT-4o-mini, GPT-4o, or LLaMA-3.1-8B), and the average accuracy is presented in Figure \ref{fig:estimation}.
Across all settings, \name{} consistently improves over Few-shot CoT while the best-performing estimator varies depending on the task; for example, Qwen-2.5-7B leads on MATH500, LLaMA-3.1-70B leads on the more challenging GPQA. 
This likely reflects differences in how well each model’s intrinsic knowledge aligns with the reasoning style of each benchmark.
Despite these task-specific differences, the overall improvements are consistent across all estimation models.   
This suggests that the effectiveness of \name{} primarily stems from its validation strategy with few-shot examples, rather than the specific choice of the estimator. 
More discussion is provided in Appendix~\ref{app:estimation_model}.

\paragraph{Robustness of \name{}.}
Since likelihood-based scoring can be sensitive to context, we examine \name{} under variations in few-shot examples, prompt format, and estimator calibration. Across independently sampled exemplar sets, \name{} consistently outperforms Few-shot CoT (Table~\ref{tab:few_shot_diff}), and the gains persist even when the exemplars are intentionally degraded in reasoning quality (Table~\ref{tab:weak_fewshot}). \name{} is also robust to prompt-format variations at both generation and verification (Table~\ref{tab:prompt_template}) and to estimator temperature, which produces negligible changes in performance (Table~\ref{tab:temperature}). Finally, we test the robustness of the nearest-neighbor approximation in: even the worst-performing fixed exemplar remains competitive, while increasing the number of retrieved neighbors leaves average performance nearly unchanged (Tables~\ref{tab:fixed_exemplar} and \ref{tab:topk_neighbors}). Together, these results show that \name{} does not rely on a particular exemplar set, prompt format, calibration choice, or retrieval configuration. Further analyses are provided in Appendix~\ref{app:few_shot}--\ref{app:nearest}.

\paragraph{Computational cost.} 
Next, we examine the trade-off between cost and accuracy.
Since all methods select from the same pre-generated pool of candidates, the generation cost is shared, so not considered. 
Figure \ref{fig:cost_main} plots accuracy against the average additional evaluation cost for response selection across the three generator models, where points closer to the upper-left indicate a better cost--accuracy trade-off.
Full results are provided in Table \ref{tab:cost}.
\name{} with a 1B estimator lies on a favorable trade-off frontier, reaching 69.0\% on MATH500, 42.7\% on GPQA, and 84.7\% on MuSR-ta, while running approximately $2.6\times$ faster than CoT-WP with the 8B estimator.
In contrast, 1B USC has the lowest computational cost, but its performance remains on par with random selection. 
Even when USC uses more candidates for voting, its performance actually degrades on benchmarks such as MATH500 (68.3\% to 68.1\%), despite requiring significantly more computation.
CoT-WP also drops significantly from 8B to 1B, showing that other likelihood-based baselines do not retain their effectiveness when scaled down.
Overall, combining \name{} with a small-scale estimator offers significant advantages and provides robust verification at a remarkably low computational cost.

\begin{table}[t]
\centering
\caption{\textbf{LLM personalization.} Evaluation results on LaMP-4 and LaMP-5 using GPT-4o as generator. 
}
\vspace{-0.05in}
\label{tab:personalization}
\begin{adjustbox}{width=\columnwidth}
\begin{tabular}{l cc|cc}
\toprule
\multirow{3}{*}{Methods} & \multicolumn{2}{c|}{LaMP-4} & \multicolumn{2}{c}{LaMP-5} \\
& Rouge-1 & Rouge-L & Rouge-1 & Rouge-L \\
\midrule
Vanilla        & 0.126 & 0.112 & 0.424 & 0.342 \\ 
Few-shot RAG   & 0.152 & 0.135 & 0.430 & 0.356 \\ 
USC & \underline{0.154} & 0.136 & 0.437 & 0.366 \\
CoT-WP & \underline{0.154} & \underline{0.137} & \underline{0.438} & \underline{0.368} \\
\midrule
\ccrow\name{}   & \textbf{0.157} & \textbf{0.140} & \textbf{0.451} & \textbf{0.387} \\ 
\bottomrule
\end{tabular}
\end{adjustbox}
\vspace{-0.1in}
\end{table}
\begin{figure}[t]
\centering
\begin{subfigure}[t]{0.49\columnwidth}
  \centering
  \includegraphics[width=\linewidth]{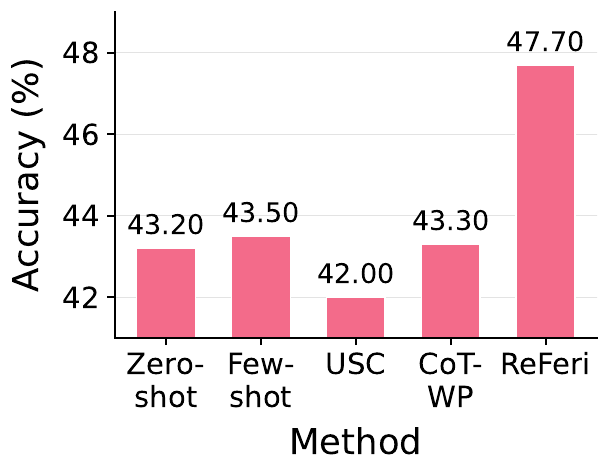}
  \caption{GPT-4o}
\end{subfigure}
\hfill
\begin{subfigure}[t]{0.49\columnwidth}
  \centering
  \includegraphics[width=\linewidth]{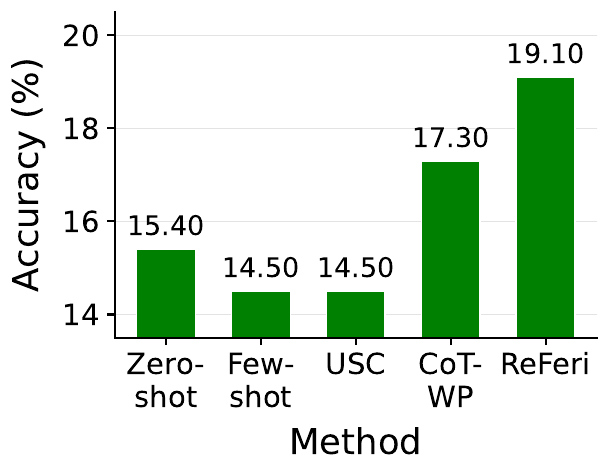}
  \caption{LLaMA-3.1-8B}
\end{subfigure}
\vspace{-0.05in}
\caption{\textbf{LiveCodeBench-code generation.} Evaluation results on LiveCodeBench code-generation.}
\label{fig:code}
\vspace{-0.15in}
\end{figure}

\paragraph{Generalization to open-ended tasks.}
We further apply \name{} to two open-ended generation tasks, \textit{LLM personalization} and \textit{code generation}, to evaluate its broader applicability beyond reasoning benchmarks, where majority voting and answer-extraction approaches are inherently limited.
Implementation details are provided in Appendix \ref{app:implementation}.
First, on the \textit{LaMP} benchmark \citep{salemi2024lamp}, which aims to steer LLM responses towards individual users, we apply \name{}\footnote{Since personalization relies heavily on retrieved examples, we use the full backward term for rigorous evaluation.} to LaMP-4 (news headline generation) and LaMP-5 (scholarly title generation) using GPT-4o as the generator.
We compare against \textit{Vanilla} (no user context) and \textit{Few-shot RAG}, which augments the prompt with $K{=}5$ examples retrieved via BM25 \citep{robertson2009probabilistic} from the user's history.
Following \citep{salemi2024lamp}, we evaluate all responses against gold references using ROUGE-1 and ROUGE-L. 
As shown in Table \ref{tab:personalization}, \name{} consistently outperforms all baselines; CoT-WP offers only limited gains on LaMP-4, where short outputs render confidence-based signals uninformative.
Next, on \textit{LiveCodeBench} \citep{jain2025livecodebench}, which requires both algorithmic correctness and executability, we use GPT-4o and LLaMA-3.1-8B as generation models.
As shown in Figure \ref{fig:code}, \name{} achieves the best accuracy: 
USC is no better than random selection, while CoT-WP struggles especially on GPT-4o, suggesting a calibration mismatch in likelihood-based selection.
\section{Related Works}\label{sec:2}

\paragraph{Few-shot in-context learning of LLM.}\label{sec:2.1}
Few-shot in-context learning (ICL) enables LLMs to generalize to unseen tasks from a handful of demonstrations \citep{brown2020language}. 
For complex reasoning, Chain-of-Thought (CoT) prompting appends intermediate steps to the few-shot examples
\citep{wei2022chain, fu2023complexitybased, jin2024impact}. 
By retrieving better examples, a method has also emerged to further enhance ICL \citep{wu2023selfadaptiveincontextlearninginformation, peng-etal-2024-revisiting}.
However, ICL does not always guarantee improvements \citep{min-etal-2022-rethinking}, and the performance gap between zero-shot and few-shot CoT is narrowing as instruction tuning becomes more effective \citep{sprague2025cot}. 
Nonetheless, carefully selected demonstrations are still effective \citep{huang2024fewer}, reducing overconfidence in multi-step reasoning  \citep{ge2025innate}, and mitigating hallucinations and memory-based mistakes in complex tasks \citep{yan2025recitation}.

\paragraph{Selection of diverse LLM outputs.} 
Due to the probabilistic nature of LLM decoding, a variety of outputs are generated, each reflecting different reasoning paths \citep{kadavath2022language, qiu2024semantic}. 
Best-of-$N$ selection aims to identify the most promising candidate among them, but existing methods face several limitations.
Self-consistency \citep{wang2023self} selects the majority answer, assuming a well-formatted answer that is often violated in open-ended tasks (\textit{e.g.}, summarization or dialogue) \citep{stiennon2020learning, salemi2024lamp}.
Outcome \citep{cobbe2021training, uesato2022solving}, or Process Reward Models \citep{lightman2024let, wang2024math} require domain-specific checkers, limiting scalability to new domains. 
To eliminate external verifiers, alternatives that require no training leverage the model's internal signals (\textit{e.g.}, LLM-as-Judge, Self-Certainty) \citep{chen2023universal, zheng2023judging, kang2025scalable, wang2024chain}, but often lead to miscalibration in unfamiliar domains.
In particular, LLM-as-Judge requires fine-tuning to be reliable, reintroducing supervision \citep{yuan2024self, mahan2024generative, zhang2025generative}.
In contrast, \name{} is training-free and task-agnostic, offering a more scalable and generalizable approach by recycling few-shot examples for verification. 

\section{Conclusion}\label{app:conclusion}
We propose \name{}, a training-free framework to find promising LLM output.
By reusing few-shot data not only for generation but also for verification, \name{} mitigates confidence miscalibration.
In experiments, \name{} is consistently effective in various LLMs and tasks, while remaining effective with lightweight estimators.
Overall, these results suggest that \name{} is a practical way to find reliable LLM output with minimal human involvement, opening future directions to reconsider the broader utility of few-shot examples.

\newpage 

\section*{Limitations}
\name{} assumes access to few-shot demonstrations. 
While their choice and quality can affect the candidate pool, \name{} remains effective across different and deliberately weakened exemplar sets (Appendix~\ref{app:few_shot}), suggesting that carefully curated demonstrations are not essential.
Proxy examples from retrieval or stronger models may therefore suffice, though settings with no available demonstrations remain an open direction.

In addition, since the selection by \name{} is determined by likelihoods produced by an estimation model, it is limited to explain why a response is incorrect, unlike PRMs, which offer step-level feedback, or LLM-as-judge, which can easily generate explanations by prompting. 
However, we believe that \name{} can potentially provide a certain level of interpretability; for example, the token-level uncertainty signals derived from the backward-term contributions of candidate responses could potentially reveal untrustworthy tokens. 
Future work could develop more systematic token-level or step-level diagnostic tools from these decomposed signals, allowing \name{} to not only select reliable responses but also explain the source of unreliability.

\section*{Broader Impact and Ethical Implications}
\name{} provides a training-free method for selecting promising outputs from LLMs. 
This makes it particularly valuable in scenarios where labeled data is scarce or where model fine-tuning is impractical such as limited access to data, or applications in emerging domains where predefined labels are unavailable. 
In addition, \name{} reduces the barrier to deploying LLMs in real-world settings without additional supervision. This may contribute to broader and more efficient adoption of LLMs in resource-constrained environments. 
All datasets used are public and widely adopted.

\section*{Acknowledgments}

Dongseok Lee and Jaehyung Kim are affiliated with the Department of Artificial Intelligence at Yonsei University.
This research was supported in part by Institute for Information \& communications Technology Planning \& Evaluation (IITP) grant funded by the Korea government (MSIT) (No. RS-2020-II201361, Artificial Intelligence Graduate School Program (Yonsei University); No. RS-2025-25442405, Development of a Self-Learning World Model-Based AGI System for Hyperspectral Imaging).
\bibliography{custom}

\appendix
\newpage
\section{More Details of Experimental Setups}\label{app:more_detail}
This section covers more details about the experiments from Section \ref{sec:4}.

\subsection{Datasets}\label{app:dataset}
This subsection provides more information about the dataset and the few-shot examples we used. 

\begin{itemize}[leftmargin=*]

\item \textbf{MATH500.}  
The MATH benchmark \citep{hendrycks2021measuring} consists of 12,500 LaTeX-formatted competition-level math problems, with topics ranging from algebra and geometry to number theory. Each problem includes a step-by-step solution and expects the model to generate a boxed final answer (e.g., an integer or simplified expression). We use MATH500, a 500-question subset introduced in \citep{lightman2024let}, uniformly sampled from the test split to preserve subject and difficulty distribution. For few-shot examples, we follow 5-shot examples from \citep{yang2024qwen2}\footnote{\url{https://github.com/QwenLM/Qwen2.5-Math}} for GPT-based models and 4-shot examples from \citep{lewkowycz2022solving}\footnote{\url{https://huggingface.co/datasets/meta-llama/Llama-3.2-3B-Instruct-evals}} for LLaMA-based models. The reason for this choice is based on our empirical observation: Simply adding "Please think step by step and put your final answer within \verb|\boxed{}|." as done in GPT-style few-shot prompts led to a significant drop in accuracy. 
Namely, LLaMA-based models require prompt formats that are aligned with their own instructions and are sensitive to deviations from the learned template.
This benchmark evaluates symbolic reasoning ability in mathematical domains.

\item \textbf{MMLU-Pro.}
MMLU-Pro \citep{wang2024mmlu} is an extension of the original MMLU benchmark \citep{hendrycks2020measuring}, which evaluates broad knowledge and reasoning over 57 subjects using 14k 4-way multiple-choice questions. MMLU-Pro introduces 12k curated 10-way multiple-choice questions across 14 professional domains, increasing task difficulty and emphasizing complex, multi-step reasoning. Instead of using the full test set, we subsample 300 questions per subject (totaling 4,200) using random seed 42 and we will share the used indices at the code. For few-shot examples, we used 5-shot examples following the format used in \citep{wang2024mmlu}. This benchmark is used to assess domain-specific and robust reasoning performance.

\item \textbf{GPQA.}
GPQA \citep{rein2024gpqa} is a graduate-level QA benchmark consisting of 448 expert-authored multiple-choice questions in domains such as physics, chemistry, and biology. Designed to be "Google-proof," it focuses on evaluating complex scientific reasoning that cannot be answered through simple retrieval. We evaluate on GPQA-Diamond, a curated subset of 198 especially difficult questions selected by the authors. Few-shot examples are taken directly from the official 5-shot examples used in \citep{rein2024gpqa}. This task measures deep scientific understanding.

\item \textbf{DROP.}
The DROP benchmark \citep{dua2019drop} contains 96k question-answer pairs requiring discrete reasoning over Wikipedia passages (e.g., numerical operations, counting, or date comparison). Answers may include spans, numbers, or dates. We evaluate on a 500-sample subset randomly selected from the dev set, and we will share the selected indices at the code. We use 3-shot examples from \citep{zhou2022least} and report both EM and F1 metrics following the official implementation. This benchmark evaluates models' symbolic reasoning grounded in natural language passages.

\item \textbf{HotpotQA.}
HotpotQA \citep{yang2018hotpotqa} consists of 113k multi-hop QA pairs requiring reasoning over multiple Wikipedia documents. The model must retrieve at least two relevant passages and combine facts to answer each question. We follow the \citep{kim2024sure}, which uses 500 samples from the dev set. Few-shot examples are taken from \citep{yao2023react} using the same 6-shot examples. This task tests compositional reasoning and the ability to aggregate distributed information across documents.

\item \textbf{MuSR.}
MuSR \citep{sprague2310musr} is a benchmark for multi-step reasoning over long-form narratives (800–1000 words), constructed via neuro-symbolic generation to embed logical dependencies into natural language. It includes structured tasks such as TeamAllocation (constraint-based planning) and ObjectPlacement (spatial consistency reasoning). We evaluate on the 250 TeamAllocation and 256 ObjectPlacement examples from the official release \citep{sprague2310musr}, using 3-shot examples tailored to each task \citep{sprague2025cot}. MuSR requires understanding of narrative flow, contextual logic, and physical feasibility. 
Notably, prior work \citep{sprague2025cot} has shown that few-shot Chain-of-Thought (CoT) prompting yields significant gains over zero-shot CoT in MuSR, highlighting the role of in-context examples in complex reasoning.

\item \textbf{LaMP.}
The LaMP benchmark \citep{salemi2024lamp} evaluates personalized text generation across multiple tasks. We focus on LaMP-4 (personalized news headline generation) and LaMP-5 (personalized scholarly title generation). In LaMP-4, the model generates headlines conditioned on article content and author profile, while LaMP-5 requires generating research paper titles conditioned on researcher profiles. We follow the official test splits and evaluate with ROUGE-L and ROUGE-1. Few-shot RAG is an example of 5 shots retrieved from user history via BM25. These tasks test personalization and style-aware generation in open-ended outputs.

\item \textbf{LiveCodeBench.} 
LiveCodeBench \citep{jain2025livecodebench} is an execution-based benchmark for evaluating LLM code-generation ability. 
We use the \texttt{code\_generation\_lite} test split with \texttt{release\_v3}. 
Each instance provides a programming problem specification and, for some problems, starter code. 
The model must generate an executable Python solution, which is evaluated by running the generated code against the corresponding test cases. 
This benchmark measures both algorithmic correctness and executability, making it suitable for evaluating open-ended code generation.

\end{itemize}

\subsection{Baselines}\label{app:baselines}
This subsection provides additional details on the baselines used in our experiments.
We further include the prompt template used for the baselines, with MATH500 serving as a representative benchmark, in Listings \ref{lst:base_begin}--\ref{lst:base_end}.

\begin{itemize}[leftmargin=*]

\item \textbf{Zero-shot CoT.}
Zero-shot Chain-of-Thought \citep{kojima2022large} is a method that induces step-by-step reasoning in a model by adding a simple reasoning prompt to the input query without requiring separate in-context examples. Typically, a phrase such as “Let's think step by step” is appended to the query to ensure that the model generates intermediate reasoning steps rather than outputting only the final answer.
However, since recent instruction-tuned models differ in their preferred prompt formats and instruction-following behaviors, the effectiveness of the same trigger phrase may vary depending on the model.
In our experiments, Zero-shot CoT is used as the baseline to evaluate how well models can generate reasoning processes and arrive at the correct answer when provided only with reasoning instructions and no additional examples.

\item \textbf{Few-shot CoT.}
Few-shot Chain-of-Thought \citep{brown2020language} is an approach that provides several CoT demonstrations alongside a test query to guide the model to follow the reasoning and response formats of the examples.
Each demonstration consists of a question, a step-by-step reasoning process, and a final answer.
Rather than simply adding examples, the prompt is constructed by combining the reasoning instructions used in Zero-shot CoT with few-shot CoT examples.
In other words, the model first learns the problem-solving approach and output format from the provided CoT examples through in-context learning, and then generates a reasoning path and final answer for the test query.
Therefore, Few-shot CoT serves as a baseline for evaluating how well a model can generalize reasoning patterns to new problems using only a small number of reasoning examples and can be viewed as a random selection method. 

\item \textbf{LEAP.}
Learning Principles (LEAP) \citep{zhang2024context} is a method that intentionally induces errors in few-shot examples and then extracts task-specific principles through the model’s self-reflection to improve prompting.
Specifically, the model first generates multiple reasoning outputs for the few-shot examples and uses the outputs that do not match the correct answer as mistake cases.
It then extracts example-specific low-level principles from each mistake case and generalizes them to generate high-level principles applicable to unseen questions.
In our experiments, following LEAP’s default settings, we set temperature to $1.0$ to generate $N=20$ reasoning outputs and add the extracted high-level principles to the few-shot CoT prompt to generate the final answer.
This procedure does not require human-written annotations, since the principles are generated by the model itself. This baseline evaluates whether explicit task-level principles can improve reasoning beyond standard few-shot demonstrations.

\item \textbf{USC.}
Universal Self-Consistency \citep{chen2023universal} is a prompt-based selection method that requires no additional training and enables the model to select a consistent answer from among multiple Chain-of-Thought outputs.
Standard Self-consistency \citep{wang2023self} methods select the most frequent answer from sampled inference paths, implicitly assuming that the majority opinion is more likely to be accurate.
However, this approach assumes that the correct answer can be easily extracted and aggregated, and it cannot be applied to open-ended tasks where this is not possible. 
To address this limitation, USC utilizes the model itself as an evaluator. Specifically, it presents the model with multiple candidate responses and asks it to select the most reliable one.

\item \textbf{CoT-WP.}
Chain-of-Thought Reasoning without Prompting (CoT-WP) \citep{wang2024chain} is a method that scores each candidate response using the token-level probabilities of a language model conditioned on the same few-shot prompt.
Specifically, it calculates the probability gap between the top-1 token and the top-2 token at each answer token position that makes up the final answer.
When an answer consists of multiple tokens, this experiment calculates the confidence gap for every answer token position and averages these values to obtain the final score for the candidate response.
A higher value indicates that the model is more confident in the selected answer token than in the alternative tokens.
Finally, the candidate response with the highest CoT-WP score is selected.
In our experiments, CoT-WP serves as a baseline closely related to \name{} because both methods require no training and rely on the model’s intrinsic likelihood signals to evaluate candidate responses.

\item \textbf{Self-Certainty.}
Self-Certainty \citep{kang2025scalable} is a method that estimates the reliability of a generated response using the model’s next-token distribution.
Specifically, it calculates the KL divergence between the model’s next-token distribution and a uniform distribution.
The closer the distribution is to a uniform distribution, the more the model’s prediction is spread across multiple tokens, indicating low certainty; conversely, the more it is concentrated on a specific token, the higher the certainty.
In this experiment, we use this certainty score to evaluate the reliability of candidate responses.
Therefore, Self-Certainty is used as a baseline to assess whether the uncertainty signal derived from the model’s internal predictive distribution can identify more reliable generations.

\end{itemize}

\subsection{Implementation}\label{app:implementation}
This section provides the detailed information needed to implement the main experiment.

\paragraph{Resource details.}
To avoid out-of-memory errors, we used two NVIDIA H100 GPUs for evaluation with the LLaMA-3.1-70B-Instruct model. All other experiments were performed on a single NVIDIA RTX A6000 GPU. 

\paragraph{Response generation.}
We use lm-eval-harness\footnote{\url{https://github.com/EleutherAI/lm-evaluation-harness}} to generate responses from LLaMA-based models, with temperature set to 1.0 and 5 responses per input. 
The prompt\footnote{\url{https://huggingface.co/datasets/meta-LLaMA/LLaMA-3.1-8B-Instruct-evals}} was written in chat template format using vllm.
For GPT-family models, we use the official OpenAI API to generate completions under the same sampling configuration. The remaining settings follow the GPT API default settings.
During evaluation, for Zero-shot CoT, Few-shot CoT, and LEAP, we average performance over five sampled generations, which can be viewed as random selection from the candidate pool.
In contrast, for USC, CoT-WP, Self-Certainty, and \name{}, we evaluate the single response selected from the same candidate pool.
All evaluations are conducted using our custom evaluation scripts to ensure consistent scoring and formatting across models.

\paragraph{Algorithm of \name.} In algorithm \ref{alg:main}, we present the formal algorithm for \name{}. We generate multiple candidate responses $\{r_1,...,r_N\}$ for each test query using Few-shot CoT, as it exhibits better quality on the average (see Table \ref{tab:main}).
\begin{algorithm}[!ht]
   \caption{\name{} algorithm}
   \label{alg:main}
\begin{algorithmic}
    \State \textbf{Input:} estimation model $P$, embedding model $E$, test-query $\widetilde{q}$, $N$ candidate responses $\{r_1, \dots, r_N\}$, $K$ few-shot examples $\mathbf{X}=\{\mathbf{x}_i\}^K_{i=1}$
    \vspace{0.05in} 
    \hrule
    \vspace{0.05in}  
  
    \State $i^\dagger = \arg\max_{i=1,\dots,K}
    \cos\big(E(\widetilde{q}), E(q_i)\big)$

    \For{$n=1$ {\bfseries to} $N$}
        \State $S_{\tt {Forw}} \leftarrow$ Compute forward score with $r_n$ as label, using $P$ and context $(\widetilde{q}, \mathbf{X})$ (Eq. \ref{eq.forward})
        \State Construct $\widetilde{\mathbf{X}}_{i^\dagger}\leftarrow$ using a leave-one-out strategy (Eq. \ref{eq.replace})

        \State $S_{\tt {Back}} \leftarrow $ backward score with $a_{i^\dagger}$ as label, using $P$ and $\widetilde{\mathbf{X}}_{i^\dagger}$ (Eq. \ref{eq.new_backward})

        \State $S_{\tt {Fin}} \leftarrow S_{\tt {Forw}} - S_{\tt {Back}}$ (Eq. \ref{eq.combined})
        \State $S_{n} \leftarrow S_{\tt {Fin}}$
  \EndFor
  \State $r_{n^*} \leftarrow \arg\max_n S_n$ 
  \State \textbf{return}~~$r_{n^*}$
\end{algorithmic}
\end{algorithm}

\paragraph{Application to LLM personalization.}
In the personalization experiments, \name{} is applied to two open-ended generative LaMP tasks: personalized news headline generation (LaMP-4) and personalized academic paper title generation (LaMP-5).
For each task, we use a 100-user subset, with 6,725 test queries for news headline generation and 106 test queries for academic title generation.
For each query, we use BM25 to search for relevant examples from the user’s history and construct a few-shot personalized prompt using $K=5$ of the retrieved examples.
We use the full backward term rather than the nearest-neighbor approximation for this experiment, since personalization relies heavily on the retrieved user examples.
Additionally, a pre-built natural language user profile is appended to the prompt for each user. 
GPT-4o is used as the generator, with the temperature set to 1.0, and 5 candidate outputs are sampled for each query.
The maximum generation length is limited to 128 tokens.
As in Table \ref{tab:main}, Llama-3.1-8B-Instruct is used as the likelihood estimator.
Following the LaMP evaluation protocol, the generated headlines and titles are evaluated using the ROUGE score by comparing them to the reference output.

\paragraph{Application to LiveCodeBench.}
The LiveCodeBench experiment uses 612 programming problems.
We follow the official LiveCodeBench code generation prompt and evaluation format.
For each problem, we configure a chat-style prompt that contains the problem statement. Once the starter code is provided, it will be included in the prompt.
Depending on the official prompt format, we use different generative demonstrations for the function completion problem and the standard input problem.
Each model is instructed to generate the Python code contained in the code block.
We set the temperature at $1.0$ and sampled $N=5$ candidate programs per problem.
The selected response is considered correct only when it passes both public and hidden test cases.
Thus, all selection methods operate on the same set of candidate programs, and test-case execution is used only for final evaluation.

\section{More Quantitative Results}\label{app:more_exp}
In this section, we use the candidate responses generated in the first run of Table~\ref{tab:main} unless otherwise specified. 

\subsection{Significance tests over matched runs}\label{sec:significance}
\begin{table}[t]
\centering
\caption{\textbf{Paired significance tests against the strongest baselines.} Two-sided paired $t$-tests on the seven-task average accuracy over the three matched runs of Table~\ref{tab:main}.}
\label{tab:significance}
\begingroup
\small
\setlength{\tabcolsep}{3.5pt}
\resizebox{\columnwidth}{!}{
\begin{tabular}{@{}llccc@{}}
\toprule
Generator & Comparison & $\Delta$ & 95\% CI & $p$ \\
\midrule
\multirow{2}{*}{GPT-4o}
 & vs.\ CoT-WP         & $+1.89$ & $[0.87, 2.90]$ & $.0153$ \\
 & vs.\ Self-Certainty & $+1.86$ & $[0.66, 3.06]$ & $.0218$ \\
\midrule
\multirow{2}{*}{GPT-4o-mini}
 & vs.\ CoT-WP         & $+2.08$ & $[1.22, 2.94]$ & $.0091$ \\
 & vs.\ Self-Certainty & $+1.71$ & $[0.36, 3.07]$ & $.0322$ \\
\midrule
\multirow{2}{*}{LLaMA-3.1-8B}
 & vs.\ CoT-WP         & $+1.90$ & $[0.33, 3.47]$ & $.0349$ \\
 & vs.\ Self-Certainty & $+1.23$ & $[-0.06, 2.53]$ & $.0547$ \\
\bottomrule
\end{tabular}}
\endgroup
\end{table}

\begin{table}[t]
\centering
\caption{\textbf{Paired significance tests for the scoring components.} Two-sided paired $t$-tests of \name{} against each single-term variant, on the matched aggregate over the three runs of Table~\ref{tab:ablation_full}.}
\label{tab:significance_ablation}
\begingroup
\small
\setlength{\tabcolsep}{4pt}
\begin{tabular}{@{}lcc@{}}
\toprule
Comparison & 95\% CI & $p$ \\
\midrule
\name{} vs.\ Forward       & $[0.021, 0.155]$ & $.0302$ \\
\name{} vs.\ Backward      & $[3.250, 5.859]$ & $.0044$ \\
\name{} vs.\ $-$Backward   & $[3.796, 6.826]$ & $.0044$ \\
\bottomrule
\end{tabular}
\endgroup
\vspace{-0.1in}
\end{table}

To evaluate whether the observed performance improvements are robust to variability across runs, we perform two-sided paired $t$-tests on three matched generative runs. 
For each run, methods are evaluated on the same candidate pool, and we compare their seven-task average accuracy. 
Table \ref{tab:significance} shows the comparison results with CoT-WP and Self-Certainty for all three generators. 
\name{} outperforms the others in all six comparisons; five of these are statistically significant at the $p<.05$ level, while the comparison with Self-Certainty on LLaMA-3.1-8B yields a $p{=}.0547$ result. 
We apply the same procedure to the scoring-related ablation experiments reported in Table~\ref{tab:ablation_full}.
As reported in Table~\ref{tab:significance_ablation}, the overall \name{} score outperforms both the forward-only variant ($p{=}.0302$) and the two backward-only variants ($p{=}.0044$).

\subsection{Additional comparison with few-shot prompting-based selection methods}\label{app:prompt_based_result}
\begin{table*}[t]
\caption{\textbf{Comparison with prompting-based selection.}  Overall performance on seven reasoning benchmarks comparing the proposed {\name{}} with different prompting-based baselines not require additional training, under three different state-of-the-art LLMs.
{For reference, we additionally include the Oracle upper bound.}}
\label{tab:prompt_based_result}
\small
\centering
\resizebox{\textwidth}{!}{
\begin{tabular}{l l ccccccc|c}
\toprule
Models & Methods & MuSR-ta & MuSR-op & GPQA & MATH500 & DROP & HotpotQA & MMLU-PRO & Avg. \\
\midrule
\multirow{5}{*}{GPT-4o}
 & USC            & 85.2 & \underline{71.1} & \underline{47.0} & \underline{77.4} & 82.2 / 90.2 & \underline{45.6} / 59.7 & \underline{74.5} & \underline{69.0} \\
 & USC-w/Fewshot  & \underline{88.8} & 69.1 & 46.0 & \underline{77.4} & 82.0 / 89.9 & 45.4 / 60.1 & 74.1 & \underline{69.0} \\
 & LLM-as-Judge   & 86.0 & 68.0 & 46.5 & \textbf{77.8} & \underline{82.8 / 91.0} & \underline{45.6 / 59.8} & 73.3 & 68.6 \\
 & \cc\name{}  & \cc\textbf{90.4} & \cc\textbf{71.9} & \cc\textbf{51.5} & \cc\textbf{77.8} & \cc\textbf{83.6 / 91.1} & \cc\textbf{47.0 / 60.7} & \cc\textbf{75.4} & \cc\textbf{71.1} \\ \cmidrule(lr){2-10}
 & Oracle & 96.4	& 87.9 &72.2 & 86.6&{89.2 / 94.7} &{55.4 / 69.2} & 84.1 & 81.7 \\ 
\midrule
\multirow{5}{*}{GPT-4o-mini}
 & USC            & 74.4 & 60.9 & \textbf{46.0} & 77.8 & 76.8 / 83.8 & \underline{35.0 / 47.2} & \underline{63.7} & \underline{62.1} \\
 & USC-w/Fewshot  & \underline{76.4} & \textbf{63.3} & 39.9 & \textbf{78.2} & 77.2 / 84.0 & 34.8 / 46.6 & 63.2 & 61.9 \\
 & LLM-as-Judge   & 75.6 & 60.6 & 34.3 & 77.0 & \underline{77.4 / 84.4} & \underline{35.0} / 46.7 & 63.3 & 60.5 \\
 & \cc\name{}  & \cc\textbf{82.8} & \cc\underline{61.3} & \cc\underline{41.9} & \cc\underline{77.8} & \cc\textbf{79.2 / 84.9} & \cc\textbf{36.2 / 48.0} & \cc\textbf{64.9} & \cc\textbf{63.4} \\ \cmidrule(lr){2-10}
 & Oracle & 97.2	& 78.1 &70.7 & 85.8&{86.6 / 91.6} &{43.4 / 56.4} & 76.8 & 76.9 \\ 
\midrule
\multirow{5}{*}{LLaMA-3.1-8B}
 & USC            & 67.2 & 52.3 & \underline{28.8} & \underline{49.6} & \textbf{69.6 / 75.8} & 24.4 / 32.5 & \textbf{45.6} & 48.2 \\
 & USC-w/Fewshot  & \underline{70.0} & 53.9 & 28.3 & 47.8 & 69.0 / 75.3 & \textbf{25.2} / \underline{32.3} & \underline{45.1} & \underline{48.5} \\
 & LLM-as-Judge   & 65.2 & \underline{55.1} & 21.2 & 46.0 & 67.7 / 74.0 & 23.4 / 31.2 & 44.1 & 46.1 \\
 & \cc\name{}        & \cc\textbf{79.6} & \cc\textbf{57.8} & \cc\textbf{35.4} & \cc\textbf{51.2} & \cc\underline{69.4 / 75.7} & \cc\underline{25.0} / \textbf{33.2} & \cc\underline{45.1} & \cc\textbf{51.9} \\ \cmidrule(lr){2-10}
 & Oracle & 97.6	& 88.3 &59.6 & 66.6&{83.4 / 88.8} &{33.8 / 45.0} & 70.8 & 71.4 \\ 
\bottomrule
\end{tabular}
}
\end{table*}

\begin{table*}[t]
\vspace{0.1in}
\centering
\caption{\textbf{Response selection distribution per task (GPT-4o-mini).} 
The selections distribution suggests that prompt-based selection method may exhibit position bias rather than relying solely on response quality.
}
\label{tab:taskwise_distribution}
\small
{
\begin{tabular}{ll|cccccc}
\toprule
{Task} & {Method} & {\#1} & {\#2} & {\#3} & {\#4} & {\#5} & {Fail (-1)} \\
\midrule
\multirow{3}{*}{MATH500}
& USC & 90.2 & 3.8 & 1.0 & 2.8 & 2.2 & 0.0 \\
& USC-w/ Fewshot & 81.0 & 12.6 & 0.6 & 3.2 & 2.6 & 0.0 \\
& LLM-as-Judge & 51.4 & 2.2 & 8.0 & 12.2 & 25.4 & 0.8 \\
\midrule
\multirow{3}{*}{MMLU-Pro}
& USC & 34.2 & 19.1 & 7.2 & 24.0 & 15.5 & 0.0 \\
& USC-w/ Fewshot & 18.8 & 31.0 & 7.8 & 22.7 & 19.7 & 0.0 \\
& LLM-as-Judge & 22.1 & 8.0 & 7.8 & 13.5 & 48.3 & 1.2 \\
\midrule
\multirow{3}{*}{GPQA}
& USC & 21.7 & 15.2 & 13.1 & 23.7 & 26.3 & 0.0 \\
& USC-w/ Fewshot & 19.7 & 17.7 & 10.6 & 23.7 & 28.3 & 0.0 \\
& LLM-as-Judge & 30.8 & 9.1 & 10.1 & 7.6 & 42.4 & 0.0 \\
\midrule
\multirow{3}{*}{DROP}
& USC & 73.8 & 21.8 & 1.8 & 1.2 & 1.4 & 0.0 \\
& USC-w/ Fewshot & 78.2 & 16.6 & 3.0 & 1.0 & 1.2 & 0.0 \\
& LLM-as-Judge & 68.8 & 9.6 & 4.6 & 4.4 & 12.4 & 0.2 \\
\midrule
\multirow{3}{*}{HotpotQA}
& USC & 77.0 & 13.4 & 2.8 & 5.4 & 1.4 & 0.0 \\
& USC-w/ Fewshot & 68.6 & 20.6 & 2.8 & 6.0 & 2.0 & 0.0 \\
& LLM-as-Judge & 65.0 & 15.0 & 4.8 & 6.0 & 9.2 & 0.0 \\
\midrule
\multirow{3}{*}{MuSR-op}
& USC & 51.6 & 18.0 & 11.7 & 7.4 & 11.3 & 0.0 \\
& USC-w/ Fewshot & 36.7 & 40.2 & 10.2 & 6.2 & 6.6 & 0.0 \\
& LLM-as-Judge & 26.9 & 21.5 & 15.2 & 20.7 & 15.2 & 0.4 \\
\midrule
\multirow{3}{*}{MuSR-ta}
& USC & 34.0 & 3.2 & 0.8 & 9.6 & 9.6 & 42.8 \\
& USC-w/ Fewshot & 46.4 & 35.2 & 2.4 & 3.6 & 12.4 & 0.0 \\
& LLM-as-Judge & 27.6 & 2.8 & 0.4 & 1.2 & 50.4 & 17.6 \\ 
\bottomrule
\end{tabular}
}
\vspace{0.1in}
\end{table*}

\begin{table*}[t]
\centering
\caption{\textbf{Evaluation of USC ordering with GPT-4o-mini.}
We evaluate USC under the default candidate order and two random permutations (\emph{perm-A}, \emph{perm-B}).
The variation across permutations shows that prompt-based judgment can be sensitive to presentation order.}
\label{tab:usc_ordering}
\small
\begin{tabular}{lcccccccc}
\toprule
{Methods} & {MuSR-ta} & {MuSR-op} & {GPQA} & {MATH500} & {DROP} & {HotpotQA} & {MMLU-PRO} & {Avg.} \\
\midrule
USC (perm-A)   & 75.6 & 59.8 & 41.9 & 78.0 & 76.6 / 83.5 & 35.6 / 47.2 & 63.2 & 61.5 \\
USC (perm-B)   & 77.2 & 56.6 & 45.0 & 77.2 & 76.6 / 83.4 & 35.2 / 46.9 & 63.6 & 61.6 \\
USC (default)  & 74.4 & 60.9 & 46.0 & 77.8 & 76.8 / 83.8 & 35.0 / 47.2 & 63.7 & 62.1 \\
\bottomrule
\end{tabular}%
\vspace{0.1in}
\end{table*}

Among the multiple answer selection methods, the simplest and most accessible approach (\textit{e.g.}, learning overhead, domain specificity, etc.) is arguably LLM-as-Judge \cite{chen2023universal, zheng2023judging}. 
It uses the LLM itself to score and select answers via in-context learning without any additional training or external verifiers. 
In particular, the addition of few-shot examples to LLM-as-Judge might be most closely aligned with the core motivation of \name{}, which is to use demonstrations not only for generation but also for validation. 
Therefore, in this section, we compare \name{} and (1) the original \textit{USC} \citep{chen2023universal}, (2) \textit{USC with few-shot} (our adaptation), and (3) \textit{LLM-as-Judge with few-shot} created with our optimized prompt (see list \ref{lst:usc_with_fewshot} and \ref{lst:llm-as-judge}). 
For a fair comparison and to maintain consistency with Table~\ref{tab:main}, LLaMA-3.1-8B-Instruct is employed as the judging model for all LLM-as-Judge variants.

As shown in Table \ref{tab:prompt_based_result}, \name{} consistently achieves the best or second-best accuracy across all LLMs and benchmarks. Interestingly, we observe that adding few-shot demonstrations to USC often degrades performance (\textit{e.g.}, on GPQA and DROP with GPT-4o-mini and LLaMA-3.1-8B), which is likely due to the sensitivity of LLMs to prompt format and positional bias of the responses. 

Notably, we observe that both prompt-based selection methods, USC and LLM-as-Judge, are highly sensitive to the order of candidate responses. 
In our experiments, USC frequently selected one of the first two responses regardless of correctness; on multiple choice question tasks this pattern is less extreme but skew toward early positions is still visible.
Moreover, since USC requires explicit answer extraction, tasks such as MuSR-ta revealed many failure cases (\textit{e.g.}, over 40\% failures in Table \ref{tab:taskwise_distribution}), further highlighting its fragility.
This highlights a critical weakness in prompt-based selection: the output is often determined more by position than content. 
Based on these observations, we conducted additional experiments where we randomly rearranged the order of candidate responses.
Indeed, we observed this issue in Table \ref{tab:usc_ordering}; on GPQA, for example, USC’s accuracy varied notably across different permutations (\textit{e.g.}, 46.0 $\rightarrow$ 41.9), demonstrating its sensitivity to presentation order. 
In contrast, our approach mitigates such ordering artifacts by decoupling few-shot demonstrations from the selection prompt and using them only for scoring. Furthermore, LLM-as-Judge does not perform reliably on more complex tasks (\textit{e.g.}, GPQA showing a noticeable accuracy degradation compared to other methods). These results emphasize that naively incorporating a few examples into prompts does not guarantee consistent gains, and that \name{} is more robust and scalable. Finally, we note that the application of prompt-based approach could be limited due to inherent input context-window length.
For reference, we also report an oracle upper bound in Table~\ref{tab:prompt_based_result}. This represents the accuracy achieved when selecting the optimal response from $N$ samples. This serves purely as a ceiling to describe in context how close each method approaches the maximum achievable performance. Although the gap with this ceiling is still noticeable, this highlights meaningful room for future improvements.

\subsection{Complementarity with majority voting}\label{app:self_cons}
\begin{table*}[t]
\caption{
\textbf{Complementarity with self-consistency.}
We compare standard Self-Consistency with \name{} and a rank-weighted voting variant that incorporates \name{} selection scores.
The results show that \name{} is competitive and can further complement majority voting.
}

\label{tab:self_consistency}
\centering
\resizebox{\textwidth}{!}{
\begin{tabular}{llccccccc|c}
\toprule
Models & Methods & MuSR-ta & MuSR-op & GPQA & MATH500 & DROP & HotpotQA & MMLU-PRO & Avg. \\
\midrule
\multirow{3}{*}{LLaMA-3.1-8B}
& Self-Consistency & 76.4 & \underline{54.7} & 26.3 & \underline{52.0} & \underline{73.0 / 78.0} & 23.0 / 29.8 & \underline{45.1} & 50.1 \\
& \name   & \textbf{79.6} & \textbf{57.8} & \textbf{35.4} & {51.2} & 69.4 / {75.7} & \underline{25.0 / 33.2} & \underline{45.1} & \underline{51.9} \\ 
& + Borda-vote(p=1) & \underline{78.4} & \textbf{57.8} & \underline{30.3} & \textbf{53.2} & \textbf{74.0 / 79.7} & \textbf{26.8 / 34.9} & \textbf{46.4} & \textbf{52.4} \\ \midrule
\multirow{3}{*}{GPT-4o-mini}
& Self-Consistency & \underline{84.8} & {60.6} & \underline{43.4} & \textbf{79.6} & \underline{80.0 / 85.3} & 35.6 / 46.8 & \underline{65.3} & \underline{64.2} \\
& \name   & {82.8} & \textbf{61.3} & 41.9 & {77.8} & {79.2} / {84.9} & \underline{36.2 / 48.0} & {64.9} & {63.4} \\
& + Borda-vote(p=1) & \textbf{86.8} & \underline{60.9} & \textbf{44.4} & \underline{79.4} & \textbf{80.4 / 86.0} & \textbf{36.8 / 48.3} & \textbf{65.7} & \textbf{64.9} \\ \midrule
\multirow{3}{*}{GPT-4o}
& Self-Consistency & 86.8 & \underline{71.5} & \underline{50.5} & \underline{80.0} & \underline{84.4 / 91.6} & 45.6 / 60.5 & \textbf{76.4} & 70.7 \\
& \name   & \textbf{90.4} & \textbf{71.9} & \textbf{51.5} & {77.8} & {83.6} / {91.1} & \underline{47.0} / \underline{60.7} & {75.4} & \underline{71.1} \\ 
& + Borda-vote(p=1) & \underline{88.8} & \underline{71.5} & {49.5} & \textbf{80.2} & \textbf{84.8 / 92.0} & \textbf{48.0 / 61.9} & \underline{76.2} & \textbf{71.3} \\ 
\bottomrule
\vspace{-0.1in}
\end{tabular}
}
\end{table*}

As denoted in Section \ref{sec:1}, \textit{Self-Consistency} (majority voting) has an inherent limitation: it is only applicable when answers can be easily extracted and normalized for voting. For example, it is hard to be applied for open-ended text generation. For this reason, rather than the original self-consistency \citep{wang2023self}, we mainly consider Universal Self-Consistency (USC) \citep{chen2023universal} as a baseline, which uses prompt-based evaluation to select among free-form outputs. In fact, HotpotQA is a representative case where standard self-consistency cannot be reliably applied, since mapping free-form outputs into consistent discrete answer categories is non-trivial.

Although standard Self-Consistency is difficult to apply in open-ended tasks, we conducted additional evaluations using standard self-consistency. To enable it on HotpotQA, we identified overlapping shared spans among free-form responses, counted how many responses contain each span, and used these aggregated counts as voting scores. In addition, inspired by the weighted voting scheme in Self-Certainty \citep{kang2025scalable}, we experimented with a Borda-voting method that uses the \name{} metric as the weight, fixing the parameter $p=1$ to avoid introducing new hyperparameters. For notational simplicity, in this subsection we let $r_n$ denote the normalized extracted answer from the $n$-th candidate response.
Namely, the standard Self-Consistency selects the answer with the highest vote,
\[
n^\star_{\text{sc}}
= \arg\max_{n \in \{1,\dots,N\}}
\sum_{j=1}^{N} \mathbf{1}[r_j = r_n],
\]
in which each candidate contributes the same weight to one.
The Borda vote generalizes this formula by replacing uniform unit weights as follows with rank-based weights derived from the \name{} score.
Let $\text{rank}(n)$ denote the rank of candidate $r_n$ based on its \name{} score.
The corresponding borda weight is calculated as follows:
\[
w_n = (N - \text{rank}(n) + 1)^p
\]
The final selection score of candidate $n$ is obtained by summing the weights of all candidates who make the same prediction:
\[
k^\star_{\text{Borda}}
= \arg\max_{n \in \{1,\dots,N\}}
\sum_{j=1}^{N}  w_j*\mathbf{1}[r_j = r_n],
\]
The results are presented in the Table \ref{tab:self_consistency}. Overall, \name{} continues to outperform self-consistency on average. For instance, on MuSR-ta with LLaMA, \name{} shows substantially higher performance (\textit{e.g.}, 79.6 vs. 76.4). More importantly, these methods are complementary: applying Borda voting with \name{} yields notable improvements over conventional self-consistency, particularly on LLaMA.

\subsection{Applying \name{} to zero-shot responses}\label{app:zero-shot}
\begin{table}[t] 
\caption{\textbf{Applying \name{} to zero-shot responses.}
We apply the same \name{} scoring function to candidates generated by Zero-shot CoT.
These results show that \name{} can also improve response selection when demonstrations are used only for verification.}
\small
\resizebox{\columnwidth}{!}{
\begin{tabular}{l l ccc} 
\toprule
Models & Methods & MATH500 & GPQA & MuSR-ta \\
\midrule
\multirow{2}{*}{GPT-4o-mini} 
& Zero-shot & 76.4 & 43.0 & 56.2 \\
& \cc\name{} & \cc\textbf{78.2} & \cc\textbf{43.9} & \cc\textbf{58.8} \\
\midrule 
\multirow{2}{*}{GPT-4o} 
& Zero-shot & 77.5 & 48.8 & 66.6 \\ 
& \cc\name{} & \cc\textbf{80.8} & \cc\textbf{54.0} & \cc\textbf{69.6} \\
\midrule 
\multirow{2}{*}{LLaMA-3.1-8B} 
& Zero-shot & 44.2 & 21.6 & 39.6 \\ 
& \cc\name{} & \cc\textbf{50.8} & \cc\textbf{24.2} & \cc\textbf{41.2} \\
\bottomrule 
\end{tabular}}
\label{tab:zeroshot} 
\end{table}

As shown in Table~\ref{tab:main}, Zero-shot CoT often achieves higher accuracy than Few-shot CoT, reflecting the intrinsic knowledge of the model. 
Although we primarily apply \name{} to Few-shot CoT candidates, the same scoring function in Eq. \ref{eq.combined} can also be applied to reasoning paths generated by Zero-shot CoT. We focus on Few-shot CoT since it usually yields better reasoning paths, as shown in Table \ref{tab:main}.
With the experiments in Table~\ref{tab:zeroshot}, we verify that applying \name{} to Zero-shot CoT yields improvements. These results further suggest that the few-shot exemplars in \name{} mainly function as a post-hoc validation pipeline, rather than as generation guidance as in conventional Few-shot CoT.
Also, this effectiveness of \name{} under decoupling between generation and selection suggests a robust alternative to conventional few-shot prompting strategies, particularly in settings where few-shot examples are ineffective with LLMs.

\subsection{Robustness to few-shot examples}\label{app:few_shot}
\begin{table}[t]
\centering
\caption{\textbf{Results of different few-shot example sets.}
We evaluate GPT-4o-mini under three different few-shot example sets.
While absolute performance varies with the quality of the demonstrations and the resulting candidate pool, \name{} consistently improves over the corresponding Few-shot CoT.
}
\label{tab:few_shot_diff}
\small
\begin{tabular}{l ccc}
\toprule
{Methods} & {MATH500} & {GPQA} & {MuSR-ta} \\
\midrule
Few-shot 1 & 75.2 & 41.3 & 77.0 \\
\cc\name{} 1 & \cc\textbf{77.8} & \cc\textbf{41.9} & \cc\textbf{82.8} \\
\midrule
Few-shot 2 & 74.5 & 41.5 & 57.8 \\
\cc\name{} 2 & \cc\textbf{79.0} & \cc\textbf{43.4} & \cc\textbf{59.2} \\
\midrule
Few-shot 3 & 75.0 & 38.9 & 60.1 \\
\cc\name{} 3 & \cc\textbf{77.8} & \cc\textbf{41.9} & \cc\textbf{62.8} \\
\bottomrule
\end{tabular}
\end{table}

\begin{table}[t]
\centering
\caption{\textbf{Performance under weak few-shot examples.}
We evaluate \name{} when the few-shot examples are intentionally weaker than the original.
Although weaker demonstrations reduce the quality of the generated candidate pool, \name{} improves over the corresponding Few-shot CoT across generation models.}
\label{tab:weak_fewshot}
\small
\resizebox{\columnwidth}{!}{
\begin{tabular}{l l ccc}
\toprule
Models & Methods & MATH500 & GPQA & MuSR-ta \\
\midrule
\multirow{2}{*}{GPT-4o-mini} 
 & Few-shot & 73.7 & 41.1 & 57.5 \\
 & \cc\name{} & \cc\textbf{76.4} & \cc\textbf{43.4} & \cc\textbf{58.0} \\
\midrule
\multirow{2}{*}{GPT-4o} 
 & Few-shot & 75.2 & 46.1 & 71.0 \\
 & \cc\name{} & \cc\textbf{79.0} & \cc\textbf{47.0} & \cc\textbf{75.6} \\
\midrule
\multirow{2}{*}{LLaMA-3.1-8B} 
 & Few-shot & 39.5 & 26.6 & 38.7 \\
 & \cc\name{} & \cc\textbf{47.6} & \cc\textbf{33.3} & \cc\textbf{41.6} \\
\bottomrule
\end{tabular}}
\end{table}

\begin{table}[t]
\centering
\small
\caption{\textbf{Quality judgment scores of weak few-shot examples.}
GPT-4o judges the original and weak few-shot examples on a 1--10 scale.}
\resizebox{\columnwidth}{!}{
\begin{tabular}{l ccc}
\toprule
Judge by GPT-4o (1--10) & MATH500 & GPQA & MuSR-ta \\
\midrule
Few-shot   & 8 & 8 & 8 \\
Low quality & 3 & 5 & 4 \\
\bottomrule
\end{tabular}}
\label{tab:weak_fewshot_judge}
\end{table}

\name{} highly relies on few-shot examples for scoring of both forward and backward terms (Section \ref{sec:3.2}). 
This raises the question of how sensitive the method is to the choice of few-shot examples.
To answer this, we conduct a sensitivity study on MATH500, GPQA and MuSR-ta using GPT-4o-mini, where we use three different few-shot examples with one original and two newly sampled.
As shown in Table \ref{tab:few_shot_diff}, both Few-shot CoT and \name{} show variation across these different sets.
Interestingly, MuSR-ta once again highlights the importance of example quality; when synthesizing new data according to \citep{sprague2310musr} to use as few-shot examples, baseline accuracy significantly degrades. 
Nevertheless, \name{} demonstrates consistent performance improvements.
These results indicate that \name{} remains robust to exemplar choice and is consistently effective, rather than overfitted to specific demonstrations.

We believe that constructing accurate few-shot examples is a minimal effort that one should invest to guide LLMs (even humans) toward a proper behavior for a target task.
However, in practical applications, the clarity of few-shot example may not always be guaranteed.
To evaluate robustness, we first synthesize low-quality few-shot examples by converting the original examples via prompting GPT-4o-mini to degrade the quality of reasoning in data.
As shown in Table~\ref{tab:weak_fewshot}, \name{} maintains consistent improvements even under degraded exemplars, indicating that recycled few-shot examples during verification remains effective without well curated examples; for instance, on MATH500 with LLaMA-3.1-8B, accuracy improves from 39.5\% to 47.6\% (+8.1), demonstrating \name{}'s robustness.
Overall, these results confirm that \name{} remains effective even with suboptimal examples and can provide meaningful gains in more practical, less curated scenarios.

To verify the degradation, we asked GPT-4o to evaluate the quality of the original set and the weak example set. The evaluation was conducted in random order, and information about each set was not provided to avoid bias. 
As shown in Table~\ref{tab:weak_fewshot_judge}, the weak set consistently received significantly lower scores (3–5 points) compared to the original examples (8 points). This further demonstrates that \name{} maintains its effectiveness even when the quality of the provided examples is low.
See list \ref{lst:generate_weak}-\ref{lst:judging_weak} for exact prompts used in the generation and evaluation assessment.

\subsection{Robustness to generation and evaluation prompts}\label{app:prompt_template}
\begin{table*}[t]
\centering
\centering
\caption{\textbf{Robustness to generation and evaluation prompts.}
We vary the prompt style used to generate candidate responses and the prompt style used for verification.
Differences across generation prompts mainly reflect changes in the quality of the initial candidate pool.
}
\label{tab:prompt_template}
\small
\begin{tabular}{c|c|cc|cc|cc}
\toprule
\multirow{2}{*}{{Gen Prompt}} & \multirow{2}{*}{{Eval Prompt}} 
  & \multicolumn{2}{c|}{{MATH500}} 
  & \multicolumn{2}{c|}{{GPQA}}
  & \multicolumn{2}{c}{{MuSR-ta}} \\
& & Few-shot & \name & Few-shot & \name & Few-shot & \name \\
\midrule
\multirow{3}{*}{{Orig}}
  & Orig & 75.2 & \cc\textbf{77.8} & 41.3 & \cc\textbf{41.9} & 77.0 & \cc\textbf{82.8} \\
  & Plan & 75.2 & \cc\textbf{78.0} & 41.3 & \cc\textbf{42.4} & 77.0 & \cc\textbf{82.8} \\
  & Role & 75.2 & \cc\textbf{77.8} & 41.3 & \cc\textbf{41.9} & 77.0 & \cc\textbf{82.4} \\
\midrule
\multirow{2}{*}{{Plan}}
  & Plan & 74.6 & \cc\textbf{78.2} & 42.6 & \cc\textbf{47.5} & 77.0 & \cc\textbf{82.4} \\
  & Orig & 74.6 & \cc\textbf{78.4} & 42.6 & \cc\textbf{47.5} & 77.0 & \cc\textbf{82.4} \\
\midrule
\multirow{2}{*}{{Role}}
  & Role & 74.5 & \cc\textbf{78.2} & 43.5 & \cc\textbf{47.5} & 75.8 & \cc\textbf{81.6} \\
  & Orig & 74.5 & \cc\textbf{78.2} & 43.5 & \cc\textbf{47.0} & 75.8 & \cc\textbf{81.6} \\
\bottomrule
\end{tabular}
\end{table*}

Additionally, to investigate the impact of various prompt choices, we conduct new experiments with two alternative prompting techniques, following prior work planning  and role-playing \citep{wang2023plan, kong2023better}. 
Specifically, we assess the robustness of \name{} by varying prompts during both the generation stage and the verification stage by adapting different prompting styles (orig, plan, and role).
For {plan} and {role} prompting at the generation, we newly sample five responses similar to Table~\ref{tab:main}.
Table \ref{tab:prompt_template} shows the performance on MATH500, GPQA and MuSR-ta under different configurations.
A key observation is that verification performance remains highly consistent across different verification prompt styles for a given generation prompt style. 
For instance, on GPQA, performance for "plan $\rightarrow$ orig" and "plan $\rightarrow$ plan" conditions is identical (47.5 vs. 47.5), with similar consistency observed for the "role" condition (47.5 vs. 47.0). 
This indicates that \name{} is inherently robust to variations in prompt formatting during the evaluation stage.

However, we also observe that the initial quality of the generated candidate set varies depending on the prompt style. 
For relatively simple tasks like MATH500, the quality of generated responses is similar across prompts. 
Conversely, on the more challenging GPQA, prompts offering structured guidance (\textit{e.g.}, plans or roles) tend to generate higher-quality seeds, reflected in slightly higher accuracy. 
Consequently, \name{} performs better when the initial candidates are of higher quality.

\subsection{Robustness to estimation model calibration}\label{app:temperature}
\begin{table*}[t]

\caption{\textbf{Robustness to estimation model calibration.} We report the performance of \name{} across different temperature scaling factors $T \in \{0.5, 1.0, 1.5, 2.0\}$ applied to the estimation model before computing token-level likelihoods.
The results show that \name{} is largely stable across calibration settings.}
\centering
\small
\resizebox{\textwidth}{!}{
\begin{tabular}{l l cccccc|c}
\toprule
Models & Temp & MuSR-ta & MuSR-op & GPQA & MATH500 & DROP & HotpotQA  & Avg. \\
\midrule
\multirow{4}{*}{GPT-4o}
& 0.5 & \underline{90.4} & \underline{71.1} & \textbf{52.0} & 77.4 & \textbf{84.0 / 91.3} & \textbf{47.6 / 60.7} & \textbf{70.4} \\
& 1 & \underline{90.4} & \textbf{71.9} & \underline{51.5} & \textbf{77.8} & \underline{83.6 / 91.1} & \underline{47.0} / \textbf{60.7} & \textbf{70.4} \\
& 1.5 & \textbf{91.2} & \underline{71.1} & 50.5 & \underline{77.6} & 83.4 / \underline{91.1} & 45.0 / 59.2 & \underline{69.8} \\
& 2 & \underline{90.4} & \textbf{71.9} & 51.0 & \underline{77.6} & 81.8 / 89.9 & 44.8 / \underline{59.3} & 69.6 \\
\midrule
\multirow{4}{*}{GPT-4o-mini}
& 0.5 & 82.0 & 59.4 & \textbf{43.9} & \underline{78.0} & \underline{78.0} / 83.8 & 35.8 /47.1 & 62.9 \\
& 1 & 82.8 & \textbf{61.3} & 41.9 & 77.8 & \textbf{79.2 / 84.9} & \underline{36.2 / 48.0} & \underline{63.2} \\
& 1.5 & \textbf{83.6} & \underline{60.2} & \underline{43.4} & \underline{78.0} & \underline{78.0 / 84.7} & \textbf{36.4 / 48.3} & \textbf{63.3} \\
& 2 & \underline{83.2} & 59.0 & 41.4 & \textbf{78.4} & 77.4 / 84.1 & 35.2 / 47.6 & 62.4 \\
\midrule
\multirow{4}{*}{LLaMA-3.1-8B}
& 0.5 & \textbf{79.6} & \underline{57.0} & 33.3 & 50.6 & 69.8 / 75.9 & 24.6 / \underline{33.0} & 52.5 \\
& 1 & \textbf{79.6} & \textbf{57.8} & \textbf{35.4} & 51.2 & 69.4 / 75.7 & \underline{25.0} / \textbf{33.2} & \textbf{53.1} \\
& 1.5 & \underline{78.8} & \underline{57.0} & \underline{34.8} & \textbf{52.0} & \underline{70.2} / \underline{76.1} & \textbf{25.2} / \textbf{33.2} & \underline{53.0} \\
& 2 & 78.0 & 56.2 & 34.3 & \underline{51.8} & \textbf{71.0} / \textbf{77.1} & 24.6 / 32.7 & 52.7 \\
\bottomrule
\label{tab:temperature}
\end{tabular}}
\end{table*}
\arrayrulecolor{black}

In the main experiments, likelihood scores are computed from the estimator's original logits, corresponding to $T=1.0$. 
A natural question is how sensitive \name{} is to the calibration of the estimator. 
To examine sensitivity to estimator calibration, we apply post-hoc temperature scaling to the logits before computing token-level likelihoods, using
$p_T(\cdot \mid x)=\mathrm{softmax}(z/T)$ with $T \in \{0.5, 1.0~(\text{default}), 1.5, 2.0\}$.
As shown in Table \ref{tab:temperature},  \name{} remains remarkably robust across temperatures, showing only minimal variation even under substantial overconfidence or underconfidence. 

This robustness is further demonstrated through cross-model and cross-scale evaluations. As shown in Figure \ref{fig:estimation} and Appendix \ref{app:estimation_model} (Table \ref{tab:estimation_full}), \name{} using the 1B estimation model maintains stable performance across various tasks and calibration environments, whereas the likelihood-based baseline CoT-WP exhibits significantly greater variability. 
These observations suggest that \name{} does not heavily rely on the precise calibration of the estimation model. The backward term provides an additional signal that helps compensate for calibration drift by capturing explanatory alignment with the few-shot examples rather than relying solely on raw likelihood peaks.

\subsection{Robustness to nearest neighbor approximation}\label{app:nearest}
\begin{table*}[t]
\caption{\textbf{Robustness analysis of fixed exemplar selection.}
We compare \name{} under three exemplar-selection settings: the lowest-performing fixed exemplar (worst), the oracle upper bound over fixed exemplars (best), and our semantic nearest-neighbor selection strategy (ours).}
\label{tab:fixed_exemplar}
\centering
\small
\resizebox{\textwidth}{!}{
\begin{tabular}{ll|ccccccc|c}
\toprule
\multirow{2}{*}{Models} & \multirow{2}{*}{Methods}
& MuSR-ta & MuSR-op & GPQA & MATH500 & DROP & HotpotQA & MMLU-PRO & \multirow{2}{*}{Avg.} \\
& & (Acc.) & (Acc.) & (Acc.) & (Acc.) & (EM / F1) & (EM / F1) & (Acc.) & \\
\midrule
\multirow{3}{*}{GPT-4o}
& \name{} (worst) & 90.0 & 70.7 & 51.0 & \underline{77.8} & 83.4 / \underline{91.2} & 46.4 / 60.2 & 75.3 & 70.7 \\
& \name{} (best)  & \textbf{90.8} & \textbf{73.4} & \textbf{51.5} & \textbf{78.8} & \textbf{84.2 / 91.5} & \textbf{47.8 / 61.7} & \textbf{76.1} & \textbf{71.8} \\
& \textbf{\name{} (ours)} & \underline{90.4} & \underline{71.9} & \textbf{51.5} & \underline{77.8} & \underline{83.6} / 91.1 & \underline{47.0 / 60.7} & \underline{75.4} & \underline{71.1} \\
\midrule
\multirow{3}{*}{GPT-4o-mini}
& \name{} (worst) & \underline{82.8} & 60.2 & 41.4 & \underline{77.8} & 79.0 / 84.7 & 35.4 / 47.5 & \underline{64.9} & 63.1 \\
& \name{} (best)  & \textbf{83.2} & \textbf{61.7} & \textbf{42.4} & \textbf{78.2} & \textbf{79.8 / 85.5} & \textbf{36.6 / 48.8} & \textbf{65.8} & \textbf{64.0} \\
& \textbf{\name{} (ours)} & \underline{82.8} & \underline{61.3} & \underline{41.9} & \underline{77.8} & \underline{79.2 / 84.9} & \underline{36.2 / 48.0} & \underline{64.9} & \underline{63.4} \\
\midrule
\multirow{3}{*}{Llama-3.1-8B}
& \name{} (worst) & 78.8 & 55.9 & 33.8 & 50.8 & \underline{69.8 / 76.0} & 24.6 / 32.8 & 44.8 & 51.2 \\
& \name{} (best)  & \textbf{81.6} & \textbf{60.2} & \textbf{35.4} & \textbf{51.8} & \textbf{71.2 / 77.1} & \textbf{25.6 / 33.9} & \textbf{46.5} & \textbf{53.2} \\
& \textbf{\name{} (ours)} & \underline{79.6} & \underline{57.8} & \textbf{35.4} & \underline{51.2} & 69.4 / 75.7 & \underline{25.0 / 33.2} & \underline{45.1} & \underline{51.9} \\
\bottomrule
\end{tabular}
}
\end{table*}

\begin{table*}[t]
\caption{\textbf{Robustness analysis of increasing the number of nearest neighbors.}
We evaluate \name{} using the top-$k$ nearest exemplars across three generation models.
This confirms that utilizing multiple examples yields marginal differences, validating the efficiency and effectiveness of our top-1 design choice.}
\label{tab:topk_neighbors}
\centering
\small
\resizebox{\textwidth}{!}{
\begin{tabular}{ll|ccccccc|c}
\toprule
\multirow{2}{*}{Model}
& \multirow{2}{*}{Methods}
& MuSR-ta & MuSR-op & GPQA & MATH500 & DROP & HotpotQA & MMLU-PRO & \multirow{2}{*}{Avg.} \\
& & (Acc.) & (Acc.) & (Acc.) & (Acc.) & (EM / F1) & (EM / F1) & (Acc.) & \\
\midrule
\multirow{3}{*}{GPT-4o}
& Top-1 & \underline{90.4}  & \textbf{71.9}  & \textbf{51.5} & 77.8 & \textbf{83.6} / \underline{91.1} & \underline{47.0 / 60.7} & \textbf{75.4} & \underline{71.1} \\
& Top-2 & \textbf{90.8} & \textbf{71.9} & \textbf{51.5} & \underline{78.0} & \textbf{83.6} / \underline{91.1} & \underline{47.0 / 60.7} & \textbf{75.4} & \textbf{71.2} \\
& Top-3 & \textbf{90.8} & \underline{70.7} & \underline{51.0} & \textbf{78.2} & \textbf{83.6 / 91.2} & \textbf{47.2 / 60.8} & \textbf{75.4} & 71.0 \\
\midrule
\multirow{3}{*}{GPT-4o-mini}
& Top-1 & \textbf{82.8} & \textbf{61.3} & \underline{41.9} & \textbf{77.8} & 79.2 / 84.9 & \textbf{36.2 / 48.0} & \underline{64.9} & \underline{63.4} \\
& Top-2 & \textbf{82.8} & \underline{60.9} & \underline{41.9} & \textbf{77.8}  & \underline{79.4 / 85.1} & 35.6 / 47.4 & \underline{64.9} & 63.3 \\
& Top-3 & \textbf{82.8} & \textbf{61.3} & \textbf{42.4} & \textbf{77.8} & \textbf{79.6 / 85.3} & \underline{35.8 / 47.7} & \textbf{65.0} & \textbf{63.5} \\
\midrule
\multirow{3}{*}{LLaMA-3.1-8B}
& Top-1 & \textbf{79.6} & \textbf{57.8} & \textbf{35.4} & \underline{51.0} & 69.4 / 75.7 & \textbf{25.0 / 33.2} & \textbf{45.1} & \textbf{51.9} \\
& Top-2 & \textbf{79.6} & \underline{56.2} & 34.3 & \textbf{51.2} & \underline{69.6 / 76.0} & \underline{24.6 / 32.8} & \underline{45.0} & 51.5 \\
& Top-3 & \textbf{79.6} & 55.9 & \underline{34.8} & 50.8 & \textbf{70.2 / 76.4} & \underline{24.6 / 32.8} & \underline{45.0} & \underline{51.6} \\
\bottomrule
\end{tabular}}
\end{table*}
To further validate our nearest-neighbor approximation, we conduct two additional experiments.
First, we evaluate the robustness of \name{} to fixed single-exemplar selection.
For this purpose, we analyze two settings: a worst-case setting that uses the lowest-performing exemplar, and an oracle upper-bound setting in which an output is considered correct if any single exemplar produces the correct answer.
As shown in Table \ref{tab:fixed_exemplar}, even the worst-case fixed-exemplar setting achieves competitive performance, while our proposed semantic nearest-neighbor selection outperforms the worst-case setting in task average for every model.
Second, we also conducted experiments using GPT-4o-mini to measure whether utilizing multiple nearest neighbors (top-k) enhances performance.
We compared the top-1 nearest neighbor against top-2 and top-3 neighbors.
As shown in Table~\ref{tab:topk_neighbors}, the average scores across benchmarks remained highly stable (63.4, 63.3, and 63.5, respectively). This confirms that utilizing multiple examples yields marginal differences, validating the efficiency and effectiveness of our top-1 design choice.
Overall, these results demonstrate that nearest-neighbor approximation is an efficient and effective design choice for \name{}.

\subsection{Robustness to different estimation models and computational efficiency}\label{app:estimation_model}
\begin{table}[t]
\centering
\caption{\textbf{Robustness to different estimation models.}
Full results with different estimation models on MATH500, GPQA, and MuSR-ta.
\name{} remains effective across estimator sizes and backbones.
}
\label{tab:estimation_full}
\small
\resizebox{\columnwidth}{!}{
\begin{tabular}{l|ccc|c}
\toprule
\multicolumn{5}{c}{{(a) MATH500}} \\
{Estimation} & {GPT-4o-mini} & {GPT-4o} & {LLaMA-3.1-8B} & {Avg} \\
\midrule
LLaMA-3.2-1B   & 78.0 & 77.6 & 51.4 & 69.0 \\
LLaMA-3.1-8B   & 77.8 & 77.8 & 51.2 & 68.9 \\
Qwen-2.5-7B    & 78.8 & 79.2 & 52.0 & 70.0 \\
LLaMA-3.1-70B  & 77.8 & 77.6 & 53.6 & 69.7 \\
\midrule
\multicolumn{5}{c}{{(b) GPQA}} \\
{Estimation} & {GPT-4o-mini} & {GPT-4o} & {LLaMA-3.1-8B} & {Avg} \\
\midrule
LLaMA-3.2-1B   & 43.9 & 50.5 & 33.8 & 42.7 \\
LLaMA-3.1-8B   & 41.9 & 51.5 & 35.4 & 42.9 \\
Qwen-2.5-7B    & 41.4 & 50.5 & 34.3 & 42.1 \\
LLaMA-3.1-70B  & 42.4 & 53.5 & 34.8 & 43.6 \\
\midrule
\multicolumn{5}{c}{{(c) MuSR-ta}} \\
{Estimation} & {GPT-4o-mini} & {GPT-4o} & {LLaMA-3.1-8B} & {Avg} \\
\midrule
LLaMA-3.2-1B   & 83.2 & 90.8 & 80.0 & 84.7 \\
LLaMA-3.1-8B   & 82.8 & 90.4 & 79.6 & 84.3 \\
Qwen-2.5-7B    & 82.0 & 90.8 & 81.6 & 84.8 \\
LLaMA-3.1-70B  & 83.6 & 91.2 & 81.6 & 85.5 \\
\bottomrule
\end{tabular}}%
\end{table}

\begin{table*}[t]
\centering
\caption{\textbf{Full results of cost--accuracy trade-off.}
Accuracy and evaluation cost across benchmarks.
Costs are measured in actual processing time (seconds) per instance on a single NVIDIA RTX A6000 GPU using the same model configuration.
All methods select from the same pre-generated candidate pool, so generation cost is shared and omitted.
The best and second-best accuracy scores are highlighted in bold and underline, respectively.}
\label{tab:cost}
\small
\resizebox{\textwidth}{!}{
\begin{tabular}{c|l|cccccccc}
\toprule
\multirow{2}{*}{Size} & \multirow{2}{*}{Methods}
& MuSR-ta & MuSR-op & GPQA & MATH500 & DROP & HotpotQA & MMLU-PRO & Avg. \\
&
& (Acc. / Time)
& (Acc. / Time)
& (Acc. / Time)
& (Acc. / Time)
& (Acc. / Time)
& (Acc. / Time)
& (Acc. / Time)
& (Acc. / Time) \\
\midrule
-- 
& Random 
& 76.3 / -- & 60.8 / -- & 37.7 / -- & 64.6 / -- & 72.9 / -- 
& 32.5 / -- & 58.5 / -- & 57.6 / -- \\
\midrule
\multirow{4}{*}{1B}
& USC
& 75.5 / 0.80 & \underline{62.1} / 0.48 & \underline{40.9} / 0.68 & 65.4 / 0.56 & 73.3 / 0.24
& 32.7 / 0.31 & 58.5 / 0.39 & 58.3 / 0.49 \\
& USC $(N{=}20)$
& 76.4 / 16.0 & 60.7 / 9.6 & \underline{40.9} / 13.6 & 64.7 / 11.2 & 74.3 / 4.8
& 32.9 / 6.2 & \underline{58.6} / 7.8 & 58.4 / 9.8 \\
& CoT-WP
& \underline{78.0} / 3.06 & 58.9 / 3.06 & \underline{40.9} / 2.12 & \underline{66.6} / 1.49 & \underline{75.9} / 0.45
& \underline{35.3} / 0.65 & \underline{58.6} / 1.39 & \underline{59.2} / 1.75 \\
& \cc\name{}
& \cc{\textbf{84.7} / 6.10} & \cc{\textbf{62.8} / 6.51} & \cc{\textbf{42.7} / 4.23} & \cc{\textbf{69.0} / 3.12} & \cc{\textbf{77.1} / 0.91}
& \cc{\textbf{35.9} / 1.32} & \cc{\textbf{61.5} / 2.78} & \cc{\textbf{62.0} / 3.57} \\
\midrule
\multirow{4}{*}{8B}
& USC
& 75.6 / 4.67 & \underline{61.4} / 3.81 & 40.6 / 4.10 & \underline{68.3} / 3.34 & 76.2 / 2.00
& 35.0 / 2.21 & 61.3 / 2.47 & 59.8 / 3.23 \\
& USC $(N{=}20)$
& 76.3 / 93.4 & 61.3 / 76.2 & \underline{40.9} / 82.0 & 68.1 / 66.8 & 76.5 / 40.0
& 35.6 / 44.2 & \textbf{62.5} / 49.4 & 60.2 / 64.6 \\
& CoT-WP
& \underline{79.7} / 15.37 & 59.9 / 15.52 & 39.7 / 11.64 & 68.0 / 8.27 & \underline{77.1} / 2.53
& \textbf{36.3} / 3.70 & 61.6 / 7.64 & \underline{60.3} / 9.24 \\
& \cc\name{}
& \cc{\textbf{84.3} / 42.56} & \cc{\textbf{63.7} / 42.39} & \cc{\textbf{42.9} / 23.64} & \cc{\textbf{68.9} / 17.41} & \cc{\textbf{77.4} / 6.99}
& \cc{\underline{36.1} / 6.94} & \cc{\underline{61.8} / 15.73} & \cc{\textbf{62.2} / 22.24} \\
\bottomrule
\end{tabular}}
\end{table*}
\begin{table}[t]
\centering
\caption{\textbf{Generator--estimator mismatch.} The same 11{,}020 candidates generated by LLaMA-3.1-8B (excluding MMLU-Pro) are rescored by a matched estimator, a smaller same-family estimator, and a different-family estimator.}
\label{tab:mismatch}
\begingroup
\small
\setlength{\tabcolsep}{4pt}
\resizebox{\columnwidth}{!}{
\begin{tabular}{@{}lcccc@{}}
\toprule
Estimator & PPL & corr(PPL, correct.) & \name{} Acc. & Random \\
\midrule
LLaMA-3.1-8B & 4.19 & $-0.411$ & 53.1 & 43.7 \\
LLaMA-3.2-1B & 5.25 & $-0.395$ & 52.4 & 43.7 \\
Qwen-2.5-7B  & 6.20 & $-0.396$ & \textbf{53.5} & 43.7 \\
\bottomrule
\end{tabular}}
\endgroup
\vspace{-0.1in}
\end{table}

Table \ref{tab:estimation_full} provides a full detailed breakdown of \name{} across all generation--estimation model combinations of MATH500, GPQA and MuSR-ta. 
This complements the average performance across different generation LLMs (GPT-4o-mini, GPT-4o, and LLaMA3.1-8B) shown in Figure \ref{fig:estimation}. 
Across all three tasks, \name{} shows consistent performance gains regardless of the estimation model used, emphasizing its robustness. 
At the same time, we observe task-dependent estimator trends: Qwen-2.5-7B performs best on MATH500, whereas LLaMA-3.1-70B performs best on GPQA and MuSR-ta, as discussed in Section~\ref{sec:4.3}.
Notably, even the smaller model (LLaMA-3.2-1B) performs competitively on (relatively) simple tasks such as MATH500, on average, indicating that large estimator capacity is not always necessary for relatively familiar or structured reasoning tasks.
These results suggest that different estimators may align better with different reasoning styles, while the proposed few-shot-based verification strategy remains broadly stable.

While the lightweight approximation offers significant computational advantages, replacing the entire Bayesian term in Eq.~\ref{eq:log-bayes} with the single most relevant example implies a theoretical simplification. This reduction may suggest a departure from the ostensibly rigorous Bayesian rule. 
However, previous work on in-context learning has observed that the relative contributions of examples are highly uneven, and that the most relevant examples often account for a disproportionately large proportion of useful signals \citep{wang2024learning, li2023compressing, liu2022makes}. From this perspective, the lightweight version does not discard the conceptual role of the entire backward component, but provides a tractable approximation to its dominant signal.
In particular, instead of computing backward term over all $K$ few-shot examples, \name{} evaluates only the most relevant exemplar selected by Eq.~\ref{eq:selection}, reducing the backward computation from $O(K)$ to $O(1)$ per candidate.
Our experiments support this interpretation. 
We provide detailed comparisons with the \textit{Full} and \textit{No replace} variants in Appendix~\ref{app:add_ablation}.

Furthermore, the modularity that separates generation from estimation allows our method to maintain its theoretical validity, even with smaller estimation models. This not only avoids the collapse of Bayesian interpretation, but also provides practical efficiency benefits.
To further examine this practical benefit, we provide an extended version of the cost-accuracy analysis from Section~\ref{sec:4.3} in Table~\ref{tab:cost}.
While the main text reports the summarized trade-off, Table~\ref{tab:cost} presents full benchmark results.
Since all methods select from the same pre-generated candidate pool, the generation cost is shared across methods; therefore, the reported cost only reflects the additional evaluation time required for response selection.

The extended results show that \name{} provides a better accuracy-cost trade-off than the baselines.
USC with a 1B estimator has the lowest evaluation cost, but its performance remains close to random selection.
Increasing the number of USC samples to $N{=}20$ with a voting scheme substantially increases computation, yet yields only limited accuracy gains.
CoT-WP benefits from a stronger estimator, but it experiences noticeable accuracy degradation when reducing the estimator from 8B to 1B.
In contrast, \name{} maintains stable performance across estimator sizes.

Specifically, \name{} with a 1B estimator achieves an average accuracy of 62.0\%, outperforming 8B CoT-WP, which achieves 60.3\%.
At the same time, \name{} with a 1B estimator requires 3.57 seconds per instance on average, compared with 9.24 seconds for the stronger baseline 8B CoT-WP, making it approximately 2.6$\times$ faster.

These results indicate that the lightweight approximation does not collapse the theoretical framework. \name{} still maintains a conceptual Bayesian structure, and maintains the benefits of backward signal.
We believe that this effectiveness, even when the generation and estimation models are not aligned, is a key strength of our approach. This design choice makes \name{} broadly applicable and can be easily integrated to existing pipelines.

However, when the estimator differs from the generator, the computed likelihoods no longer reflect the generator's own distribution, which introduces additional variability. 
This setting is unavoidable in practice: for closed-source generators the token likelihoods are inaccessible, so verification necessarily relies on a separately chosen estimator. 
The relevant question is therefore not whether the estimator reproduces the generator's intrinsic distribution, but whether it still ranks candidate quality effectively. 
To test this directly, we fix the candidate set and vary only the estimator, rescoring the same 11,020 candidates generated by LLaMA-3.1-8B (excluding MMLU-Pro) with a matched estimator (LLaMA-3.1-8B), a smaller same-family estimator (LLaMA-3.2-1B), and a different-family estimator (Qwen-2.5-7B).

As shown in Table~\ref{tab:mismatch}, absolute likelihoods shift as expected: perplexity rises from 4.19 to 5.25 and then to 6.20, confirming that cross-family mismatch in particular affects calibration. The induced ranking, however, remains stable. 
The correlation between perplexity and correctness stays near $-0.40$ in all three cases, and selection accuracy ranges only from 52.4\% to 53.5\% against a 43.7\% random baseline, with the different-family estimator marginally the best despite the worst calibration. 
This is the property \name{} relies on, since candidates are ranked within a single query under one fixed estimator and one fixed demonstration set, so estimator-specific offsets are shared across candidates and largely cancel. 
We note nonetheless that the scores under mismatch express the estimator's preferences rather than the generator's own confidence, and can be shaped by calibration, knowledge, reasoning style, verbosity, and tokenization;
applicability to closed-source generators therefore rests on the ranking evidence above rather than on the derivation.

\subsection{Additional ablation}\label{app:add_ablation}
\begin{table*}[t]
\caption{\textbf{Ablation study on backward terms.}
We compare the proposed leave-one-out replacement strategy with a simpler no-replacement variant, across all three generation models.
Full variants average the backward signal over all few-shot examples, while lightweight variants use a single selected exemplar. The best average within each generator is in \textbf{bold}.}
\label{tab:no_replace}
\centering
\small
\resizebox{\textwidth}{!}{
\begin{tabular}{c|l|ccccccc|c}
\toprule
\multirow{2}{*}{Models} & \multirow{2}{*}{Methods} & MuSR-ta & MuSR-op & GPQA & MATH500 & DROP & HotpotQA & MMLU-PRO & \multirow{2}{*}{Avg.} \\
& & (Acc.) & (Acc.) & (Acc.) & (Acc.) & (EM / F1) & (EM / F1) & (Acc.) & \\
\midrule
\multirow{4}{*}{{GPT-4o}}
& No replace (full) & 90.4 & \textbf{74.2} & \textbf{52.0} & 78.0 & 83.4 / 90.8 & 45.6 / 59.6 & 75.3 & \textbf{71.3} \\
& No replace        & \textbf{90.8} & 72.3 & 50.5 & 77.8 & \textbf{83.6} / 90.9 & \textbf{47.2 / 61.1} & 75.1 & 71.0 \\
& \name{} (Full)    & \textbf{90.8} & 70.7 & 51.0 & \textbf{78.4} & \textbf{83.6 / 91.2} & 47.0 / 60.6 & \textbf{75.5} & 71.0 \\
&  \name{}     & 90.4 & 71.9 & 51.5 & 77.8 & \textbf{83.6} / 91.1 & 47.0 / 60.7 & 75.4 & 71.1 \\
\midrule
\multirow{4}{*}{{GPT-4o-mini}}
& No replace (full) & \textbf{82.8} & 60.2 & 42.4 & \textbf{78.0} & 78.4 / 84.2 & \textbf{36.2 / 48.0} & \textbf{65.0} & 63.3 \\
& No replace        & 82.4 & 60.2 & \textbf{42.9} & 77.6 & 78.4 / 84.1 & 35.8 / 47.6 & 64.7 & 63.1 \\
& \name{} (Full)    & \textbf{82.8} & \textbf{61.3} & 42.4 & 77.8 & \textbf{79.6 / 85.3} & 35.8 / 47.9 & \textbf{65.0} & \textbf{63.5} \\
&  \name{}     & \textbf{82.8} & \textbf{61.3} & 41.9 & 77.8 & 79.2 / 84.9 & \textbf{36.2 / 48.0} & 64.9 & 63.4 \\
\midrule
\multirow{4}{*}{{LLaMA-3.1-8B}}
& No replace (full) & 73.2 & 55.5 & 34.8 & 50.8 & 67.6 / 73.4 & 24.6 / 32.7 & 44.5 & 50.2 \\
& No replace        & 72.4 & 55.9 & 33.8 & 50.8 & 66.2 / 72.7 & 24.6 / 32.6 & 44.7 & 49.8 \\
& \name{} (Full)    & \textbf{79.6} & 55.9 & 34.8 & 50.8 & \textbf{70.2 / 76.4} & 24.6 / 32.8 & 44.9 & 51.5 \\
&  \name{}     & \textbf{79.6} & \textbf{57.8} & \textbf{35.4} & \textbf{51.2} & 69.4 / 75.7 & \textbf{25.0 / 33.2} & \textbf{45.1} & \textbf{51.9} \\
\bottomrule
\end{tabular}}
\end{table*}
\begin{table*}[t]
\centering
\caption{\textbf{Extended ablation across all generators and tasks.} Each row uses the same candidate pool and differs only in the selection score, averaged over three runs. \textit{Backward} and \textit{$-$Backward} select the largest and the smallest raw backward score, and \textit{\name{} (Full)} uses the full backward term of Eq.~\ref{eq.backward}. The best score within each generator is in \textbf{bold}.}
\label{tab:ablation_full}
\begingroup
\resizebox{\textwidth}{!}{
\begin{tabular}{c|c|ccccccc|c}
\toprule
\multirow{2}{*}{Models} & \multirow{2}{*}{Methods} & MuSR-ta & MuSR-op & GPQA & MATH500 & DROP & HotpotQA & MMLU-PRO & \multirow{2}{*}{Avg.} \\
& & (Acc.) & (Acc.) & (Acc.) & (Acc.) & (EM / F1) & (EM / F1) & (Acc.) & \\
\midrule
\multirow{6}{*}{\rotatebox{90}{GPT-4o}}
&Backward        & 87.1 \ms{0.2} & 70.7 \ms{2.3} & 47.6 \ms{1.9} & 75.3 \ms{0.7} & 80.9 \ms{0.9} / 89.3 \ms{0.9} & 45.7 \ms{0.4} / 59.1 \ms{0.4} & 73.4 \ms{0.3} & 68.7 \\
&$-$Backward     & 87.5 \ms{0.7} & 69.7 \ms{1.4} & 48.3 \ms{1.7} & 75.0 \ms{0.2} & 78.7 \ms{0.6} / 88.1 \ms{0.5} & 45.7 \ms{1.2} / 59.1 \ms{0.7} & 74.0 \ms{0.2} & 68.4 \\
&Direct          & 90.8 \ms{1.1} & 73.3 \ms{2.4} & \textbf{52.2 \ms{1.3}} & 77.2 \ms{1.7} & 81.9 \ms{2.0} / 89.8 \ms{1.2} & 46.7 \ms{0.5} / 60.5 \ms{0.8} & 75.4 \ms{0.1} & 71.1 \\
&Forward         & 90.8 \ms{0.4} & 73.5 \ms{2.8} & 51.5 \ms{0.5} & 78.4 \ms{0.5} & 83.1 \ms{0.3} / 90.6 \ms{0.5} & \textbf{47.2 \ms{0.2} / 61.4 \ms{0.4}} & 75.4 \ms{0.0} & 71.4 \\
&\cc\name{}      & \cc90.8 \ms{0.4} & \cc\textbf{73.7 \ms{2.2}} & \cc51.8 \ms{0.3} & \cc78.5 \ms{0.6} & \cc\textbf{83.5 \ms{0.1} / 90.9 \ms{0.6}} & \cc46.9 \ms{0.4} / 61.0 \ms{0.7} & \cc75.4 \ms{0.3} & \cc\textbf{71.5} \\
&\cc\name{} (Full) & \cc\textbf{90.9 \ms{0.2}} & \cc73.0 \ms{2.1} & \cc51.2 \ms{0.2} & \cc\textbf{78.7 \ms{0.2}} & \cc83.3 \ms{0.4} / 90.8 \ms{0.5} & \cc47.1 \ms{0.2} / 61.3 \ms{0.5} & \cc75.4 \ms{0.0} & \cc71.4 \\
\midrule
\multirow{6}{*}{\rotatebox{90}{GPT-4o-mini}}
&Backward        & 78.7 \ms{0.7} & 61.6 \ms{2.3} & 40.9 \ms{1.6} & 74.1 \ms{0.7} & 77.3 \ms{0.8} / 83.5 \ms{0.5} & 34.3 \ms{0.7} / 45.5 \ms{0.6} & 62.7 \ms{0.5} & 61.4 \\
&$-$Backward     & 76.0 \ms{1.6} & 60.2 \ms{1.7} & 41.4 \ms{2.3} & 74.5 \ms{0.7} & 75.5 \ms{0.5} / 82.3 \ms{0.4} & 33.6 \ms{0.3} / 44.9 \ms{0.2} & 62.8 \ms{0.0} & 60.6 \\
&Direct          & 81.3 \ms{0.8} & 62.1 \ms{2.1} & 44.3 \ms{1.3} & \textbf{78.3 \ms{0.2}} & 78.9 \ms{0.6} / 84.2 \ms{0.9} & 35.1 \ms{0.6} / 46.9 \ms{0.4} & 64.8 \ms{0.2} & 63.5 \\
&Forward         & 82.9 \ms{0.2} & 61.8 \ms{0.8} & 44.3 \ms{2.3} & 78.1 \ms{0.3} & 79.1 \ms{0.4} / 84.6 \ms{0.7} & 35.3 \ms{0.5} / \textbf{47.4 \ms{0.6}} & 64.8 \ms{0.2} & 63.8 \\
&\cc\name{}      & \cc\textbf{83.1 \ms{0.2}} & \cc62.0 \ms{0.6} & \cc\textbf{44.6 \ms{2.6}} & \cc78.2 \ms{0.4} & \cc79.1 \ms{0.5} / \textbf{84.7 \ms{0.8}} & \cc\textbf{35.6 \ms{0.5}} / \textbf{47.4 \ms{0.6}} & \cc64.9 \ms{0.4} & \cc\textbf{63.9} \\
&\cc\name{} (Full) & \cc82.8 \ms{0.3} & \cc\textbf{62.2 \ms{0.7}} & \cc44.1 \ms{1.3} & \cc78.1 \ms{0.2} & \cc79.1 \ms{0.7} / 84.6 \ms{0.9} & \cc35.4 \ms{0.3} / \textbf{47.4 \ms{0.4}} & \cc\textbf{65.0 \ms{0.2}} & \cc63.8 \\
\midrule
\multirow{6}{*}{\rotatebox{90}{LLaMA-3.1-8B}}
&Backward        & 65.1 \ms{0.9} & 51.6 \ms{0.8} & 22.4 \ms{2.1} & 41.0 \ms{1.6} & 64.7 \ms{0.9} / 69.5 \ms{0.9} & 19.5 \ms{1.9} / 25.8 \ms{1.9} & 36.9 \ms{0.3} & 43.0 \\
&$-$Backward     & 61.5 \ms{2.4} & 51.8 \ms{2.4} & 25.9 \ms{2.3} & 44.3 \ms{1.3} & 50.8 \ms{1.5} / 56.6 \ms{1.3} & 17.7 \ms{0.7} / 22.8 \ms{1.2} & 40.4 \ms{0.3} & 41.8 \\
&Direct          & 68.4 \ms{3.0} & 56.0 \ms{4.8} & 32.6 \ms{2.0} & \textbf{50.3 \ms{1.3}} & 68.2 \ms{0.0} / 74.0 \ms{0.1} & 25.1 \ms{1.0} / 33.2 \ms{1.5} & 44.0 \ms{0.3} & 49.2 \\
&Forward         & \textbf{76.9 \ms{4.6}} & 57.0 \ms{0.4} & 32.3 \ms{1.3} & 50.1 \ms{1.6} & \textbf{71.5 \ms{1.7} / 76.7 \ms{0.5}} & 25.6 \ms{0.8} / \textbf{33.6 \ms{0.7}} & 44.6 \ms{0.2} & 51.1 \\
&\cc\name{}      & \cc76.5 \ms{3.4} & \cc\textbf{57.5 \ms{0.5}} & \cc\textbf{33.5 \ms{1.7}} & \cc50.2 \ms{1.6} & \cc70.9 \ms{1.6} / 76.3 \ms{0.5} & \cc\textbf{25.7 \ms{0.7}} / \textbf{33.6 \ms{0.6}} & \cc\textbf{44.8 \ms{0.3}} & \cc\textbf{51.3} \\
&\cc\name{} (Full) & \cc76.8 \ms{2.0} & \cc56.5 \ms{0.5} & \cc32.7 \ms{1.7} & \cc50.1 \ms{1.2} & \cc70.9 \ms{1.2} / 76.3 \ms{0.4} & \cc25.6 \ms{0.8} / 33.5 \ms{0.7} & \cc44.6 \ms{0.3} & \cc51.0 \\
\bottomrule
\end{tabular}
}
\endgroup
\vspace{-0.1in}
\end{table*}

Here, we conduct additional experiments to provide comprehensive ablation study for \name{}.
We first evaluate the \textit{Full} variant (Eq.~\ref{eq.backward}), which uses the complete set of examples and therefore provides the most faithful estimate of the backward reconstruction signal (Table~\ref{tab:no_replace}).
However, as discussed in Sec.~\ref{sec:3.2}, the computational overhead increases linearly with the number of few-shot examples, which renders the \textit{Full} variant less appealing for large-scale or resource-constrained scenarios.

To further analyze this trade-off, we examine the effectiveness of the proposed \textit{prompt replacement} (Eq.~\ref{eq.replace}) for better estimation of backward score. 
To this end, we consider a simplified variant of our backward reconstruction signal, termed \textit{No replace}, where each few-shot example $\mathbf{x}_i = (q_i, a_i)$ is evaluated in a one-shot manner using the test query $\widetilde{q}$ and the candidate response $r_n$ as additional context. 
Specifically, this variant modifies the backward term in Eq.~\ref{eq.backward} by replacing the leave-one-out prompt $\widetilde{\mathbf{X}}_{i}$ with a single pair $\mathbf{y}_n=(\widetilde{q}, r_n)$:
\begin{equation}
    \small
    \begin{aligned}
    S'_{\tt Back}(\mathbf{y}_n)
    &:= \log P(\mathbf{X} \mid \mathbf{y}_n) - \log P(\mathbf{X}) \\
    &= \frac{1}{K}\sum_{i=1}^{K}
    \left[
    \log P(a_i \mid q_i, \widetilde{q}, r_n)
    - \log P(a_i \mid q_i)
    \right].
    \end{aligned}
\end{equation}
We note that, as in our main method, a cost-efficient variant can be obtained by incorporating the $i^\dagger$ selection strategy (Eq. \ref{eq:selection}), which adaptively chooses the most relevant exemplar to the test query.
\begin{equation}
    \small
    S'_{\tt Back}(\mathbf{y}_n) := 
    \log P(a_{i^\dagger} \mid q_{i^\dagger}, \widetilde{q}, r_n) - \log P(a_{i^\dagger} \mid q_{i^\dagger}) ,
    \label{eq.no_replace}
\end{equation}
This formulation can be interpreted as the most straightforward implementation of backward term (see Eq.~\ref{eq:log-bayes}) under the assumption of mutual independence between few-shot examples.
Table~\ref{tab:no_replace} reports all four combinations across the three generation models, and two observations follow. 
First, prompt replacement is consistently beneficial: \name{} improves over \textit{No replace} on every generator (71.1 vs.\ 71.0, 63.4 vs.\ 63.1, and 51.9 vs.\ 49.8), and the same holds between their \textit{Full} counterparts. 
Second, \name{} is robust to how many exemplars enter the backward term, matching the \textit{Full} variant within 0.4 points on all three generators while requiring only $O(1)$ computation per candidate.
This indicates that \textit{No replace} still provides a useful backward penalty, but the penalty is less contextualized because it reconstructs each few-shot example using only the candidate pair $\mathbf{y}_n$.
Nonetheless, \textit{No replace} could serve as a practical alternative that trades off a small performance drop with the greater simplicity.

Table~\ref{tab:ablation_full} further extends the component ablation in Table~\ref{tab:ablation} to all three generation models and all seven tasks.
In addition to the \textit{Direct} and \textit{Forward} variants, we consider two single-term variants that select candidates using only the backward score, either by choosing the largest (\textit{Backward}) or the smallest (\textit{$-$Backward}) value. 
As shown in Table~\ref{tab:ablation_full}, neither variant performs competitively across the three generators, suggesting that the backward score alone does not serve as a reliable estimate of response correctness. 
We also observe that \textit{Direct} and \textit{Forward} exhibit complementary behavior across tasks, with neither consistently outperforming the other. 
In contrast, \name{} combines the two signals and is never worse than both variants across the 21 generator--task combinations, while achieving the best average accuracy for all three generators. 
The improvement over the forward-only variant is also consistent across the three matched runs, as further supported by the significance analysis in Table~\ref{tab:significance_ablation}.

\subsection{Additional comparison with direct scoring}\label{app:add_direct}
\begin{table*}[t]
\caption{\textbf{Additional comparison to direct terms.}
Model-averaged results comparing \name{} with direct likelihood scoring and zero-shot variants of uncertainty-based selection.
All methods use the same candidate responses, but differ in whether the estimator is conditioned on few-shot examples.
}
\label{tab:direct}
\centering
\small
\begin{tabular}{l|cccccc|c}
\toprule
\multirow{2}{*}{Methods} & MuSR-ta & MuSR-op & GPQA & MATH500 & DROP & HotpotQA & \multirow{2}{*}{Avg.} \\
 & (Acc.) & (Acc.) & (Acc.) & (Acc.) & (EM / F1) & (EM / F1) & \\
\midrule
Direct & \underline{80.2} & \underline{63.8} & \underline{43.0} & \underline{68.6} & \underline{76.3} & \underline{35.6} & \underline{61.2} \\
Self-Certainty (zero) & 79.2 & 62.5 & 40.8 & 68.2 & 76.2 & 34.7 & 60.3 \\
\ccrow\textbf{\name{}} & \textbf{83.5} & \textbf{64.4} & \textbf{43.3} & \textbf{69.0} & \textbf{77.8} & \textbf{36.1} & \textbf{62.4} \\
\bottomrule
\end{tabular}
\end{table*}

\begin{table*}[t]
\centering
\small
\caption{\textbf{Agreement between forward and backward selections.}
We measure how often the candidate preferred by the forward confidence score matches the candidate with the largest raw backward reconstruction score.}
\label{tab:agreement}
\resizebox{\textwidth}{!}{
\begin{tabular}{l l cccccccc}
\toprule
Models & Agreement & MuSR-ta & MuSR-op & GPQA & MATH500 & DROP & HotpotQA & MMLU-PRO & Avg \\
\midrule
\multirow{1}{*}{GPT-4o-mini} 
 & (F,B) & 0.188 & 0.180 & 0.242 & 0.242 & 0.234 & 0.244 & 0.208 & 0.220 \\
\multirow{1}{*}{GPT-4o} 
 & (F,B) & 0.212 & 0.238 & 0.303 & 0.222 & 0.268 & 0.260 & 0.226 & 0.247 \\
\multirow{1}{*}{Llama-3.1-8B-It} 
 & (F,B) & 0.140 & 0.312 & 0.066 & 0.182 & 0.246 & 0.226 & 0.143 & 0.188 \\
\bottomrule
\end{tabular}
}
\end{table*}

To further verify the concern raised in Section \ref{sec:3.2} that directly scoring candidate is unreliable when the model is miscalibrated in unfamiliar domains, we perform an additional comparison by fixing the candidate set and varying only the input condition. 
We reuse the same $N$ candidate responses ${r_n}$ generated from Table \ref{tab:main} under few-shot prompting $(\tilde q,\mathbf{X})$. 
The results in Table~\ref{tab:direct} were calculated by averaging the three independent sampling runs used and consolidating the modeled averages into a single table.
For \textit{direct confidence score}, we estimate each candidate the same as Eq.~\ref{eq.forward}, but without few-shot examples:
\begin{equation}
S_{\tt Direct}(r_n)
:= \log P(r_n\mid \tilde q).
\end{equation}
We also consider \textit{Self-Certainty (zero)}, a zero-shot conditioning variant of Self-Certainty, where uncertainty is computed without demonstrations.
Unlike the main table, we replace the conditioning $(\tilde q,\mathbf{X})$ with $(\tilde q)$ when computing the uncertainty for each $r_n$.

As shown in Table \ref{tab:direct}, candidates that score directly without demonstrations can be competitive in relatively familiar tasks such as MATH500, but become unreliable as the domain shifts.
In particular, the largest gap appears in the MuSR-ta. \name{} improves by 3.3\% (83.5 vs. 80.2) over the direct score. 
This demonstrates that relying solely on internal model knowledge is unreliable in unfamiliar domains. 
We also observe significant performance drop with Self-Certainty (zero). 
Self-Certainty quantifies the uncertainty by measuring how close the model's next token distribution is to the uniform distribution. 
When eliminating a few-shot examples, Self-Certainty can make the predictive distribution flat, especially in unfamiliar domain with weak calibration. 
In this case, it may not provide sufficient signals to select a plausible candidate among the candidates.
On the other hand, \name{} consistently benefits from recycling a few-shot context during validation, which provides external grounding signals and improves reliability, especially on benchmarks where few-shot guidance is important.

\subsection{Detailed analysis of the backward term}\label{app:backward_analysis}
\begin{table}[t]
\centering
\caption{\textbf{What the raw backward score correlates with.} Candidate-level Pearson correlations with correctness, response length, and similarity to the retrieved demonstration answer, measured by stemmed ROUGE-L and \texttt{all-mpnet-base-v2} cosine. \textit{Replace} is the score used by \name{} (Eq.~\ref{eq.new_backward}) and \textit{No-replace} the diagnostic variant of Eq.~\ref{eq.no_replace}.}
\label{tab:backward_corr}
\begingroup
\resizebox{\columnwidth}{!}{
\begin{tabular}{l cc cc cc}
\toprule
\multirow{2}{*}{Task} & \multirow{2}{*}{Correct.} & \multirow{2}{*}{Length} & \multicolumn{2}{c}{Replace} & \multicolumn{2}{c}{No-replace} \\
\cmidrule(lr){4-5}\cmidrule(lr){6-7}
 & & & ROUGE-L & cosine & ROUGE-L & cosine \\
\midrule
MuSR-ta   & $-0.011$ & $0.133$  & $-0.028$ & $0.041$  & $0.758$ & $0.578$ \\
MuSR-op   & $-0.049$ & $-0.043$ & $0.311$  & $0.126$  & $0.638$ & $0.296$ \\
GPQA      & $-0.029$ & $0.114$  & $0.046$  & $0.001$  & $0.038$ & $0.037$ \\
MATH500   & $-0.034$ & $0.056$  & $-0.026$ & $-0.013$ & $0.052$ & $0.001$ \\
DROP      & $0.088$  & $-0.225$ & $0.302$  & $0.078$  & $0.455$ & $0.365$ \\
HotpotQA  & $-0.012$ & $-0.012$ & $0.120$  & $0.091$  & $0.424$ & $-0.016$ \\
MMLU-Pro  & $0.030$  & $-0.117$ & $0.132$  & $0.049$  & $0.364$ & $0.165$ \\
\bottomrule
\end{tabular}}
\endgroup
\vspace{-0.1in}
\end{table}

\begin{table*}[t]
\centering
\caption{\textbf{Perturbation with demonstration fragments.} Raw forward and backward scores before and after prepending the retrieved demonstration answer, with 95\% confidence intervals; the tabular counterpart of Figure~\ref{fig:perturb}. The change is larger for the backward term in every cell.}
\label{tab:perturb_ci}
\begingroup
\small
\setlength{\tabcolsep}{6pt}
\begin{tabular}{@{}ll ccc@{}}
\toprule
Task & Generator & Original & $\rho{=}100$ & $\Delta$ \\
\midrule
\multicolumn{5}{@{}c}{\textit{Forward score}} \\
\multirow{3}{*}{MATH500}
 & LLaMA-3.1-8B & $-0.677_{\,[-0.722,-0.634]}$ & $-0.544_{\,[-0.575,-0.515]}$ & $+0.133$ \\
 & GPT-4o-mini  & $-0.668_{\,[-0.689,-0.648]}$ & $-0.509_{\,[-0.526,-0.493]}$ & $+0.159$ \\
 & GPT-4o       & $-0.744_{\,[-0.777,-0.712]}$ & $-0.551_{\,[-0.577,-0.527]}$ & $+0.193$ \\
\multirow{3}{*}{MuSR-ta}
 & LLaMA-3.1-8B & $-1.550_{\,[-1.640,-1.460]}$ & $-0.831_{\,[-0.870,-0.793]}$ & $+0.719$ \\
 & GPT-4o-mini  & $-1.236_{\,[-1.250,-1.222]}$ & $-0.766_{\,[-0.775,-0.758]}$ & $+0.470$ \\
 & GPT-4o       & $-1.180_{\,[-1.195,-1.165]}$ & $-0.709_{\,[-0.718,-0.700]}$ & $+0.471$ \\
\midrule
\multicolumn{5}{@{}c}{\textit{Backward score}} \\
\multirow{3}{*}{MATH500}
 & LLaMA-3.1-8B & $0.092_{\,[0.092,0.092]}$ & $0.396_{\,[0.391,0.401]}$ & $\mathbf{+0.304}$ \\
 & GPT-4o-mini  & $0.227_{\,[0.227,0.228]}$ & $1.002_{\,[0.985,1.021]}$ & $\mathbf{+0.775}$ \\
 & GPT-4o       & $0.227_{\,[0.227,0.228]}$ & $1.002_{\,[0.984,1.021]}$ & $\mathbf{+0.775}$ \\
\multirow{3}{*}{MuSR-ta}
 & LLaMA-3.1-8B & $0.641_{\,[0.640,0.642]}$ & $1.482_{\,[1.453,1.512]}$ & $\mathbf{+0.841}$ \\
 & GPT-4o-mini  & $0.602_{\,[0.602,0.603]}$ & $1.536_{\,[1.511,1.561]}$ & $\mathbf{+0.934}$ \\
 & GPT-4o       & $0.601_{\,[0.601,0.601]}$ & $1.536_{\,[1.510,1.561]}$ & $\mathbf{+0.935}$ \\
\bottomrule
\end{tabular}
\endgroup
\vspace{-0.1in}
\end{table*}

In this subsection, we provide the details of the analyses summarized in Section~\ref{sec:4.3}. 
As described in Section~\ref{sec:3.2}, the forward confidence score measures the likelihood of a candidate given the few-shot examples, while the backward reconstruction penalty suppresses candidates that are overly specific to the demonstrations. 
Intuitively, a large backward term implies that the candidate mimics the surface patterns or artifacts of the few-shot examples rather than reflecting the underlying reasoning logic. 
However, the same term also increases when a candidate correctly reuses a reasoning path that the demonstrations instantiate, and both cases occur in practice. 
The role of the backward term therefore cannot be determined from its magnitude alone. 
To this end, we examine what the raw score does not capture, what it actually tracks, and whether this sensitivity changes the selected response.

We first verify that the raw backward score is not a redundant estimate of response quality. 
As shown in Table~\ref{tab:agreement}, the candidate preferred by the forward score coincides with the one receiving the largest raw backward score in only 0.22 of the cases on average; since $N{=}5$, chance agreement is approximately 0.2, and the two terms therefore rank candidates almost independently. 
Table~\ref{tab:backward_corr} supports the same conclusion at the candidate level. 
The correlation with correctness is close to zero on all seven tasks ($|r|\le0.09$) and is as often negative as positive. 
If subtracting the backward term mainly penalized candidates that are simply better task instances, this correlation should be clearly positive. 
The correlation with response length is also limited ($|r|\le0.23$, with most values below 0.15), showing that the signal is not a proxy for verbosity.

We next examine what the raw score actually tracks. 
As shown in Table~\ref{tab:backward_corr}, the replacement score used by \name{} is positively associated with the lexical similarity between the candidate and the retrieved demonstration answer on most tasks. 
This association becomes substantially stronger under the no-replacement variant of Eq.~\ref{eq.no_replace} (up to 0.758 lexical and 0.578 semantic on MuSR-ta), where the contribution of the candidate is more directly isolated. 
This indicates that the raw backward term captures proximity to the demonstrations rather than response quality, which is why it serves as a penalty only after the forward score has established that a candidate is plausible.

The perturbation experiment in Figure~\ref{fig:perturb} indicates that this sensitivity is stronger for the backward term, and Table~\ref{tab:perturb_ci} reports the same measurements with 95\% confidence intervals, where the increase in the backward term exceeds that of the forward score in every generator--task cell. 
We then examine whether this difference changes the selected response, by injecting a demonstration artifact into a candidate that is already known to be incorrect. 
For each test query, we reuse the five candidates generated in the first run of Table~\ref{tab:main} and retrieve the demonstration with Eq.~\ref{eq:selection}, so that all five candidates of a query share the same exemplar. 
Among the incorrect candidates, we inject the artifact into the one with the \textit{lowest} forward score rather than one that the forward score already favors, which makes the test conservative: the artifact must be strong enough to overturn a selection that was initially unlikely. 
Concretely, we take the answer text of the retrieved demonstration, tokenize it with the tokenizer of the estimator in use, and prepend its first $\rho\%$ of tokens to that candidate, with $\rho \in \{0,25,50,100\}$. 
Only the modified candidate is rescored, so $\rho{=}0$ reproduces the original selection, and we keep the correctness labels fixed at their pre-injection values so that any change is attributable to the selector. 
This leaves 1,687 candidate sets pooled from three generators and three tasks.

As shown in Table~\ref{tab:perturb_selection}, the resulting pattern is monotone. 
Forward-only accuracy degrades from 45.29\% to 33.31\% with the 8B estimator and from 45.41\% to 30.71\% with the 1B estimator, since the injected prefix makes an incorrect candidate appear more consistent with the demonstrations. 
In contrast, \name{} improves to 47.48\% and 47.66\%, because the same insertion inflates the backward term more than the forward term and is subtracted away. 
Notably, the two estimator scales behave almost identically, indicating that this behavior stems from the scoring design rather than from estimator capacity.

Finally, we turn from an injected dependence to the demonstration similarity that arises naturally in the candidate pool. 
Here, we modify nothing and reuse the candidates and 8B estimator scores from the first run of Table~\ref{tab:main}. 
Within each query that contains both correct and incorrect candidates, we take the strongest candidate of either kind under the forward score, $c^{+}$ and $c^{-}$, so that the two most competitive candidates are directly opposed. 
For both, we measure the ROUGE-L similarity against the retrieved demonstration answer $a_{i^\dagger}$ and sort their difference within each generator$\times$task cell, comparing the two extreme quartiles. 
These correspond to the cases where the correct and the incorrect candidate is far more demonstration-similar, and are consistent with the reuse of a shared reasoning path and with demonstration mimicry, respectively.

Writing $\Delta B=\widetilde{S}_{\tt Back}(c^{-})-\widetilde{S}_{\tt Back}(c^{+})$, the margin by which \name{} prefers the correct candidate decomposes as
\begin{equation}
    \small
    \underbrace{S_{\tt Forw}(c^{+}) - S_{\tt Forw}(c^{-})}_{\text{forward margin}}
    \; + \;
    \underbrace{\Delta B}_{\text{backward correction}} ,
\end{equation}
so that the penalty helps exactly when the incorrect candidate receives the larger reconstruction score.
As shown in Table~\ref{tab:matched_pair}, both cases occur. 
When the correct candidate is more demonstration-similar, it does receive the larger backward score, confirming that reconstruction can reflect a correctly reused reasoning path; 
subtracting it nevertheless reduces pair accuracy by only 0.63 points, since that candidate also retains strong forward support. 
When the incorrect candidate is more similar, it receives the larger penalty and \name{} improves by 2.23 points. 
The penalty therefore remains close to inert when demonstration similarity is benign and becomes useful when it is not, which is how subtracting it discounts candidates whose forward support is overly explained by the few-shot examples.

\section{Qualitative Examples}\label{app:qual_examples}
In this section, we present qualitative examples to further analyze the proposed \name{}. For better readability, we only present the examples from MATH500, GPQA, and HotpotQA. All the responses are generated by GPT-4o-mini, and we use the \name{} \textit{(Full)} variant for illustration to provide the clearest comparisons.
\subsection{Token level analysis}
To better understand how \name{} identifies high-quality response using given few-shot examples, we perform a token-level analysis of following backward reconstruction penalty (Eq. \ref{eq.backward}). 
The visualization results are shown in Table~\ref{tab:token1}--Table~\ref{tab:token3}.

For a given test query $\widetilde{q}$, we divide the candidate responses into correct and incorrect groups using ground-truth labels. 
For each token, we calculate the difference between its token-level backward score under the two groups.
When the backward term for the tokens in the few-shot examples exhibit lower score in the correct group compared to incorrect one, the tokens are colored {\color{red}\textit{red}}. 
In other case, the tokens are colored {\color{blue}\textit{blue}}.
For visual clarity, we only highlight the top 60\% of tokens based on the absolute difference in values. The remaining 40\% remain uncolored. 
The value in parentheses is the ratio of tokens highlighted in red to the total number of tokens.

Since \name{} subtracts the backward term from the forward score, lower token-level backward scores for the correct candidate represent the token whose backward term contributes to distinguishing plausible responses.
Thus, the red token can be interpreted as part of several examples in which the right candidate induces less demonstration-specific reconstruction than the wrong candidate.
For example, in MATH500 and GPQA, numbers, symbols, final-answer formats, and intermediate reasoning steps are often highlighted in red, suggesting that the backward term helps identify candidates that avoid merely reproducing demonstration-specific structures.
In HotpotQA can also see that red is dominant for most, but not all, meaningful words.
\subsection{Response level analysis}
For each response selected by \name{}, we compute the Eq. \ref{eq.combined} for both best and worst response.
The value in parentheses is the final selection score.
The highest score among all candidates corresponds to the best response, and the lowest score represents the worst response. 
The visualization results are shown in Table \ref{tab:response1}–Table \ref{tab:response3}.

As shown in below examples, the selected response by \name{} has more accurate reasoning. 
For instance, in the example of MATH500, the best-scoring response anchors their reasoning in a coordinate system and follow clean logic. 
The worst response, on the other hand, ignore spatial cues, misapply subtraction, and over-complicate with lengthy and internally inconsistent steps. 
Similarly, in GPQA, the best response reasonably combines the results from the two analyses to arrive at the correct answer. 
The worst response, on the other hand, appears to rely primarily on the idea that \textit{“the heavy branching in (A) is consistent with the splitting observed”} without making a clear connection to the analyzed results.
Thus, one response integrates the data, while the other reduces it to a vague notion of complexity. 
In HotpotQA, selected answer is consistent with HotpotQA's multi-hop requirements by making intermediate hops explicit and factually correct, while rejected answer provides unsupported single-hop claims.

\section{Usage of AI Assistants}\label{app:usage}
This paper used AI-based writing aids to improve sentence structure, correct grammar, and improve readability. These tools were applied only to language refinement and did not affect the advancement of technical content, research methodology, or experimental analysis. All scientific ideas, results, and conclusions were conceived and written entirely by researchers. The use of AI aids was limited to editorial purposes and did not impair the originality or intellectual contribution of the work.

\onecolumn
\begin{longtable}{@{}p{\dimexpr\textwidth-2\tabcolsep\relax}@{}}
\caption{\textbf{Token-level qualitative analysis on MATH500.}} \\
\toprule
Few-shot examples for MATH500 (0.6455):\\
\midrule
\endfirsthead

\toprule
Few-shot examples for MATH500 (0.6455):\\
\midrule
\endhead

\small \textbf{Test-Query}: \newline
Twelve 1 by 1 squares form a rectangle, as shown. What is the total area of the shaded region? [asy] \newline
size(4cm);defaultpen(linewidth(0.75)); \newline \newline
// Filled portions \newline
fill((0, 4)--(0, 0)--(2, 0)--cycle, gray(0.75)); \newline
fill((0, 4)--(3, 4)--(3, 0)--cycle, gray(0.75)); \newline \newline
// grid \newline
int j; \newline
for (j = 0; j < 4; ++j) {draw((j, 0)--(j, 4));} \newline
for (j = 0; j < 5; ++j) {draw((0, j)--(3, j));} \newline \newline
//diagonals \newline
draw((0, 4)--(3, 0)); draw((0, 4)--(2, 0)); \newline
[/asy]

\\ \midrule 
\small \textbf{Few-shot 1 Query}: \newline
Kevin Kangaroo begins hopping on a number line at 0. He wants to get to 1, but he can hop only $\frac{1}{3}$ of the distance. Each hop tires him out so that he continues to hop $\frac{1}{3}$ of the remaining distance. How far has he hopped after five hops? Express your answer as a common fraction.

\textbf{Few-shot 1 Response:}\newline
\small {\color{red}\verb|Let|}{\color{red}\verb|'s|} {\color{red}\verb|think|} {\color{black}\verb|step|} {\color{black}\verb|by|} {\color{blue}\verb|step|}

{\color{red}\verb|Kevin|} {\color{red}\verb|hops|} {\color{blue}\verb|$|}{\color{red}\verb|1|}{\color{red}\verb|/|}{\color{black}\verb|3|}{\color{blue}\verb|$|} {\color{red}\verb|of|} {\color{black}\verb|the|} {\color{red}\verb|remaining|} {\color{black}\verb|distance|} {\color{red}\verb|with|} {\color{blue}\verb|every|} {\color{black}\verb|hop|}{\color{blue}\verb|.|}

{\color{blue}\verb|His|} {\color{blue}\verb|first|} {\color{black}\verb|hop|} {\color{blue}\verb|takes|} {\color{red}\verb|$|}{\color{red}\verb|1|}{\color{red}\verb|/|}{\color{black}\verb|3|}{\color{blue}\verb|$|} {\color{blue}\verb|closer|}{\color{red}\verb|.|}

{\color{red}\verb|For|} {\color{blue}\verb|his|} {\color{red}\verb|second|} {\color{black}\verb|hop|}{\color{blue}\verb|,|} {\color{blue}\verb|he|} {\color{black}\verb|has|} {\color{red}\verb|$|}{\color{blue}\verb|2|}{\color{black}\verb|/|}{\color{black}\verb|3|}{\color{red}\verb|$|} {\color{red}\verb|left|} {\color{blue}\verb|to|} {\color{blue}\verb|travel|}{\color{red}\verb|,|} {\color{blue}\verb|so|} {\color{black}\verb|he|} {\color{blue}\verb|hops|} {\color{blue}\verb|forward|} {\color{blue}\verb|$(|}{\color{blue}\verb|2|}{\color{black}\verb|/|}{\color{black}\verb|3|}{\color{red}\verb|)(|}{\color{black}\verb|1|}{\color{black}\verb|/|}{\color{black}\verb|3|}{\color{red}\verb|)$|}{\color{blue}\verb|.|}

{\color{red}\verb|For|} {\color{red}\verb|his|} {\color{black}\verb|third|} {\color{red}\verb|hop|}{\color{blue}\verb|,|} {\color{red}\verb|he|} {\color{black}\verb|has|} {\color{red}\verb|$(|}{\color{red}\verb|2|}{\color{black}\verb|/|}{\color{black}\verb|3|}{\color{red}\verb|)^|}{\color{blue}\verb|2|}{\color{black}\verb|$|} {\color{red}\verb|left|} {\color{red}\verb|to|} {\color{black}\verb|travel|}{\color{red}\verb|,|} {\color{black}\verb|so|} {\color{black}\verb|he|} {\color{black}\verb|hops|} {\color{red}\verb|forward|} {\color{red}\verb|$(|}{\color{black}\verb|2|}{\color{black}\verb|/|}{\color{black}\verb|3|}{\color{black}\verb|)^|}{\color{black}\verb|2|}{\color{red}\verb|(|}{\color{black}\verb|1|}{\color{black}\verb|/|}{\color{black}\verb|3|}{\color{red}\verb|)$|}{\color{red}\verb|.|}

{\color{red}\verb|In|} {\color{black}\verb|general|}{\color{black}\verb|,|} {\color{red}\verb|Kevin|} {\color{red}\verb|hops|} {\color{red}\verb|forward|} {\color{red}\verb|$(|}{\color{red}\verb|2|}{\color{black}\verb|/|}{\color{black}\verb|3|}{\color{black}\verb|)^|}{\color{red}\verb|{|}{\color{blue}\verb|k|}{\color{black}\verb|-|}{\color{black}\verb|1|}{\color{red}\verb|}(|}{\color{black}\verb|1|}{\color{black}\verb|/|}{\color{black}\verb|3|}{\color{black}\verb|)$|} {\color{red}\verb|on|} {\color{red}\verb|his|} {\color{red}\verb|$|}{\color{black}\verb|k|}{\color{red}\verb|$|}{\color{red}\verb|th|} {\color{black}\verb|hop|}{\color{red}\verb|.|}

{\color{blue}\verb|We|} {\color{blue}\verb|want|} {\color{blue}\verb|to|} {\color{blue}\verb|find|} {\color{blue}\verb|how|} {\color{red}\verb|far|} {\color{blue}\verb|he|} {\color{red}\verb|has|} {\color{red}\verb|h|}{\color{black}\verb|opped|} {\color{black}\verb|after|} {\color{red}\verb|five|} {\color{black}\verb|hops|}{\color{blue}\verb|.|}

{\color{red}\verb|This|} {\color{red}\verb|is|} {\color{blue}\verb|a|} {\color{blue}\verb|finite|} {\color{red}\verb|geometric|} {\color{red}\verb|series|} {\color{blue}\verb|with|} {\color{red}\verb|first|} {\color{black}\verb|term|} {\color{red}\verb|$|}{\color{red}\verb|1|}{\color{black}\verb|/|}{\color{black}\verb|3|}{\color{red}\verb|$,|} {\color{red}\verb|common|} {\color{black}\verb|ratio|} {\color{red}\verb|$|}{\color{black}\verb|2|}{\color{black}\verb|/|}{\color{black}\verb|3|}{\color{red}\verb|$,|} {\color{black}\verb|and|} {\color{red}\verb|five|} {\color{black}\verb|terms|}{\color{black}\verb|.|}

{\color{red}\verb|Thus|}{\color{blue}\verb|,|} {\color{red}\verb|Kevin|} {\color{red}\verb|has|} {\color{red}\verb|h|}{\color{black}\verb|opped|} {\color{blue}\verb|$\|}{\color{black}\verb|frac|}{\color{red}\verb|{\|}{\color{red}\verb|frac|}{\color{blue}\verb|{|}{\color{black}\verb|1|}{\color{black}\verb|}{|}{\color{black}\verb|3|}{\color{red}\verb|}\|}{\color{red}\verb|left|}{\color{red}\verb|(|}{\color{red}\verb|1|}{\color{red}\verb|-\|}{\color{black}\verb|left|}{\color{red}\verb|(\|}{\color{black}\verb|frac|}{\color{black}\verb|{|}{\color{black}\verb|2|}{\color{black}\verb|}{|}{\color{black}\verb|3|}{\color{red}\verb|}\|}

{\color{black}\verb|right|}{\color{black}\verb|)^|}{\color{blue}\verb|5|}{\color{red}\verb|\|}{\color{black}\verb|right|}{\color{black}\verb|)}|}{\color{red}\verb|{|}{\color{black}\verb|1|}{\color{red}\verb|-\|}{\color{blue}\verb|frac|}{\color{black}\verb|{|}{\color{black}\verb|2|}{\color{black}\verb|}{|}{\color{black}\verb|3|}{\color{red}\verb|}}|} {\color{red}\verb|=|} {\color{red}\verb|\|}{\color{blue}\verb|boxed|}{\color{blue}\verb|{\|}{\color{black}\verb|frac|}{\color{red}\verb|{|}{\color{red}\verb|211|}{\color{red}\verb|}{|}{\color{black}\verb|243|}{\color{black}\verb|}}|}{\color{blue}\verb|$|}{\color{blue}\verb|.|}

{\color{black}\verb|The|} {\color{black}\verb|answer|} {\color{black}\verb|is|} {\color{blue}\verb|\|}{\color{red}\verb|frac|}{\color{black}\verb|{|}{\color{black}\verb|211|}{\color{black}\verb|}{|}{\color{black}\verb|243|}{\color{red}\verb|}}|}
\\ \midrule

\small \textbf{Few-shot 2 Query:}\newline
What is the area of the region defined by the equation $x^2+y^2 - 7 = 4y-14x+3$?

\textbf{Few-shot 2 Response:}\newline
\small {\color{red}\verb|Let|}{\color{red}\verb|'s|} {\color{red}\verb|think|} {\color{black}\verb|step|} {\color{black}\verb|by|} {\color{red}\verb|step|}

{\color{red}\verb|We|} {\color{red}\verb|rewrite|} {\color{red}\verb|the|} {\color{blue}\verb|equation|} {\color{red}\verb|as|} {\color{red}\verb|$|}{\color{blue}\verb|x|}{\color{black}\verb|^|}{\color{black}\verb|2|} {\color{red}\verb|+|} {\color{red}\verb|14|}{\color{black}\verb|x|} {\color{red}\verb|+|} {\color{blue}\verb|y|}{\color{black}\verb|^|}{\color{black}\verb|2|} {\color{black}\verb|-|} {\color{black}\verb|4|}{\color{red}\verb|y|} {\color{blue}\verb|=|} {\color{black}\verb|10|}{\color{red}\verb|$|} {\color{red}\verb|and|} {\color{blue}\verb|then|} {\color{red}\verb|complete|} {\color{red}\verb|the|} {\color{red}\verb|square|}{\color{red}\verb|,|}

{\color{blue}\verb|result|}{\color{black}\verb|ing|} {\color{black}\verb|in|}  {\color{blue}\verb|$(|}{\color{black}\verb|x|}{\color{red}\verb|+|}{\color{black}\verb|7|}{\color{black}\verb|)^|}{\color{black}\verb|2|}{\color{red}\verb|-|}{\color{red}\verb|49|} {\color{red}\verb|+|} {\color{black}\verb|(|}{\color{black}\verb|y|}{\color{red}\verb|-|}{\color{black}\verb|2|}{\color{black}\verb|)^|}{\color{black}\verb|2|}{\color{red}\verb|-|}{\color{black}\verb|4|}{\color{blue}\verb|=|}{\color{red}\verb|10|}{\color{red}\verb|$|}{\color{blue}\verb|,|}

{\color{blue}\verb|or|} {\color{red}\verb|$(|}{\color{black}\verb|x|}{\color{red}\verb|+|}{\color{black}\verb|7|}{\color{black}\verb|)^|}{\color{black}\verb|2|}{\color{blue}\verb|+(|}{\color{black}\verb|y|}{\color{black}\verb|-|}{\color{black}\verb|2|}{\color{black}\verb|)^|}{\color{black}\verb|2|}{\color{red}\verb|=|}{\color{black}\verb|63|}{\color{red}\verb|$|}{\color{red}\verb|.|}

{\color{red}\verb|This|} {\color{blue}\verb|is|} {\color{blue}\verb|the|} {\color{black}\verb|equation|} {\color{blue}\verb|of|} {\color{black}\verb|a|} {\color{red}\verb|circle|} {\color{red}\verb|with|} {\color{red}\verb|center|} {\color{red}\verb|$|}{\color{black}\verb|(-|}{\color{black}\verb|7|}{\color{black}\verb|,|} {\color{black}\verb|2|}{\color{red}\verb|)$|} {\color{red}\verb|and|} {\color{black}\verb|radius|} {\color{red}\verb|$\|}{\color{black}\verb|sqrt|}{\color{blue}\verb|{|}{\color{black}\verb|63|}{\color{blue}\verb|},|}{\color{red}\verb|$|}

{\color{red}\verb|so|} {\color{red}\verb|the|} {\color{red}\verb|area|} {\color{blue}\verb|of|} {\color{blue}\verb|this|} {\color{red}\verb|region|} {\color{red}\verb|is|} {\color{red}\verb|$\|}{\color{red}\verb|pi|} {\color{blue}\verb|r|}{\color{black}\verb|^|}{\color{black}\verb|2|} {\color{red}\verb|=|} {\color{blue}\verb|\|}{\color{red}\verb|boxed|}{\color{red}\verb|{|}{\color{black}\verb|63|}{\color{red}\verb|\|}{\color{black}\verb|pi|}{\color{blue}\verb|}$|}{\color{blue}\verb|.|}

{\color{black}\verb|The|} {\color{black}\verb|answer|} {\color{black}\verb|is|} {\color{red}\verb|63|}{\color{red}\verb|\|}{\color{black}\verb|pi|}
\\ \midrule

\small \textbf{Few-shot 3 Query:}\newline
If $x^2+y^2=1$, what is the largest possible value of $|x|+|y|$?

\textbf{Few-shot 3 Response:}\newline
\small {\color{red}\verb|Let|}{\color{red}\verb|'s|} {\color{black}\verb|think|} {\color{black}\verb|step|} {\color{black}\verb|by|} {\color{red}\verb|step|}

{\color{blue}\verb|If|} {\color{red}\verb|$(|}{\color{blue}\verb|x|}{\color{red}\verb|,y|}{\color{black}\verb|)$|} {\color{red}\verb|lies|} {\color{black}\verb|on|} {\color{red}\verb|the|} {\color{blue}\verb|circle|}{\color{red}\verb|,|}

{\color{blue}\verb|so|} {\color{red}\verb|does|} {\color{red}\verb|$(|}{\color{red}\verb|x|}{\color{red}\verb|,-|}{\color{black}\verb|y|}{\color{red}\verb|),$|} {\color{blue}\verb|$|}{\color{black}\verb|(-|}{\color{black}\verb|x|}{\color{red}\verb|,-|}{\color{black}\verb|y|}{\color{blue}\verb|),$|} {\color{red}\verb|and|} {\color{black}\verb|$|}{\color{black}\verb|(-|}{\color{black}\verb|x|}{\color{blue}\verb|,-|}{\color{blue}\verb|y|}{\color{red}\verb|),$|} {\color{blue}\verb|(|}{\color{red}\verb|which|} {\color{blue}\verb|all|} {\color{red}\verb|give|} {\color{red}\verb|the|} {\color{blue}\verb|same|} {\color{blue}\verb|value|} {\color{blue}\verb|of|} {\color{blue}\verb|$|}{\color{blue}\verb!|!}{\color{black}\verb|x|}{\color{black}\verb!|!} {\color{red}\verb|+|} {\color{blue}\verb!|!}{\color{black}\verb|y|}{\color{blue}\verb!|$!}{\color{red}\verb|),|}

{\color{red}\verb|so|} {\color{red}\verb|we|} {\color{red}\verb|can|} {\color{red}\verb|assume|} {\color{red}\verb|that|} {\color{red}\verb|$|}{\color{blue}\verb|x|} {\color{blue}\verb|\|}{\color{blue}\verb|ge|} {\color{black}\verb|0|}{\color{red}\verb|$|} {\color{black}\verb|and|} {\color{black}\verb|$|}{\color{black}\verb|y|} {\color{blue}\verb|\|}{\color{red}\verb|ge|} {\color{black}\verb|0|}{\color{red}\verb|.$|}{\color{black}\verb|Then|} {\color{red}\verb|$|}{\color{red}\verb!|!}{\color{black}\verb|x|}{\color{red}\verb!|!} {\color{red}\verb|+|} {\color{black}\verb!|!}{\color{black}\verb|y|}{\color{red}\verb!|!} {\color{red}\verb|=|} {\color{red}\verb|x|} {\color{red}\verb|+|} {\color{black}\verb|y|}{\color{blue}\verb|.$|}  {\color{red}\verb|Squ|}{\color{black}\verb|aring|}{\color{red}\verb|,|} {\color{red}\verb|we|} {\color{red}\verb|get|}
{\color{red}\verb|\|}{\color{red}\verb|[(|}{\color{red}\verb|x|} {\color{red}\verb|+|} {\color{black}\verb|y|}{\color{black}\verb|)^|}{\color{black}\verb|2|} {\color{red}\verb|=|} {\color{red}\verb|x|}{\color{black}\verb|^|}{\color{black}\verb|2|} {\color{red}\verb|+|} {\color{black}\verb|2|}{\color{black}\verb|xy|} {\color{black}\verb|+|} {\color{black}\verb|y|}{\color{black}\verb|^|}{\color{black}\verb|2|} {\color{red}\verb|=|} {\color{blue}\verb|1|} {\color{red}\verb|+|} {\color{black}\verb|2|}{\color{black}\verb|xy|}{\color{red}\verb|.\|}{\color{red}\verb|]|}

{\color{red}\verb|Note|} {\color{black}\verb|that|} {\color{red}\verb|$(|}{\color{red}\verb|x|} {\color{blue}\verb|-|} {\color{red}\verb|y|}{\color{black}\verb|)^|}{\color{black}\verb|2|} {\color{blue}\verb|\|}{\color{black}\verb|ge|} {\color{black}\verb|0|}{\color{blue}\verb|.$|}{\color{red}\verb|Exp|}{\color{black}\verb|anding|}{\color{red}\verb|,|} {\color{black}\verb|we|} {\color{red}\verb|get|} {\color{blue}\verb|$|}{\color{blue}\verb|x|}{\color{black}\verb|^|}{\color{black}\verb|2|} {\color{red}\verb|-|} {\color{black}\verb|2|}{\color{black}\verb|xy|} {\color{black}\verb|+|} {\color{black}\verb|y|}{\color{black}\verb|^|}{\color{black}\verb|2|} {\color{blue}\verb|\|}{\color{black}\verb|ge|} {\color{black}\verb|0|}{\color{red}\verb|,$|} {\color{red}\verb|so|} {\color{red}\verb|$|}{\color{red}\verb|2|}{\color{black}\verb|xy|} {\color{black}\verb|\|}{\color{black}\verb|le|} {\color{blue}\verb|x|}{\color{black}\verb|^|}{\color{black}\verb|2|} {\color{black}\verb|+|} {\color{black}\verb|y|}{\color{black}\verb|^|}{\color{black}\verb|2|} {\color{red}\verb|=|} {\color{black}\verb|1|}{\color{red}\verb|.$|}
{\color{blue}\verb|H|}{\color{black}\verb|ence|}{\color{blue}\verb|,\|}{\color{red}\verb|[|}{\color{red}\verb|1|} {\color{red}\verb|+|} {\color{black}\verb|2|}{\color{black}\verb|xy|} {\color{black}\verb|\|}{\color{black}\verb|le|} {\color{red}\verb|2|}{\color{red}\verb|,\|}{\color{blue}\verb|]|}{\color{red}\verb|which|} {\color{blue}\verb|means|} {\color{blue}\verb|$|}{\color{red}\verb|x|} {\color{red}\verb|+|} {\color{black}\verb|y|} {\color{red}\verb|\|}{\color{black}\verb|le|} {\color{black}\verb|\|}{\color{red}\verb|sqrt|}{\color{blue}\verb|{|}{\color{black}\verb|2|}{\color{red}\verb|}|}{\color{red}\verb|.$|}{\color{red}\verb|Equality|} {\color{blue}\verb|occurs|} {\color{red}\verb|when|} {\color{red}\verb|$|}{\color{red}\verb|x|} {\color{red}\verb|=|} {\color{red}\verb|y|} {\color{black}\verb|=|} {\color{blue}\verb|\|}{\color{blue}\verb|frac|}{\color{red}\verb|{|}{\color{black}\verb|1|}{\color{black}\verb|}{|}{\color{black}\verb|\|}{\color{black}\verb|sqrt|}{\color{black}\verb|{|}{\color{black}\verb|2|}{\color{red}\verb|}},|}{\color{red}\verb|$|}

{\color{black}\verb|so|} {\color{red}\verb|the|} {\color{blue}\verb|maximum|} {\color{red}\verb|value|} {\color{red}\verb|of|} {\color{red}\verb|$|}{\color{red}\verb!|!}{\color{black}\verb|x|}{\color{black}\verb!|!} {\color{red}\verb|+|} {\color{black}\verb!|!}{\color{black}\verb|y|}{\color{black}\verb!|$!} {\color{red}\verb|is|} {\color{red}\verb|$\|}{\color{blue}\verb|boxed|}{\color{black}\verb|{\|}{\color{black}\verb|sqrt|}{\color{black}\verb|{|}{\color{black}\verb|2|}{\color{black}\verb|}}|}{\color{red}\verb|.$|}

{\color{blue}\verb|The|} {\color{black}\verb|answer|} {\color{black}\verb|is|} {\color{red}\verb|\|}{\color{red}\verb|sqrt|}{\color{blue}\verb|{|}{\color{black}\verb|2|}{\color{blue}\verb|}|}
\\ \midrule

\small \textbf{Few-shot 4 Query:}\newline
If $f(x)=\frac{ax+b}{cx+d}, abcd\not=0$ and $f(f(x))=x$ for all $x$ in the domain of $f$, what is the value of $a+d$?

\textbf{Few-shot 4 Response:}\newline
\small {\color{red}\verb|Let|}{\color{black}\verb|'s|} {\color{black}\verb|think|} {\color{black}\verb|step|} {\color{black}\verb|by|} {\color{black}\verb|step|}

{\color{red}\verb|The|} {\color{red}\verb|condition|} {\color{red}\verb|$|}{\color{black}\verb|f|}{\color{black}\verb|(f|}{\color{black}\verb|(x|}{\color{black}\verb|))|}{\color{black}\verb|$|} {\color{red}\verb|means|} {\color{blue}\verb|that|} {\color{black}\verb|$|}{\color{red}\verb|f|}{\color{red}\verb|$|} {\color{black}\verb|is|} {\color{red}\verb|the|} {\color{black}\verb|inverse|} {\color{black}\verb|of|} {\color{blue}\verb|itself|}{\color{blue}\verb|,|}

{\color{black}\verb|so|} {\color{blue}\verb|its|} {\color{black}\verb|graph|} {\color{red}\verb|is|} {\color{blue}\verb|sym|}{\color{black}\verb|metrical|} {\color{red}\verb|about|} {\color{black}\verb|the|} {\color{black}\verb|line|} {\color{black}\verb|$|}{\color{black}\verb|y|} {\color{red}\verb|=|} {\color{black}\verb|x|}{\color{red}\verb|$|}{\color{red}\verb|.|}

{\color{red}\verb|With|} {\color{blue}\verb|a|} {\color{blue}\verb|rational|} {\color{black}\verb|function|} {\color{blue}\verb|of|} {\color{blue}\verb|this|} {\color{black}\verb|form|}{\color{black}\verb|,|} {\color{red}\verb|we|} {\color{red}\verb|will|} {\color{red}\verb|have|} {\color{red}\verb|two|} {\color{red}\verb|asympt|}{\color{black}\verb|otes|}{\color{red}\verb|:|}

{\color{blue}\verb|a|} {\color{blue}\verb|vertical|} {\color{red}\verb|one|} {\color{red}\verb|at|} {\color{red}\verb|$|}{\color{red}\verb|x|}{\color{blue}\verb|=-|}{\color{red}\verb|d|}{\color{black}\verb|/c|}{\color{red}\verb|$|} {\color{blue}\verb|if|} {\color{red}\verb|$|}{\color{blue}\verb|cx|}{\color{black}\verb|+d|}{\color{blue}\verb|$|} {\color{blue}\verb|does|} {\color{black}\verb|not|} {\color{red}\verb|divide|} {\color{black}\verb|$|}{\color{black}\verb|ax|}{\color{black}\verb|+b|}{\color{black}\verb|$|}{\color{blue}\verb|,|}

{\color{black}\verb|and|} {\color{black}\verb|a|} {\color{black}\verb|horizontal|} {\color{black}\verb|one|} {\color{black}\verb|at|} {\color{black}\verb|$|}{\color{black}\verb|y|}{\color{red}\verb|=a|}{\color{black}\verb|/c|}{\color{red}\verb|$|}{\color{blue}\verb|,|}

{\color{red}\verb|if|} {\color{blue}\verb|we|} {\color{red}\verb|take|} {\color{red}\verb|the|} {\color{red}\verb|limit|} {\color{red}\verb|of|} {\color{blue}\verb|$|}{\color{red}\verb|f|}{\color{red}\verb|(x|}{\color{black}\verb|)$|} {\color{red}\verb|as|} {\color{black}\verb|$|}{\color{black}\verb|x|}{\color{red}\verb|$|} {\color{blue}\verb|goes|} {\color{black}\verb|to|} {\color{red}\verb|$\|}{\color{blue}\verb|pm|}{\color{red}\verb|\|}{\color{black}\verb|in|}{\color{black}\verb|fty|}{\color{blue}\verb|$|}{\color{red}\verb|.|}

{\color{red}\verb|In|} {\color{blue}\verb|order|} {\color{blue}\verb|for|} {\color{red}\verb|$|}{\color{red}\verb|f|}{\color{red}\verb|$|} {\color{black}\verb|to|} {\color{red}\verb|be|} {\color{black}\verb|its|} {\color{black}\verb|own|} {\color{black}\verb|inverse|}{\color{blue}\verb|,|} {\color{red}\verb|the|} {\color{blue}\verb|intersection|} {\color{blue}\verb|of|} {\color{red}\verb|the|} {\color{red}\verb|asympt|}{\color{black}\verb|otes|} {\color{red}\verb|must|} {\color{blue}\verb|lie|} {\color{red}\verb|on|} {\color{black}\verb|the|} {\color{blue}\verb|line|} {\color{black}\verb|$|}{\color{black}\verb|y|}{\color{red}\verb|=x|}{\color{red}\verb|$|}

{\color{red}\verb|so|} {\color{blue}\verb|that|} {\color{red}\verb|it|} {\color{blue}\verb|and|} {\color{red}\verb|its|} {\color{blue}\verb|asympt|}{\color{black}\verb|otes|} {\color{red}\verb|reflect|} {\color{blue}\verb|onto|} {\color{red}\verb|themselves|}{\color{red}\verb|.|}

{\color{red}\verb|This|} {\color{blue}\verb|means|} {\color{red}\verb|that|} {\color{red}\verb|$|}{\color{red}\verb|-|}{\color{black}\verb|d|}{\color{black}\verb|/c|}{\color{blue}\verb|=a|}{\color{black}\verb|/c|}{\color{black}\verb|$|}{\color{blue}\verb|,|}

{\color{red}\verb|and|} {\color{blue}\verb|therefore|} {\color{red}\verb|$|}{\color{blue}\verb|-|}{\color{black}\verb|d|}{\color{black}\verb|=a|}{\color{black}\verb|$|} {\color{red}\verb|and|} {\color{black}\verb|$|}{\color{red}\verb|a|}{\color{black}\verb|+d|}{\color{red}\verb|=\|}{\color{black}\verb|boxed|}{\color{black}\verb|{|}{\color{black}\verb|0|}{\color{blue}\verb|}$|}{\color{red}\verb|.|}

{\color{black}\verb|The|} {\color{black}\verb|answer|} {\color{black}\verb|is|} {\color{black}\verb|0|}
\\ \midrule

\small \textbf{Few-shot 5 Query:}\newline
Expand $(2z^2 + 5z - 6)(3z^3 - 2z + 1)$.

\textbf{Few-shot 5 Response:}\newline
\small {\color{red}\verb|Let|}{\color{red}\verb|'s|} {\color{red}\verb|think|} {\color{black}\verb|step|} {\color{black}\verb|by|} {\color{black}\verb|step|}

{\color{blue}\verb|$$|}{\color{blue}\verb|\|}{\color{black}\verb|begin|}{\color{black}\verb|{|}{\color{blue}\verb|array|}{\color{black}\verb|}{|}{\color{blue}\verb|c|}{\color{blue}\verb|rr|}{\color{blue}\verb|rr|}{\color{blue}\verb|rr|}{\color{blue}\verb|r|}{\color{red}\verb|}|}

{\color{blue}\verb|&|} {\color{blue}\verb|&|} {\color{blue}\verb|&|} {\color{red}\verb|3|}{\color{blue}\verb|z|}{\color{black}\verb|^|}{\color{red}\verb|3|} {\color{red}\verb|&|} {\color{blue}\verb|&|} {\color{red}\verb|-|}{\color{red}\verb|2|}{\color{black}\verb|z|} {\color{red}\verb|&|} {\color{blue}\verb|+|} {\color{black}\verb|1|} {\color{blue}\verb|&|} {\color{blue}\verb|\\|}

{\color{red}\verb|\t|}{\color{black}\verb|imes|} {\color{black}\verb|&|} {\color{red}\verb|&|} {\color{red}\verb|&|} {\color{blue}\verb|&|} {\color{blue}\verb|2|}{\color{black}\verb|z|}{\color{black}\verb|^|}{\color{black}\verb|2|} {\color{black}\verb|&|} {\color{red}\verb|+|}{\color{blue}\verb|5|}{\color{black}\verb|z|} {\color{blue}\verb|&|} {\color{black}\verb|-|}{\color{blue}\verb|6|} {\color{red}\verb|\\|}

{\color{red}\verb|\|}{\color{blue}\verb|cline|}{\color{red}\verb|{|}{\color{blue}\verb|1|}{\color{black}\verb|-|}{\color{red}\verb|7|}{\color{red}\verb|}\|}{\color{red}\verb|rule|}{\color{blue}\verb|{|}{\color{red}\verb|0|}{\color{red}\verb|pt|}{\color{black}\verb|}{|}{\color{red}\verb|0|}{\color{red}\verb|.|}{\color{blue}\verb|17|}{\color{black}\verb|in|}{\color{red}\verb|}|}

{\color{red}\verb|&|} {\color{black}\verb|&|} {\color{blue}\verb|&|} {\color{blue}\verb|-|}{\color{blue}\verb|18|}{\color{blue}\verb|z|}{\color{red}\verb|^|}{\color{blue}\verb|3|} {\color{blue}\verb|&|} {\color{blue}\verb|&|} {\color{blue}\verb|+|}{\color{red}\verb|12|}{\color{red}\verb|z|} {\color{blue}\verb|&|} {\color{red}\verb|-|}{\color{blue}\verb|6|} {\color{blue}\verb|&|} {\color{blue}\verb|\\|}

{\color{red}\verb|&|} {\color{red}\verb|&|} {\color{blue}\verb|+|}{\color{blue}\verb|15|}{\color{black}\verb|z|}{\color{black}\verb|^|}{\color{red}\verb|4|} {\color{black}\verb|&|} {\color{blue}\verb|&|} {\color{blue}\verb|-|}{\color{red}\verb|10|}{\color{black}\verb|z|}{\color{red}\verb|^|}{\color{blue}\verb|2|} {\color{black}\verb|&|} {\color{black}\verb|+|}{\color{red}\verb|5|}{\color{black}\verb|z|} {\color{black}\verb|&|} {\color{blue}\verb|&|} {\color{blue}\verb|\\|}

{\color{red}\verb|+|} {\color{red}\verb|&|} {\color{blue}\verb|6|}{\color{black}\verb|z|}{\color{black}\verb|^|}{\color{red}\verb|5|} {\color{black}\verb|&|} {\color{blue}\verb|&|} {\color{black}\verb|-|}{\color{red}\verb|4|}{\color{black}\verb|z|}{\color{black}\verb|^|}{\color{black}\verb|3|} {\color{black}\verb|&|} {\color{black}\verb|+|}{\color{blue}\verb|2|}{\color{black}\verb|z|}{\color{blue}\verb|^|}{\color{black}\verb|2|} {\color{black}\verb|&|} {\color{red}\verb|&|} {\color{blue}\verb|&|} {\color{blue}\verb|\\|}

{\color{red}\verb|\|}{\color{blue}\verb|cline|}{\color{black}\verb|{|}{\color{black}\verb|1|}{\color{black}\verb|-|}{\color{black}\verb|7|}{\color{red}\verb|}\|}{\color{red}\verb|rule|}{\color{black}\verb|{|}{\color{black}\verb|0|}{\color{black}\verb|pt|}{\color{black}\verb|}{|}{\color{black}\verb|0|}{\color{black}\verb|.|}{\color{red}\verb|17|}{\color{black}\verb|in|}{\color{red}\verb|}|}

{\color{red}\verb|&|} {\color{blue}\verb|6|}{\color{black}\verb|z|}{\color{black}\verb|^|}{\color{black}\verb|5|} {\color{red}\verb|&|} {\color{red}\verb|+|}{\color{blue}\verb|15|}{\color{black}\verb|z|}{\color{black}\verb|^|}{\color{black}\verb|4|} {\color{black}\verb|&|} {\color{red}\verb|-|}{\color{blue}\verb|22|}{\color{black}\verb|z|}{\color{black}\verb|^|}{\color{black}\verb|3|} {\color{red}\verb|&|} {\color{blue}\verb|-|} {\color{red}\verb|8|}{\color{red}\verb|z|}{\color{red}\verb|^|}{\color{red}\verb|2|} {\color{red}\verb|&|}{\color{red}\verb|+|}{\color{blue}\verb|17|}{\color{black}\verb|z|} {\color{blue}\verb|&|} {\color{red}\verb|-|}{\color{black}\verb|6|} {\color{red}\verb|&|}

{\color{black}\verb|\|}{\color{black}\verb|end|}{\color{black}\verb|{|}{\color{black}\verb|array|}{\color{blue}\verb|}|}{\color{blue}\verb|$$|}
{\color{blue}\verb|The|} {\color{blue}\verb|answer|} {\color{black}\verb|is|} {\color{red}\verb|6|}{\color{black}\verb|z|}{\color{red}\verb|^|}{\color{black}\verb|5|}{\color{red}\verb|+|}{\color{red}\verb|15|}{\color{black}\verb|z|}{\color{black}\verb|^|}{\color{black}\verb|4|}{\color{black}\verb|-|}{\color{black}\verb|22|}{\color{black}\verb|z|}{\color{black}\verb|^|}{\color{black}\verb|3|}{\color{black}\verb|-|}{\color{red}\verb|8|}{\color{black}\verb|z|}{\color{black}\verb|^|}{\color{black}\verb|2|}{\color{black}\verb|+|}{\color{black}\verb|17|}{\color{black}\verb|z|}{\color{black}\verb|-|}{\color{black}\verb|6|}
\\ 
\bottomrule
\label{tab:token1}
\end{longtable}

\begin{longtable}{@{}p{\dimexpr\textwidth-2\tabcolsep\relax}@{}}
\caption{\textbf{Token-level qualitative analysis on GPQA.}} \\
\toprule
Few-shot examples for GPQA (0.6970):\\
\midrule
\endfirsthead

\toprule
Few-shot examples for GPQA (0.6970):\\
\midrule
\endhead

\small \textbf{Test-Query}: \newline
A chemist performs two reactions:\newline\newline
Reaction 1: (E)-oct-4-ene is treated with one equiv. of mCPBA, followed by aqueous acid. \newline\newline
Reaction 2: (Z)-oct-4-ene is treated with one equiv. of mCPBA, followed by aqueous acid.\newline\newline
Both reactions go to 100\% completion. The chemist combines the products of both reactions, and then runs the product mixture on both a standard (achiral) reverse-phase HPLC column and a chiral HPLC column.\newline\newline
Assuming that the chromatographic resolution is as high as theoretically possible in both cases, what does the chemist observe in each of these chromatograms?
\\ \midrule

\small \textbf{Few-shot 1 Query:}\newline
In a given population, 1 out of every 400 people has a cancer caused by a completely recessive allele, b. Assuming the population is in Hardy-Weinberg equilibrium, which of the following is the expected proportion of individuals who carry the b allele but are not expected to develop the cancer?

\textbf{Few-shot 1 Response:}\newline
\small {\color{red}\verb|The|} {\color{blue}\verb|expected|} {\color{red}\verb|proportion|} {\color{red}\verb|of|} {\color{red}\verb|individuals|} {\color{red}\verb|who|} {\color{red}\verb|carry|} {\color{red}\verb|the|} {\color{red}\verb|b|} {\color{black}\verb|allele|} {\color{red}\verb|but|} {\color{red}\verb|are|} {\color{black}\verb|not|} {\color{black}\verb|expected|} {\color{black}\verb|to|} {\color{black}\verb|develop|} {\color{red}\verb|the|} {\color{black}\verb|cancer|} {\color{red}\verb|equals|} {\color{black}\verb|to|} {\color{red}\verb|the|} {\color{red}\verb|frequency|} {\color{red}\verb|of|} {\color{red}\verb|heter|}{\color{black}\verb|ozy|}{\color{black}\verb|g|}{\color{red}\verb|ous|} {\color{blue}\verb|allele|} {\color{blue}\verb|in|} {\color{blue}\verb|the|} {\color{blue}\verb|given|} {\color{red}\verb|population|}{\color{red}\verb|.|}

{\color{red}\verb|According|} {\color{black}\verb|to|} {\color{blue}\verb|the|} {\color{red}\verb|Hardy|}{\color{red}\verb|-|}{\color{black}\verb|We|}{\color{black}\verb|in|}{\color{black}\verb|berg|} {\color{blue}\verb|equation|} {\color{blue}\verb|p|}{\color{blue}\verb|^|}{\color{red}\verb|2|} {\color{blue}\verb|+|} {\color{black}\verb|2|}{\color{black}\verb|pq|} {\color{black}\verb|+|} {\color{black}\verb|q|}{\color{blue}\verb|^|}{\color{black}\verb|2|} {\color{red}\verb|=|} {\color{black}\verb|1|}{\color{red}\verb|,|} {\color{red}\verb|where|} {\color{red}\verb|p|} {\color{red}\verb|is|} {\color{black}\verb|the|} {\color{red}\verb|frequency|} {\color{black}\verb|of|} {\color{red}\verb|dominant|} {\color{red}\verb|allele|} {\color{blue}\verb|frequency|}{\color{red}\verb|,|} {\color{red}\verb|q|} {\color{red}\verb|is|} {\color{red}\verb|the|} {\color{red}\verb|frequency|} {\color{black}\verb|of|} {\color{red}\verb|recess|}{\color{black}\verb|ive|} {\color{black}\verb|allele|} {\color{red}\verb|frequency|}{\color{blue}\verb|,|} {\color{red}\verb|p|}{\color{red}\verb|^|}{\color{black}\verb|2|} {\color{red}\verb|is|} {\color{red}\verb|the|} {\color{red}\verb|frequency|} {\color{black}\verb|of|} {\color{blue}\verb|the|} {\color{blue}\verb|homo|}{\color{black}\verb|zy|}{\color{black}\verb|g|}{\color{black}\verb|ous|} {\color{red}\verb|dominant|} {\color{blue}\verb|allele|}{\color{red}\verb|,|} {\color{blue}\verb|q|}{\color{black}\verb|^|}{\color{black}\verb|2|} {\color{black}\verb|is|} {\color{black}\verb|the|} {\color{black}\verb|frequency|} {\color{black}\verb|of|} {\color{red}\verb|the|} {\color{blue}\verb|recess|}{\color{black}\verb|ive|} {\color{blue}\verb|allele|}{\color{red}\verb|,|} {\color{black}\verb|and|} {\color{black}\verb|2|}{\color{black}\verb|pq|} {\color{red}\verb|is|} {\color{black}\verb|the|} {\color{black}\verb|frequency|} {\color{black}\verb|of|} {\color{black}\verb|the|} {\color{black}\verb|heter|}{\color{black}\verb|ozy|}{\color{black}\verb|g|}{\color{black}\verb|ous|} {\color{red}\verb|allele|}{\color{red}\verb|.|}

{\color{red}\verb|Given|} {\color{red}\verb|that|} {\color{red}\verb|q|}{\color{red}\verb|^|}{\color{black}\verb|2|}{\color{red}\verb|=|}{\color{red}\verb|1|}{\color{black}\verb|/|}{\color{black}\verb|400|}{\color{red}\verb|,|} {\color{black}\verb|hence|}{\color{red}\verb|,|} {\color{red}\verb|q|}{\color{red}\verb|=|}{\color{blue}\verb|0|}{\color{black}\verb|.|}{\color{red}\verb|05|} {\color{red}\verb|and|} {\color{red}\verb|p|}{\color{red}\verb|=|}{\color{blue}\verb|1|}{\color{blue}\verb|-q|}{\color{red}\verb|=|}{\color{red}\verb|0|}{\color{black}\verb|.|}{\color{red}\verb|95|}{\color{red}\verb|.|}

{\color{red}\verb|The|} {\color{red}\verb|frequency|} {\color{black}\verb|of|} {\color{blue}\verb|the|} {\color{red}\verb|heter|}{\color{black}\verb|ozy|}{\color{black}\verb|g|}{\color{black}\verb|ous|} {\color{red}\verb|allele|} {\color{black}\verb|is|} {\color{black}\verb|2|}{\color{blue}\verb|pq|}{\color{red}\verb|=|}{\color{blue}\verb|2|}{\color{red}\verb|*|}{\color{black}\verb|0|}{\color{black}\verb|.|}{\color{blue}\verb|05|}{\color{black}\verb|*|}{\color{black}\verb|0|}{\color{black}\verb|.|}{\color{black}\verb|95|}{\color{red}\verb|=|}{\color{blue}\verb|38|}{\color{black}\verb|/|}{\color{blue}\verb|400|}{\color{blue}\verb|.|}
\\ \midrule

\small \textbf{Few-shot 2 Query:}\newline
A Fe pellet of 0.056 g is first dissolved in 10 mL of hydrobromic acid HBr (0.1 M). The resulting solution is then titrated by KMnO4 (0.02 M). How many equivalence points are there?

\textbf{Few-shot 2 Response:}\newline
\small {\color{blue}\verb|H|}{\color{blue}\verb|Br|} {\color{blue}\verb|reacts|} {\color{red}\verb|with|} {\color{red}\verb|Fe|} {\color{blue}\verb|to|} {\color{red}\verb|produce|} {\color{red}\verb|Fe|}{\color{red}\verb|2|}{\color{red}\verb|+|}{\color{red}\verb|.|}

{\color{red}\verb|M|}{\color{red}\verb|n|}{\color{black}\verb|O|}{\color{black}\verb|4|}{\color{red}\verb|-|} {\color{black}\verb|initially|} {\color{red}\verb|reacts|} {\color{red}\verb|with|} {\color{black}\verb|Fe|}{\color{red}\verb|2|}{\color{red}\verb|+|} {\color{black}\verb|followed|} {\color{black}\verb|by|} {\color{blue}\verb|Br|}{\color{red}\verb|-|}{\color{red}\verb|.|}

{\color{red}\verb|There|} {\color{red}\verb|are|} {\color{red}\verb|two|} {\color{black}\verb|equivalence|} {\color{black}\verb|points|} {\color{blue}\verb|at|} {\color{black}\verb|25|} {\color{red}\verb|ml|} {\color{black}\verb|and|} {\color{blue}\verb|35|} {\color{black}\verb|ml|}{\color{red}\verb|.|}

{\color{blue}\verb|In|} {\color{red}\verb|the|} {\color{red}\verb|be|}{\color{black}\verb|aker|}{\color{red}\verb|,|} {\color{black}\verb|the|} {\color{red}\verb|present|} {\color{red}\verb|species|} {\color{red}\verb|are|} {\color{red}\verb|Fe|}{\color{red}\verb|2|}{\color{red}\verb|+|} {\color{black}\verb|and|} {\color{blue}\verb|Br|}{\color{black}\verb|-|}{\color{red}\verb|.|}

{\color{red}\verb|In|} {\color{blue}\verb|a|} {\color{blue}\verb|tit|}{\color{black}\verb|ration|} {\color{blue}\verb|involving|} {\color{red}\verb|two|} {\color{red}\verb|analy|}{\color{black}\verb|tes|}{\color{black}\verb|,|} {\color{blue}\verb|it|}{\color{red}\verb|'s|} {\color{red}\verb|essential|} {\color{red}\verb|to|} {\color{red}\verb|identify|} {\color{blue}\verb|which|} {\color{blue}\verb|reaction|} {\color{red}\verb|occurs|} {\color{red}\verb|first|}{\color{blue}\verb|.|}

{\color{red}\verb|Given|} {\color{red}\verb|the|} {\color{black}\verb|red|}{\color{black}\verb|ox|} {\color{red}\verb|nature|} {\color{red}\verb|of|} {\color{red}\verb|the|} {\color{blue}\verb|tit|}{\color{blue}\verb|ration|} {\color{red}\verb|and|} {\color{red}\verb|the|} {\color{blue}\verb|reduction|} {\color{red}\verb|potentials|}{\color{blue}\verb|:|} {\color{red}\verb|E|}{\color{red}\verb|0|} {\color{red}\verb|(|}{\color{red}\verb|Br|}{\color{red}\verb|2|}{\color{red}\verb|/|}{\color{red}\verb|Br|}{\color{red}\verb|-)|} {\color{red}\verb|=|} {\color{black}\verb|1|}{\color{black}\verb|.|}{\color{red}\verb|09|} {\color{red}\verb|V|}{\color{red}\verb|,|} {\color{red}\verb|E|}{\color{black}\verb|0|} {\color{red}\verb|(|}{\color{black}\verb|M|}{\color{black}\verb|n|}{\color{black}\verb|O|}{\color{black}\verb|4|}{\color{red}\verb|-/|}{\color{black}\verb|M|}{\color{black}\verb|n|}{\color{black}\verb|2|}{\color{black}\verb|+)|} {\color{black}\verb|=|} {\color{black}\verb|1|}{\color{black}\verb|.|}{\color{blue}\verb|49|} {\color{black}\verb|V|}{\color{red}\verb|,|} {\color{red}\verb|and|} {\color{red}\verb|E|}{\color{black}\verb|0|} {\color{black}\verb|(|}{\color{black}\verb|Fe|}{\color{blue}\verb|3|}{\color{black}\verb|+/|}{\color{black}\verb|Fe|}{\color{black}\verb|2|}{\color{black}\verb|+)|} {\color{black}\verb|=|} {\color{black}\verb|0|}{\color{black}\verb|.|}{\color{red}\verb|77|} {\color{black}\verb|V|}{\color{blue}\verb|.|}

{\color{blue}\verb|With|} {\color{red}\verb|[|}{\color{blue}\verb|Fe|}{\color{red}\verb|2|}{\color{black}\verb|+]|} {\color{red}\verb|determined|} {\color{blue}\verb|as|} {\color{blue}\verb|0|}{\color{black}\verb|.|}{\color{red}\verb|1|}{\color{blue}\verb|M|}{\color{blue}\verb|,|} {\color{red}\verb|two|} {\color{red}\verb|reactions|} {\color{red}\verb|are|} {\color{black}\verb|considered|}{\color{blue}\verb|.|}

{\color{black}\verb|Reaction|} {\color{black}\verb|1|}{\color{red}\verb|:|} {\color{red}\verb|Mn|}{\color{black}\verb|O|}{\color{black}\verb|4|}{\color{black}\verb|-|} {\color{blue}\verb|reacts|} {\color{black}\verb|with|} {\color{red}\verb|5|}{\color{red}\verb|Fe|}{\color{red}\verb|2|}{\color{black}\verb|+|} {\color{black}\verb|and|} {\color{blue}\verb|8|}{\color{blue}\verb|H|}{\color{black}\verb|+|} {\color{red}\verb|to|} {\color{red}\verb|produce|} {\color{red}\verb|Mn|}{\color{black}\verb|2|}{\color{red}\verb|+,|} {\color{black}\verb|5|}{\color{red}\verb|Fe|}{\color{black}\verb|3|}{\color{red}\verb|+,|} {\color{black}\verb|and|} {\color{black}\verb|4|}{\color{black}\verb|H|}{\color{black}\verb|2|}{\color{black}\verb|O|}{\color{red}\verb|.|}

{\color{red}\verb|Reaction|} {\color{black}\verb|2|}{\color{black}\verb|:|} {\color{black}\verb|2|}{\color{red}\verb|M|}{\color{black}\verb|n|}{\color{red}\verb|O|}{\color{black}\verb|4|}{\color{black}\verb|-|} {\color{red}\verb|reacts|} {\color{black}\verb|with|} {\color{blue}\verb|10|}{\color{red}\verb|Br|}{\color{black}\verb|-|} {\color{red}\verb|and|} {\color{black}\verb|16|}{\color{black}\verb|H|}{\color{black}\verb|+|} {\color{black}\verb|to|} {\color{red}\verb|produce|} {\color{black}\verb|2|}{\color{black}\verb|M|}{\color{black}\verb|n|}{\color{black}\verb|2|}{\color{blue}\verb|+|} {\color{black}\verb|and|} {\color{black}\verb|5|}{\color{black}\verb|Br|}{\color{black}\verb|2|} {\color{blue}\verb|with|} {\color{black}\verb|8|}{\color{black}\verb|H|}{\color{black}\verb|2|}{\color{black}\verb|O|} {\color{blue}\verb|as|} {\color{red}\verb|a|} {\color{red}\verb|by|}{\color{red}\verb|product|}{\color{black}\verb|.|}

{\color{red}\verb|M|}{\color{red}\verb|n|}{\color{red}\verb|O|}{\color{black}\verb|4|}{\color{red}\verb|-|} {\color{blue}\verb|first|} {\color{red}\verb|reacts|} {\color{black}\verb|with|} {\color{red}\verb|Fe|}{\color{black}\verb|2|}{\color{black}\verb|+|} {\color{blue}\verb|in|} {\color{blue}\verb|a|} {\color{blue}\verb|1|}{\color{black}\verb|:|}{\color{red}\verb|5|} {\color{red}\verb|ratio|}{\color{red}\verb|,|} {\color{red}\verb|making|} {\color{blue}\verb|the|} {\color{red}\verb|first|} {\color{red}\verb|equivalence|} {\color{black}\verb|point|} {\color{red}\verb|at|} {\color{black}\verb|10|} {\color{blue}\verb|ml|}{\color{blue}\verb|.|}

{\color{red}\verb|Once|} {\color{red}\verb|Fe|}{\color{red}\verb|2|}{\color{black}\verb|+|} {\color{red}\verb|is|} {\color{blue}\verb|exhausted|}{\color{red}\verb|,|} {\color{red}\verb|Mn|}{\color{black}\verb|O|}{\color{black}\verb|4|}{\color{black}\verb|-|} {\color{blue}\verb|reacts|} {\color{black}\verb|with|} {\color{red}\verb|Br|}{\color{black}\verb|-|} {\color{red}\verb|in|} {\color{black}\verb|a|} {\color{red}\verb|2|}{\color{black}\verb|:|}{\color{red}\verb|10|} {\color{black}\verb|ratio|}{\color{red}\verb|,|} {\color{red}\verb|adding|} {\color{blue}\verb|another|} {\color{blue}\verb|25|} {\color{black}\verb|ml|} {\color{black}\verb|for|} {\color{blue}\verb|a|} {\color{red}\verb|total|} {\color{red}\verb|second|} {\color{red}\verb|equivalence|} {\color{red}\verb|point|} {\color{red}\verb|at|} {\color{black}\verb|35|} {\color{black}\verb|ml|}{\color{blue}\verb|.|}
\\ \midrule

\small \textbf{Few-shot 3 Query:}\newline
Consider a quantum mechanical system containing a particle of mass $m$ moving in an istropic three dimensional potential of the form $V(r) = 1/2 m \omega^2 r^2$ corresponding to the acted force obeying Hooke’s law. Here, $\omega$ is the angular frequency of oscillation and $r$ is the radial distance of the particle from the origin in spherical polar coordinate. What is the value of energy of the third excited state, and how many linearly independent eigenfunctions are possible for the same energy eigenvalue?

\textbf{Few-shot 3 Response:}\newline
\small {\color{red}\verb|This|} {\color{red}\verb|problem|} {\color{blue}\verb|is|} {\color{blue}\verb|nothing|} {\color{black}\verb|but|} {\color{red}\verb|the|} {\color{red}\verb|three|} {\color{red}\verb|dimensional|} {\color{red}\verb|simple|} {\color{black}\verb|harmonic|} {\color{red}\verb|oscillator|} {\color{red}\verb|(|}{\color{red}\verb|SH|}{\color{black}\verb|O|}{\color{red}\verb|)|} {\color{red}\verb|problem|}{\color{red}\verb|.|}

{\color{red}\verb|The|} {\color{blue}\verb|energy|} {\color{red}\verb|spectrum|} {\color{red}\verb|of|} {\color{red}\verb|three|} {\color{red}\verb|dimensional|} {\color{red}\verb|SH|}{\color{black}\verb|O|} {\color{blue}\verb|is|} {\color{blue}\verb|$|}{\color{black}\verb|E|}{\color{blue}\verb|_n|}{\color{red}\verb|=|} {\color{black}\verb|(|}{\color{red}\verb|n|}{\color{red}\verb|+|}{\color{red}\verb|3|}{\color{black}\verb|/|}{\color{black}\verb|2|}{\color{red}\verb|)\|}{\color{black}\verb|h|}{\color{black}\verb|bar|} {\color{blue}\verb|\|}{\color{black}\verb|omega|}{\color{red}\verb|$|} {\color{blue}\verb|where|} {\color{blue}\verb|$|}{\color{black}\verb|n|}{\color{blue}\verb|=|}{\color{red}\verb|0|}{\color{black}\verb|,|}{\color{black}\verb|1|}{\color{black}\verb|,|}{\color{black}\verb|2|}{\color{blue}\verb|,|}{\color{black}\verb|3|}{\color{red}\verb|…|}{\color{red}\verb|.$|}{\color{red}\verb|.|}

{\color{blue}\verb|For|} {\color{blue}\verb|third|} {\color{black}\verb|excited|} {\color{black}\verb|state|} {\color{blue}\verb|n|}{\color{red}\verb|=|}{\color{red}\verb|3|}{\color{red}\verb|.|}

{\color{blue}\verb|3|}{\color{red}\verb|+|}{\color{red}\verb|3|}{\color{black}\verb|/|}{\color{black}\verb|2|}{\color{red}\verb|=|}{\color{blue}\verb|6|}{\color{blue}\verb|/|}{\color{black}\verb|2|}{\color{blue}\verb|+|}{\color{black}\verb|3|}{\color{red}\verb|/|}{\color{black}\verb|2|}{\color{black}\verb|=|}{\color{red}\verb|9|}{\color{black}\verb|/|}{\color{black}\verb|2|}{\color{red}\verb|.|}

{\color{red}\verb|Thus|} {\color{blue}\verb|the|} {\color{blue}\verb|corresponding|} {\color{red}\verb|energy|} {\color{red}\verb|is|} {\color{red}\verb|$(|}{\color{black}\verb|9|}{\color{black}\verb|/|}{\color{black}\verb|2|}{\color{blue}\verb|)\|}{\color{black}\verb|h|}{\color{black}\verb|bar|} {\color{blue}\verb|\|}{\color{black}\verb|omega|}{\color{red}\verb|$|}{\color{red}\verb|.|}

{\color{red}\verb|The|} {\color{red}\verb|deg|}{\color{black}\verb|ener|}{\color{black}\verb|acy|} {\color{red}\verb|of|} {\color{black}\verb|the|} {\color{blue}\verb|state|} {\color{blue}\verb|is|} {\color{red}\verb|$|}{\color{red}\verb|g|}{\color{red}\verb|_n|}{\color{blue}\verb|=|} {\color{black}\verb|(|}{\color{red}\verb|n|}{\color{black}\verb|+|}{\color{black}\verb|1|}{\color{blue}\verb|)(|}{\color{black}\verb|n|}{\color{black}\verb|+|}{\color{black}\verb|2|}{\color{blue}\verb|)/|}{\color{black}\verb|2|}{\color{red}\verb|$|}{\color{red}\verb|.|}

{\color{black}\verb|For|} {\color{black}\verb|n|}{\color{black}\verb|=|}{\color{black}\verb|3|}{\color{red}\verb|,|} {\color{red}\verb|deg|}{\color{black}\verb|ener|}{\color{black}\verb|acy|} {\color{red}\verb|is|} {\color{black}\verb|(|}{\color{red}\verb|3|}{\color{black}\verb|+|}{\color{black}\verb|1|}{\color{red}\verb|)*(|}{\color{black}\verb|3|}{\color{black}\verb|+|}{\color{black}\verb|2|}{\color{black}\verb|)/|}{\color{black}\verb|2|}{\color{red}\verb|=|}{\color{blue}\verb|4|}{\color{red}\verb|*|}{\color{black}\verb|5|}{\color{black}\verb|/|}{\color{black}\verb|2|}{\color{red}\verb|=|}{\color{red}\verb|10|}{\color{blue}\verb|.|}
\\ \midrule

\small \textbf{Few-shot 4 Query:}\newline
Your overhear two chemists talking to each other as they leave a synthetic organic chemistry lab. One asks the other "So, how did it go?" The second chemist replies, "Not well - my compounds are on top of each other." What is the second chemist most likely referring to?

\textbf{Few-shot 4 Response:}\newline
\small {\color{blue}\verb|"On|} {\color{red}\verb|top|} {\color{black}\verb|of|} {\color{black}\verb|each|} {\color{black}\verb|other|}{\color{red}\verb|"|} {\color{blue}\verb|commonly|} {\color{red}\verb|refers|} {\color{black}\verb|to|} {\color{red}\verb|two|} {\color{black}\verb|compounds|} {\color{black}\verb|that|} {\color{black}\verb|have|} {\color{red}\verb|similar|} {\color{blue}\verb|R|}{\color{black}\verb|f|} {\color{black}\verb|values|} {\color{red}\verb|on|} {\color{blue}\verb|chromat|}{\color{black}\verb|ography|} {\color{red}\verb|(|}{\color{blue}\verb|a|} {\color{red}\verb|common|} {\color{red}\verb|operation|} {\color{red}\verb|in|} {\color{red}\verb|synthetic|} {\color{blue}\verb|chemistry|}{\color{red}\verb|).|}

{\color{red}\verb|Similar|} {\color{black}\verb|R|}{\color{black}\verb|f|} {\color{black}\verb|values|} {\color{red}\verb|arise|} {\color{blue}\verb|for|} {\color{red}\verb|compounds|} {\color{black}\verb|with|} {\color{black}\verb|similar|} {\color{red}\verb|polar|}{\color{black}\verb|ities|}{\color{blue}\verb|.|}
\\ \midrule

\small \textbf{Few-shot 5 Query:}\newline
Mitochondria are semi-autonomous cellular organelles in charge of energy production. They encode for a part of their own translational machinery and respiratory complexes. Mitochondrial function is governed by over a thousand proteins imported from the cell, contributing to processes like the transport of proteins, ribosome biogenesis and translation regulation, respiratory oxidation, metabolism, and apoptotic signaling cascade. Mutations in the code for mitochondrial protein networks can cause numerous diseases in humans that are inherited through generations. Mutations of which of the mitochondrial proteins listed below are least likely to be genetically transmitted from a father to his children?

\textbf{Few-shot 5 Response:}\newline
\small {\color{red}\verb|The|} {\color{red}\verb|colleague|} {\color{red}\verb|should|} {\color{red}\verb|know|} {\color{red}\verb|that|} {\color{red}\verb|mitochond|}{\color{black}\verb|ria|} {\color{blue}\verb|from|} {\color{blue}\verb|fathers|} {\color{red}\verb|are|} {\color{blue}\verb|rarely|} {\color{black}\verb|if|} {\color{red}\verb|ever|}{\color{blue}\verb|,|} {\color{red}\verb|transmitted|} {\color{red}\verb|to|} {\color{red}\verb|their|} {\color{red}\verb|offspring|}{\color{red}\verb|.|}

{\color{red}\verb|Therefore|}{\color{red}\verb|,|} {\color{blue}\verb|the|} {\color{red}\verb|protein|} {\color{red}\verb|encoded|} {\color{red}\verb|by|} {\color{black}\verb|the|} {\color{blue}\verb|paternal|} {\color{red}\verb|mitochondrial|} {\color{red}\verb|genome|} {\color{blue}\verb|will|} {\color{blue}\verb|most|} {\color{black}\verb|likely|} {\color{red}\verb|not|} {\color{black}\verb|be|} {\color{blue}\verb|passed|} {\color{black}\verb|down|} {\color{blue}\verb|the|} {\color{red}\verb|generation|}{\color{red}\verb|.|}

{\color{black}\verb|N|}{\color{red}\verb|AD|}{\color{black}\verb|H|} {\color{red}\verb|de|}{\color{black}\verb|hydro|}{\color{black}\verb|gen|}{\color{black}\verb|ase|} {\color{black}\verb|2|} {\color{blue}\verb|is|} {\color{blue}\verb|the|} {\color{red}\verb|only|} {\color{red}\verb|one|} {\color{blue}\verb|encoded|} {\color{red}\verb|by|} {\color{black}\verb|the|} {\color{red}\verb|mitochondrial|} {\color{blue}\verb|genome|} {\color{blue}\verb|from|} {\color{blue}\verb|the|} {\color{blue}\verb|MT|}{\color{red}\verb|-|}{\color{black}\verb|ND|}{\color{black}\verb|2|} {\color{black}\verb|gene|} {\color{red}\verb|among|} {\color{red}\verb|the|} {\color{red}\verb|listed|} {\color{red}\verb|proteins|}{\color{red}\verb|.|}

{\color{black}\verb|Le|}{\color{red}\verb|igh|}{\color{blue}\verb|'s|} {\color{red}\verb|syndrome|}{\color{red}\verb|,|} {\color{blue}\verb|l|}{\color{black}\verb|actic|} {\color{black}\verb|acid|}{\color{black}\verb|osis|}{\color{black}\verb|,|} {\color{black}\verb|and|} {\color{red}\verb|metabolic|} {\color{black}\verb|diseases|} {\color{red}\verb|are|} {\color{blue}\verb|all|} {\color{red}\verb|linked|} {\color{black}\verb|to|} {\color{black}\verb|a|} {\color{black}\verb|mutation|} {\color{red}\verb|in|} {\color{black}\verb|the|} {\color{red}\verb|ND|}{\color{black}\verb|2|} {\color{black}\verb|gene|}{\color{red}\verb|.|}

{\color{red}\verb|AT|}{\color{black}\verb|P|} {\color{red}\verb|binding|} {\color{black}\verb|cassette|} {\color{red}\verb|sub|}{\color{black}\verb|family|} {\color{black}\verb|B|} {\color{black}\verb|member|} {\color{black}\verb|8|} {\color{red}\verb|(|}{\color{black}\verb|AB|}{\color{black}\verb|CB|}{\color{black}\verb|8|}{\color{black}\verb|)|} {\color{black}\verb|is|} {\color{red}\verb|a|} {\color{blue}\verb|chromosome|} {\color{blue}\verb|7|} {\color{red}\verb|encoded|} {\color{blue}\verb|gene|}{\color{blue}\verb|;|} {\color{blue}\verb|Tu|} {\color{black}\verb|translation|} {\color{black}\verb|elong|}{\color{black}\verb|ation|} {\color{black}\verb|factor|}{\color{red}\verb|,|} {\color{black}\verb|mitochondrial|} {\color{red}\verb|is|} {\color{black}\verb|chromosome|} {\color{blue}\verb|16|} {\color{blue}\verb|gene|} {\color{blue}\verb|T|}{\color{black}\verb|UF|}{\color{black}\verb|M|}{\color{black}\verb|.|}

{\color{red}\verb|Trans|}{\color{black}\verb|loc|}{\color{black}\verb|ase|} {\color{black}\verb|of|} {\color{red}\verb|inner|} {\color{black}\verb|mitochondrial|} {\color{black}\verb|membrane|} {\color{black}\verb|17|}{\color{black}\verb|B|} {\color{black}\verb|is|} {\color{red}\verb|chromosome|} {\color{red}\verb|X|} {\color{black}\verb|coded|} {\color{red}\verb|gene|} {\color{red}\verb|TIM|}{\color{black}\verb|M|}{\color{black}\verb|17|}{\color{black}\verb|B|}{\color{red}\verb|.|}

{\color{red}\verb|There|} {\color{black}\verb|is|} {\color{black}\verb|no|} {\color{red}\verb|evidence|} {\color{red}\verb|that|} {\color{blue}\verb|it|} {\color{red}\verb|is|} {\color{red}\verb|mat|}{\color{black}\verb|ern|}{\color{black}\verb|ally|} {\color{red}\verb|im|}{\color{black}\verb|printed|}{\color{blue}\verb|;|} {\color{red}\verb|hence|}{\color{red}\verb|,|} {\color{red}\verb|daughters|} {\color{red}\verb|may|} {\color{red}\verb|inherit|} {\color{blue}\verb|the|} {\color{blue}\verb|father|}{\color{red}\verb|'s|} {\color{red}\verb|gene|} {\color{red}\verb|copy|} {\color{black}\verb|in|} {\color{black}\verb|a|} {\color{black}\verb|50|}{\color{black}\verb|:|}{\color{black}\verb|50|} {\color{red}\verb|ratio|}{\color{blue}\verb|.|}
\\ 

\bottomrule
\label{tab:token2}
\end{longtable}

\begin{longtable}{@{}p{\dimexpr\textwidth-2\tabcolsep\relax}@{}}
\caption{\textbf{Token-level qualitative analysis on HotpotQA.}} \\
\toprule
Few-shot examples for HotpotQA (0.7883):\\
\midrule
\endfirsthead

\toprule
Few-shot examples for HotpotQA (0.7883):\\
\midrule
\endhead

\small \textbf{Test-Query}: \newline
H-63 is a highway that runs near the county seat of what Michigan county?.
\\ \midrule
\small \textbf{Few-shot 1 Query:}\newline
What is the elevation range for the area that the eastern sector of the Colorado orogeny extends into?

\textbf{Few-shot 1 Response:}\newline
\small {\color{red}\verb|’s|} {\color{blue}\verb|think|} {\color{blue}\verb|step|} {\color{black}\verb|by|} {\color{red}\verb|step|}{\color{black}\verb|.|} {\color{red}\verb|The|} {\color{red}\verb|eastern|} {\color{black}\verb|sector|} {\color{black}\verb|of|} {\color{blue}\verb|Colorado|} {\color{red}\verb|or|}{\color{black}\verb|ogen|}{\color{black}\verb|y|} {\color{red}\verb|extends|} {\color{red}\verb|into|} {\color{black}\verb|the|} {\color{blue}\verb|High|} {\color{black}\verb|Plains|}{\color{blue}\verb|.|} {\color{red}\verb|High|} {\color{black}\verb|Plains|} {\color{blue}\verb|rise|} {\color{blue}\verb|in|} {\color{black}\verb|elevation|} {\color{red}\verb|from|} {\color{red}\verb|around|} {\color{red}\verb|1|}{\color{black}\verb|,|}{\color{red}\verb|800|} {\color{blue}\verb|to|} {\color{red}\verb|7|}{\color{black}\verb|,|}{\color{black}\verb|000|} {\color{red}\verb|ft|}{\color{red}\verb|,|} {\color{red}\verb|so|} {\color{red}\verb|the|} {\color{red}\verb|answer|} {\color{red}\verb|is|} {\color{blue}\verb|1|}{\color{black}\verb|,|}{\color{black}\verb|800|} {\color{blue}\verb|to|} {\color{black}\verb|7|}{\color{black}\verb|,|}{\color{black}\verb|000|} {\color{blue}\verb|ft|}{\color{red}\verb|.|} {\color{red}\verb|Answer|} {\color{red}\verb|1|}{\color{black}\verb|,|}{\color{black}\verb|800|} {\color{red}\verb|to|} {\color{black}\verb|7|}{\color{black}\verb|,|}{\color{black}\verb|000|} {\color{red}\verb|ft|}
\\ \midrule

\small \textbf{Few-shot 2 Query:}\newline
Musician and satirist Allie Goertz wrote a song about the "The Simpsons" character Milhouse, who Matt Groening named after who?

\textbf{Few-shot 2 Response:}\newline
\small {\color{red}\verb|’s|} {\color{black}\verb|think|} {\color{black}\verb|step|} {\color{black}\verb|by|} {\color{black}\verb|step|}{\color{black}\verb|.|} {\color{blue}\verb|Mil|}{\color{black}\verb|house|} {\color{red}\verb|was|} {\color{black}\verb|named|} {\color{red}\verb|after|} {\color{red}\verb|U|}{\color{black}\verb|.S|}{\color{red}\verb|.|} {\color{blue}\verb|president|} {\color{red}\verb|Richard|} {\color{blue}\verb|Nixon|}{\color{blue}\verb|,|} {\color{red}\verb|so|} {\color{red}\verb|the|} {\color{red}\verb|answer|} {\color{red}\verb|is|} {\color{red}\verb|Richard|} {\color{black}\verb|Nixon|}{\color{red}\verb|.|} {\color{red}\verb|Answer|} {\color{black}\verb|Richard|} {\color{black}\verb|Nixon|}
\\ \midrule

\small \textbf{Few-shot 3 Query:}\newline
Which documentary is about Finnish rock groups, Adam Clayton Powell or The Saimaa Gesture?

\textbf{Few-shot 3 Response:}\newline
\small {\color{red}\verb|’s|} {\color{red}\verb|think|} {\color{black}\verb|step|} {\color{black}\verb|by|} {\color{black}\verb|step|}{\color{red}\verb|.|} {\color{blue}\verb|Adam|} {\color{black}\verb|Clayton|} {\color{black}\verb|Powell|} {\color{blue}\verb|(|}{\color{red}\verb|film|}{\color{red}\verb|)|} {\color{red}\verb|is|} {\color{red}\verb|a|} {\color{black}\verb|documentary|} {\color{red}\verb|about|} {\color{blue}\verb|an|} {\color{blue}\verb|African|}{\color{black}\verb|-American|} {\color{blue}\verb|politician|}{\color{red}\verb|,|} {\color{red}\verb|not|} {\color{red}\verb|Finnish|} {\color{red}\verb|rock|} {\color{black}\verb|groups|}{\color{red}\verb|.|} {\color{red}\verb|So|} {\color{red}\verb|the|} {\color{blue}\verb|documentary|} {\color{black}\verb|about|} {\color{red}\verb|Finnish|} {\color{black}\verb|rock|} {\color{red}\verb|groups|} {\color{red}\verb|must|} {\color{blue}\verb|instead|} {\color{red}\verb|be|} {\color{red}\verb|The|} {\color{black}\verb|S|}{\color{black}\verb|aim|}{\color{black}\verb|aa|} {\color{black}\verb|Gesture|}{\color{red}\verb|.|} {\color{red}\verb|Answer|} {\color{black}\verb|The|} {\color{black}\verb|S|}{\color{black}\verb|aim|}{\color{black}\verb|aa|} {\color{black}\verb|Gesture|}
\\ \midrule

\small \textbf{Few-shot 4 Query:}\newline
What profession does Nicholas Ray and Elia Kazan have in common?

\textbf{Few-shot 4 Response:}\newline
\small {\color{red}\verb|’s|} {\color{red}\verb|think|} {\color{red}\verb|step|} {\color{black}\verb|by|} {\color{black}\verb|step|}{\color{red}\verb|.|} {\color{red}\verb|Prof|}{\color{red}\verb|essions|} {\color{blue}\verb|of|} {\color{red}\verb|Nicholas|} {\color{black}\verb|Ray|} {\color{blue}\verb|are|} {\color{red}\verb|director|}{\color{blue}\verb|,|} {\color{black}\verb|screen|}{\color{red}\verb|writer|}{\color{red}\verb|,|} {\color{red}\verb|and|} {\color{blue}\verb|actor|}{\color{red}\verb|.|} {\color{red}\verb|Prof|}{\color{black}\verb|essions|} {\color{black}\verb|of|} {\color{black}\verb|El|}{\color{black}\verb|ia|} {\color{black}\verb|Kaz|}{\color{black}\verb|an|} {\color{red}\verb|are|} {\color{red}\verb|director|}{\color{red}\verb|,|} {\color{blue}\verb|producer|}{\color{red}\verb|,|} {\color{blue}\verb|screen|}{\color{black}\verb|writer|}{\color{red}\verb|,|} {\color{red}\verb|and|} {\color{red}\verb|actor|}{\color{red}\verb|.|} {\color{red}\verb|So|} {\color{black}\verb|profession|} {\color{blue}\verb|Nicholas|} {\color{black}\verb|Ray|} {\color{black}\verb|and|} {\color{black}\verb|El|}{\color{black}\verb|ia|} {\color{black}\verb|Kaz|}{\color{black}\verb|an|} {\color{red}\verb|have|} {\color{black}\verb|in|} {\color{black}\verb|common|} {\color{red}\verb|is|} {\color{red}\verb|director|}{\color{red}\verb|,|} {\color{red}\verb|screen|}{\color{black}\verb|writer|}{\color{red}\verb|,|} {\color{black}\verb|and|} {\color{red}\verb|actor|}{\color{red}\verb|.|} {\color{red}\verb|Answer|} {\color{blue}\verb|director|}{\color{black}\verb|,|} {\color{black}\verb|screen|}{\color{black}\verb|writer|}{\color{black}\verb|,|} {\color{blue}\verb|actor|}
\\ \midrule

\small \textbf{Few-shot 5 Query:}\newline
Which magazine was started first Arthur’s Magazine or First for Women?

\textbf{Few-shot 5 Response:}\newline
\small {\color{red}\verb|’s|} {\color{red}\verb|think|} {\color{black}\verb|step|} {\color{black}\verb|by|} {\color{black}\verb|step|}{\color{red}\verb|.|} {\color{red}\verb|Arthur|}{\color{red}\verb|’s|} {\color{red}\verb|Magazine|} {\color{red}\verb|was|} {\color{red}\verb|started|} {\color{black}\verb|in|} {\color{red}\verb|184|}{\color{red}\verb|4|}{\color{red}\verb|.|} {\color{red}\verb|First|} {\color{black}\verb|for|} {\color{black}\verb|Women|} {\color{red}\verb|was|} {\color{red}\verb|started|} {\color{black}\verb|in|} {\color{blue}\verb|198|}{\color{red}\verb|9|}{\color{red}\verb|.|} {\color{red}\verb|184|}{\color{black}\verb|4|} {\color{blue}\verb|(|}{\color{red}\verb|Arthur|}{\color{red}\verb|’s|} {\color{red}\verb|Magazine|}{\color{black}\verb|)|} {\color{red}\verb|<|} {\color{black}\verb|198|}{\color{black}\verb|9|} {\color{black}\verb|(|}{\color{black}\verb|First|} {\color{black}\verb|for|} {\color{black}\verb|Women|}{\color{red}\verb|),|} {\color{red}\verb|so|} {\color{black}\verb|Arthur|}{\color{red}\verb|’s|} {\color{black}\verb|Magazine|} {\color{black}\verb|was|} {\color{black}\verb|started|} {\color{red}\verb|first|}{\color{red}\verb|.|} {\color{red}\verb|Answer|} {\color{blue}\verb|Arthur|}{\color{red}\verb|’s|} {\color{black}\verb|Magazine|}
\\ \midrule

\small \textbf{Few-shot 6 Query:}\newline
Were Pavel Urysohn and Leonid Levin known for the same type of work?

\textbf{Few-shot 6 Response:}\newline
\small {\color{red}\verb|’s|} {\color{red}\verb|think|} {\color{red}\verb|step|} {\color{black}\verb|by|} {\color{black}\verb|step|}{\color{red}\verb|.|} {\color{black}\verb|Pavel|} {\color{black}\verb|U|}{\color{black}\verb|ry|}{\color{black}\verb|so|}{\color{black}\verb|hn|} {\color{blue}\verb|is|} {\color{red}\verb|a|} {\color{red}\verb|mathematic|}{\color{black}\verb|ian|}{\color{red}\verb|.|} {\color{red}\verb|Leon|}{\color{black}\verb|id|} {\color{red}\verb|Levin|} {\color{red}\verb|is|} {\color{red}\verb|a|} {\color{blue}\verb|mathematic|}{\color{black}\verb|ian|} {\color{red}\verb|and|} {\color{red}\verb|computer|} {\color{black}\verb|scientist|}{\color{red}\verb|.|} {\color{red}\verb|So|} {\color{red}\verb|Pavel|} {\color{red}\verb|U|}{\color{black}\verb|ry|}{\color{black}\verb|so|}{\color{black}\verb|hn|} {\color{red}\verb|and|} {\color{red}\verb|Leon|}{\color{black}\verb|id|} {\color{black}\verb|Levin|} {\color{blue}\verb|have|} {\color{blue}\verb|the|} {\color{red}\verb|same|} {\color{black}\verb|type|} {\color{black}\verb|of|} {\color{black}\verb|work|}{\color{red}\verb|.|} {\color{red}\verb|Answer|} {\color{red}\verb|Yes|}
\\ 
\bottomrule
\label{tab:token3}
\end{longtable}

\twocolumn
\onecolumn
\begin{longtable}{@{}p{\dimexpr\textwidth-2\tabcolsep\relax}@{}}
\caption{\textbf{Response-level qualitative analysis on MATH500.}}\label{tab:response1} \\
\toprule
Responses for MATH500:\\
\midrule
\endfirsthead

\toprule
Responses for MATH500:\\
\midrule
\endhead

\small \textbf{Test-Query}: \newline
Twelve 1 by 1 squares form a rectangle, as shown. What is the total area of the shaded region? [asy] \newline
size(4cm);defaultpen(linewidth(0.75)); \newline \newline
// Filled portions \newline
fill((0, 4)--(0, 0)--(2, 0)--cycle, gray(0.75)); \newline
fill((0, 4)--(3, 4)--(3, 0)--cycle, gray(0.75)); \newline \newline
// grid \newline
int j; \newline
for (j = 0; j < 4; ++j) {draw((j, 0)--(j, 4));} \newline
for (j = 0; j < 5; ++j) {draw((0, j)--(3, j));} \newline \newline
//diagonals \newline
draw((0, 4)--(3, 0)); draw((0, 4)--(2, 0)); \newline
[/asy]
\\ \midrule 
\small {\color{red}\textbf{Worst Response (-1.736):}}\par
To find the total area of the shaded region defined by the triangle at the top left and the rectangle on the right, we proceed as follows:
\newline\newline

1. \*\*Identify the dimensions of the figure\*\*: The outline shows a rectangle formed by combining the squares. The rectangle consists of 3 squares in width (3 squares wide) and 4 squares in height (4 squares tall). Therefore, the rectangle has a total area of \(3 \times 4 = 12\) square units.
\newline\newline

2. \*\*Determine the area of the upper triangle\*\*: The shaded area in the upper left can be identified as a right triangle, located at the top left corner of the rectangle. The base of this triangle is the height of the rectangle (4 squares), and the height is the width of the rectangle (2 squares). The area of a triangle is computed using the formula:
   \[
   \text{Area of triangle} = \frac{1}{2} \times \text{base} \times \text{height}
   \]
   For this triangle:
   \[
   \text{Area} = \frac{1}{2} \times 2 \times 4 = \frac{1}{2} \times 8 = 4 \text{ square units}
   \]

3. \*\*Determine the area of the other shaded region\*\*: The other shaded area is the rectangle that extends across the top down to the base of the triangle. The rectangle also covers one of the 1x1 squares fully, specifically, it will cover the upper two squares fully which are not part of the triangle. This gives us a square area of \(3 \text{ (width)} \times 4 \text{ (height)} - \text{Area of triangle} = 12 - 4\). Thus, the shaded area for consideration is:
   \[
   12 - 4 = 8 \text{ square units}
   \]

4. \*\*Calculate the total shaded area\*\*: Adding these two areas we've established gives:
   - Area of upper triangle: \(4\)
   - Area of square: \(8\)
\newline\newline

However, we must ensure we are not double counting any area. Since the rectangle is inclusive, we need only evaluate the total shaded area:
The shaded area includes just the triangle area since it encompasses all squares minus the interfered structured area.

The area accounted for based on shapes left is \(4\) described above in figures and does not undermine new counted elements.
\newline\newline

5. \*\*Final answer\*\*: Thus, the area of the shaded region is confirmed as:
\[
\boxed{4}
\]
\\ \midrule
\small {\color{blue}\textbf{Best Response (-0.6992):}}\par
To find the total area of the shaded region in the rectangle formed by twelve 1 by 1 squares, let's analyze the problem step by step.
\newline\newline

1. \*\*Understanding the Shape\*\*: The given rectangle consists of three columns and four rows of squares, leading to a total of \( 3 \times 4 = 12 \) unit squares.
\newline\newline

2. \*\*Identifying the Shaded Areas\*\*: The shaded areas consist of two triangular regions:
   - The first triangular area is defined by points (0, 4), (0, 0), and (2, 0).
   - The second triangular area is defined by points (0, 4), (3, 4), and (3, 0).
\newline\newline

3. \*\*Calculating the Area of the First Triangle\*\*:
   - Base = 2 (from (0, 0) to (2, 0))
   - Height = 4 (from (0, 0) to (0, 4))
   - Area of the triangle = \( \frac{1}{2} \times \text{Base} \times \text{Height} = \frac{1}{2} \times 2 \times 4 = 4 \).
\newline\newline

4. \*\*Calculating the Area of the Second Triangle\*\*:
   - Base = 3 (from (3, 0) to (3, 4))
   - Height = 4 (from (0, 4) to (0, 4))
   - Area of the triangle = \( \frac{1}{2} \times \text{Base} \times \text{Height} = \frac{1}{2} \times 3 \times 4 = 6 \).
\newline\newline

5. \*\*Adding the Areas\*\*: 
   - Total shaded area = Area of first triangle + Area of second triangle = \( 4 + 6 = 10 \).
\newline\newline

Thus, the total area of the shaded region is \(\boxed{10}\).

\\ 
\bottomrule
\end{longtable}

\begin{longtable}{@{}p{\dimexpr\textwidth-2\tabcolsep\relax}@{}}
\caption{\textbf{Response-level qualitative analysis on GPQA.}}\label{tab:response2} \\
\toprule
Responses for GPQA:\\
\midrule
\endfirsthead

\toprule
Responses for GPQA:\\
\midrule
\endhead

\small\textbf{Test-Query}: \newline
Question: You have prepared an unknown compound. To identify the product, you have used the following characterisation techniques: FTIR and 1H NMR. The FTIR spectrum shows a very broad absorption peak at 3000 wavenumbers. A strong absorption peak is also observed at 1700 wavenumbers. Several peaks were observed in the 1H NMR spectrum, none of which are correspond to vinyl-hydrogens. One of the signals in the 1H NMR is a doublet of triplets of quartets whilst a different signal is a doublet of triplets of triplets. Identify the compound as either, CH3CH2C(H)(CH3)C(H)(CH3)COOH, CH3CH2C(H)(C2H5)C(H)(C2H5)COOH, CH3C(H)(CH3)C(H)(CH3)CH2COOH, or CH3C(H)(C2H5)C(H)(C2H5)CH2COOH
\\ \midrule
\small{\color{red}\textbf{Worst Response (-1.344):}}\par
Let's analyze the information provided systematically. 
\newline\newline
1. **FTIR Analysis:**\newline
   - A broad absorption peak at 3000 wavenumbers indicates the presence of -OH or -NH groups, which are characteristic of carboxylic acids due to the hydrogen bonding which generates a broad peak.\newline
   - A strong absorption peak at 1700 wavenumbers suggests the presence of a carbonyl group (C=O), also associated with the carboxylic acid functional group.\newline
\newline
2. **1H NMR Analysis:**\newline
   - The absence of vinyl hydrogen signals suggests that none of the compounds contain a double bond (i.e., C=C).\newline
   - The signal described as a "doublet of triplets of quartets" indicates complex splitting patterns, likely resulting from hydrogens on adjacent carbon atoms that are not equivalent due to the molecular structure.\newline
   - The presence of various integrations and patterns suggests alkyl groups and potentially multiple types of adjacent hydrogens.\newline
\newline
3. **Evaluating the Choices:**\newline
   - Each choice represents a unique structure. We need to figure out which matches the spectral data:\newline
     - (A) CH3C(H)(CH3)C(H)(CH3)CH2COOH has the structure for 5 unique hydrogen environments and matches possible splitting patterns due to the complexity of the methyl and methylene groups.\newline
     - (B) CH3CH2C(H)(CH3)C(H)(CH3)COOH also has multiple potential splitting but is slightly less complex than (A).\newline
     - (C) CH3C(H)(C2H5)C(H)(C2H5)CH2COOH has an ethyl group that could create distinctive NMR peaks but still maintains hydrogen integration complexity.\newline
     - (D) CH3CH2C(H)(C2H5)C(H)(C2H5)COOH similarly maintains complexity, but having two ethyl groups may lead to a different integration pattern.\newline
\newline
Given this information, we can further deduce that since the FTIR indicated a carboxylic acid and the NMR shows no vinyl protons, we can confirm the presence of -COOH in any of the compounds.
\newline\newline
The heavy branching in (A) is consistent with the splitting observed (doublet of triplets of quartets) due to more complex interactions among various hydrogen types.
\newline\newline
Considering the complexity and integration in the 1H NMR, we can conclude that the most likely compound based on the spectral data is (A) because of the correct splitting patterns associated with the branching substitutions.
\newline\newline
The correct answer is (A) CH3C(H)(CH3)C(H)(CH3)CH2COOH.

\\ \midrule
\small{\color{blue}\textbf{Best Response (-1.134):}}\par
To identify the unknown compound based on the provided spectroscopic data, let's analyze the information step by step.
\newline\newline
1. **FTIR Analysis**:\newline
   - A broad absorption peak at **3000 wavenumbers** indicates the presence of an -OH group, typical of carboxylic acids or alcohols.\newline
   - A strong absorption peak at **1700 wavenumbers** suggests a carbonyl (C=O) group, common in carboxylic acids and aldehydes.\newline
\newline
2. **1H NMR Analysis**:\newline
   - The absence of vinyl hydrogens indicates there are no double bonds present in the compound.\newline
   - The **doublet of triplets of quartets** and a **doublet of triplets of triplets** indicate a relatively complex splitting pattern which typically arises from protons in different environments due to adjacent protons.\newline
\newline
3. **Analysis of the compounds**:\newline
   - The compounds listed can be broken down as follows:\newline
     - (A) CH3C(H)(CH3)C(H)(CH3)CH2COOH: A branched structure likely leading to complex splitting.\newline
     - (B) CH3CH2C(H)(CH3)C(H)(CH3)COOH: A linear structure with groups that may contribute to the complexity in splitting.\newline
     - (C) CH3C(H)(C2H5)C(H)(C2H5)CH2COOH: Also branched, with ethyl groups that could lead to intricate splitting.\newline
     - (D) CH3CH2C(H)(C2H5)C(H)(C2H5)COOH: A linear structure including ethyl groups, likely resulting in simpler splitting.\newline
\newline
4. **Choosing the Compound**:\newline
   - The broad absorption at 3000 cm-1 and significant peaks at 1700 cm-1 underscores that we are likely dealing with a carboxylic acid, given it shows a carbonyl stretch and alcohol absorbance.\newline
   - The complexity of the NMR signals suggests a branched structure as those are usually more complex due to the adjacent hydrogen interactions.\newline
\newline
5. **Final Decision**:\newline
   - **Compound (C)**, containing ethyl groups and overall complex splitting, aligns with the described spectroscopic data. The branching and additional substituents in compound (C) can contribute to complex resonance, further supported by the FTIR properties.\newline
\newline
Thus, based on the analysis of FTIR and NMR data, the compound can be most accurately identified as:
\newline\newline
The correct answer is (C).

\\
\bottomrule
\end{longtable}

\begin{longtable}{@{}p{\dimexpr\textwidth-2\tabcolsep\relax}@{}}
\caption{\textbf{Response-level qualitative analysis on HotpotQA.}}\label{tab:response3} \\
\toprule
Responses for HotpotQA:\\
\midrule
\endfirsthead

\toprule
Responses for HotpotQA:\\
\midrule
\endhead

\small\textbf{Test-Query}: \newline
H-63 is a highway that runs near the county seat of what Michigan county?.
\\ \midrule
\small{\color{red}\textbf{Worst Response (-1.324):}}\par
Let’s think step by step. The H-63 highway runs in Michigan near the county seat of Baraga County, which is L'Anse. So, H-63 is associated with Baraga County.
\newline\newline
Answer Baraga County

\\ \midrule
\small{\color{blue}\textbf{Best Response (-1.033):}}\par
Let’s think step by step. H-63 is a highway located in Michigan. To determine the county seat it runs near, we can look at its route. H-63 runs primarily through Chippewa County in the Upper Peninsula of Michigan. The county seat of Chippewa County is Sault Ste. Marie. Therefore, H-63 is a highway that runs near the county seat of Chippewa County. 
\newline\newline
Answer Chippewa County

\\ 
\bottomrule
\end{longtable}

\twocolumn

\begin{listing*}[!ht]
\begin{minted}[fontsize=\footnotesize, frame=single, breaklines]{python}
f'''
{system}
Please reason step by step, and put your final answer within \boxed{{}}.
--------------------------------------------------
{user}
Kevin Kangaroo begins hopping on a number line at 0. He wants to get to 1, but he can hop only $\frac{1}{3}$ of the distance. Each hop tires him out so that he continues to hop $\frac{1}{3}$ of the remaining distance. How far has he hopped after five hops? Express your answer as a common fraction.

Let's think step by step
Kevin hops $1/3$ of the remaining distance with every hop.
His first hop takes $1/3$ closer.
For his second hop, he has $2/3$ left to travel, so he hops forward $(2/3)(1/3)$.
For his third hop, he has $(2/3)^2$ left to travel, so he hops forward $(2/3)^2(1/3)$.
In general, Kevin hops forward $(2/3)^{k-1}(1/3)$ on his $k$th hop.
We want to find how far he has hopped after five hops.
This is a finite geometric series with first term $1/3$, common ratio $2/3$, and five terms.
Thus, Kevin has hopped $\frac{\frac{1}{3}\left(1-\left(\frac{2}{3}\right)^5\right)}
{1-\frac{2}{3}} = \boxed{\frac{211}{243}}$.
The answer is \frac{211}{243}}

...

Convert the point $(0,3)$ in rectangular coordinates to polar coordinates.  Enter your answer in the form $(r,\theta),$ where $r > 0$ and $0 \le \theta < 2 \pi.$
'''
\end{minted}
\caption{Few-shot CoT prompt on MATH500}
\label{lst:base_begin}
\end{listing*}

\begin{listing*}[!ht]
\begin{minted}[fontsize=\footnotesize, frame=single, breaklines]{python}
f'''
{system}
Please reason step by step, and put your final answer within \boxed{{}}.
--------------------------------------------------
{user}
Convert the point $(0,3)$ in rectangular coordinates to polar coordinates.  Enter your answer in the form $(r,\theta),$ where $r > 0$ and $0 \le \theta < 2 \pi.$
'''
\end{minted}
\caption{Zero-shot CoT prompt on MATH500}
\end{listing*}

\begin{listing*}[!ht]
\begin{minted}[fontsize=\footnotesize, frame=single, breaklines]{python}
f'''
I have generated the following responses to the question: Convert the point $(0,3)$ in rectangular coordinates to polar coordinates.  Enter your answer in the form $(r,\theta),$ where $r > 0$ and $0 \le \theta < 2 \pi.$

Response 0: {response0}

...

Response 4: {response4}

Evaluate these responses.
Select the most consistent response based on majority consensus.
Start your answer with "The most consistent response is Response X" (without quotes).
'''
\end{minted}
\caption{Prompt for USC}
\end{listing*}

\begin{listing*}[!ht]
\begin{minted}[fontsize=\footnotesize, frame=single, breaklines]{python}
f'''
{system}
Please reason step by step, and put your final answer within \boxed{{}}.
--------------------------------------------------
{user}
Kevin Kangaroo begins hopping on a number line at 0. He wants to get to 1, but he can hop only $\frac{1}{3}$ of the distance. Each hop tires him out so that he continues to hop $\frac{1}{3}$ of the remaining distance. How far has he hopped after five hops? Express your answer as a common fraction.
'''
\end{minted}
\caption{Prompt for LEAP mistakes}
\label{lst:leap_mistake}
\end{listing*}

\begin{listing*}[!ht]
\begin{minted}[fontsize=\footnotesize, frame=single, breaklines]{python}
f'''
Question: {question}
Generated Reasoning: {response}

Generated Answer: {generated_answer}

Correct Reasoning: {correct_reasoning}

Correct Answer: {correct_answer}

Instruction: Conduct a thorough analysis of the generated answer in comparison to the correct answer. Also observe how the generated reasoning differs from the correct reasoning. Identify any discrepancies, misunderstandings, or errors. Provide clear insights, principles, or guidelines that can be derived from this analysis to improve future responses. We are not focused on this one data point, but rather on the general principle.

Reasoning: <discuss why the generated answer is wrong>
Insights: <what principle should be looked at carefully to improve the performance in the future>

'''
\end{minted}
\caption{Prompt for LEAP low-level principles}
\label{lst:leap_low}
\end{listing*}

\begin{listing*}[!ht]
\begin{minted}[fontsize=\footnotesize, frame=single, breaklines]{python}
f'''
Low-level principles:
{low_level_principles}

Create a list of *unique* and insightful principles to improve future responses based on the analysis above.
Focus on capturing the essence of the feedback while eliminating redundancies.
Ensure that each point is clear, concise, and directly derived from the introspection results.
Create a numbered list of principles. Leave specific details in place.
Limit to at most 8 principles.

List of Principles:
'''
\end{minted}
\caption{Prompt for LEAP high-level principles}
\label{lst:leap_high}
\end{listing*}

\begin{listing*}[!ht]
\begin{minted}[fontsize=\footnotesize, frame=single, breaklines]{python}
f'''
{system}
Please reason step by step, and put your final answer within \boxed{{}}.
--------------------------------------------------
{user}
Please carefully note the following principles:

Principles: 1. **Meticulous Verification**: Always verify each step in algebraic processes to prevent errors that can lead to incorrect conclusions.

... 

8. **Continuous Learning and Adaptation**: Stay open to learning from mistakes and adapting methods to improve future problem-solving approaches.

Kevin Kangaroo begins hopping on a number line at 0. He wants to get to 1, but he can hop only $\frac{1}{3}$ of the distance. Each hop tires him out so that he continues to hop $\frac{1}{3}$ of the remaining distance. How far has he hopped after five hops? Express your answer as a common fraction.

Let's think step by step
Kevin hops $1/3$ of the remaining distance with every hop.
His first hop takes $1/3$ closer.
...

Convert the point $(0,3)$ in rectangular coordinates to polar coordinates.  Enter your answer in the form $(r,\theta),$ where $r > 0$ and $0 \le \theta < 2 \pi.$
'''
\end{minted}
\caption{Prompt for LEAP generations}
\label{lst:base_end}
\end{listing*}

\begin{listing*}[!ht]
\begin{minted}[fontsize=\footnotesize, frame=single, breaklines]{python}
f'''
Kevin Kangaroo begins hopping on a number line at 0. He wants to get to 1, but he can hop only $\frac{1}{3}$ of the distance. Each hop tires him out so that he continues to hop $\frac{1}{3}$ of the remaining distance. How far has he hopped after five hops? Express your answer as a common fraction.

Let's think step by step
Kevin hops $1/3$ of the remaining distance with every hop.
His first hop takes $1/3$ closer.
...

I have generated the following responses to the question: Convert the point $(0,3)$ in rectangular coordinates to polar coordinates.  Enter your answer in the form $(r,\theta),$ where $r > 0$ and $0 \le \theta < 2 \pi.$

Response 0: {response0}

...

Response 4: {response4}

Evaluate these responses.
Select the most consistent response based on majority consensus.
Start your answer with "The most consistent response is Response X" (without quotes).
'''
\end{minted}
\caption{Prompt for USC-w/ Fewshot}
\label{lst:usc_with_fewshot}
\end{listing*}

\begin{listing*}[!ht]
\begin{minted}[fontsize=\footnotesize, frame=single, breaklines]{python}
f'''
{system}
Your job is selecting the most accurate response among multiple candidates. You will receive a question and several candidate answers labeled candidate1, candidate2, etc. Please summarize the debate very briefly and then conclude which single candidate is the most plausible. Output exactly in this format:
Summary: <brief summary>
Conclusion: candidate<number>
Remember to choose only one candidate as the final answer.
--------------------------------------------------
{user}
Please reason step by step, and put your final answer within \boxed{{}}.

The below examples are well-constructed gold question and answer pairs for the same task.

Kevin Kangaroo begins hopping on a number line at 0. He wants to get to 1, but he can hop only $\frac{1}{3}$ of the distance. Each hop tires him out so that he continues to hop $\frac{1}{3}$ of the remaining distance. How far has he hopped after five hops? Express your answer as a common fraction.

Let's think step by step
Kevin hops $1/3$ of the remaining distance with every hop.
His first hop takes $1/3$ closer.
...

Now, let’s select the most proper answer for the given question
Question: Convert the point $(0,3)$ in rectangular coordinates to polar coordinates.  Enter your answer in the form $(r,\theta),$ where $r > 0$ and $0 \le \theta < 2 \pi.$
candidate1: {response 0}
...
candidate5: {response 4}
'''
\end{minted}
\caption{Prompt for LLM-as-Judge}
\label{lst:llm-as-judge}
\end{listing*}

\begin{listing*}[!ht]
\begin{minted}[fontsize=\footnotesize, frame=single, breaklines]{python}
f'''
 "You will receive a QUESTION and its original ANSWER.\n"
"Rewrite ONLY the ANSWER; do NOT alter the QUESTION.\n"
"Treat the original as a 10/10 reference. Produce a deliberately degraded explanation (target quality 1/10):\n"
"- Keep the final answer tokens EXACT (e.g., '\\boxed{...}' or 'The correct answer is (X)').\n"
"- Keep the original CoT style label if present (e.g., 'Let's think step by step:' / 'Reasoning:').\n"
"- Make reasoning weak: shallow, vague, incomplete; omit steps, avoid precise formulas/numbers.\n"
"- Prefer generic phrases over concrete derivations. Lower clarity and rigor compared to the original.\n"
"OUTPUT FORMAT: Return EXACTLY ONE JSON object and NOTHING ELSE:\n"
'{"answer":"<rewritten weaker answer>"}'
'''
\end{minted}
\caption{Prompt for generate weak few-shot}
\label{lst:generate_weak}
\end{listing*}

\begin{listing*}[!ht]
\begin{minted}[fontsize=\footnotesize, frame=single, breaklines]{python}
f'''
 "You are judging FEW-SHOT QUALITY only.\n"
"Compare TWO blocks side-by-side. Assume BLOCK A and BLOCK B are candidate few-shot demonstrations.\n\n"
"Ignore question quality entirely — the question is context only.\n\n"

"What “answer quality” means here:\n"
"- clarity, structure, and coherence of the reasoning.\n"
"- specific steps, concrete numbers/equations when relevant, and justified transitions.\n"
"- a single, clearly marked final answer token format (e.g., \"\\boxed{...}\" or \"The correct answer is (X)\") if present;\n"

"Instructions:\n"
"- Assign an integer score 1-10 to EACH block (higher = better few-shot quality).\n"
"- The evaluation should be comparative: scores must reflect their relative quality.\n"
"- Provide brief notes explaining each score.\n\n"
"OUTPUT FORMAT:\n"
"Return exactly ONE JSON object with this schema (and nothing else):\n"
"{"
"\"A\":{\"score\":int,\"notes\":string},"
"\"B\":{\"score\":int,\"notes\":string},"
"\"comparative_notes\":string"
'''
\end{minted}
\caption{Prompt for judging weak few-shot}
\label{lst:judging_weak}
\end{listing*}

\begin{listing*}[!ht]
\begin{minted}[fontsize=\footnotesize, frame=single, breaklines]{python}
f'''
"Let’s first understand the problem, extract relevant variables and their corresponding numerals, and make a complete plan. Then, let’s carry out the plan, calculate intermediate variables (pay attention to correct numerical calculation and commonsense), solve the problem step by step, and put your final answer within \\boxed{{}}.\n"
'''
\end{minted}
\caption{Prompt for plan-and-solve on MATH500}
\label{lst:plan_and_solve}
\end{listing*}

\begin{listing*}[!ht]
\begin{minted}[fontsize=\footnotesize, frame=single, breaklines]{python}
f'''
"From now on, you are an excellent math teacher and always teach your students math problems correctly. And I am one of your students. Put your final answer within \\boxed{{}}.\n"
'''
\end{minted}
\caption{Prompt for role-playing on MATH500}
\label{lst:role_playing}
\end{listing*}

\end{document}